\documentclass[10pt,journal,cspaper,compsoc]{IEEEtran}

\usepackage[margin=10pt,font=footnotesize,labelfont=bf,labelsep=endash]{caption}

\usepackage{float}
\usepackage[font=footnotesize,labelfont=it]{subcaption}

\usepackage{algorithmic}
\usepackage[linesnumbered, algoruled, vlined]{algorithm2e}

\usepackage{graphicx}
\usepackage{amssymb}
\usepackage{amsmath}
\usepackage{amsfonts}
\usepackage{color}
\usepackage{dsfont}
\usepackage{upgreek}
\usepackage{bbm}
\usepackage{url}
\usepackage{multirow}
\usepackage{xspace}
\usepackage{eucal}
\usepackage{proof}

\usepackage{hyperref}

\usepackage[table]{xcolor}

\usepackage{arydshln}

\newcommand{\fnorm}[2][2]{\ensuremath{ \left\| #2 \right\|^2_{ \mathrm{#1} } } }
\newcommand\norm[1]{\left\lVert#1\right\rVert}
\newcommand{\ie}{i.e.\@\xspace}
\newcommand{\eg}{e.g.\@\xspace}
\newcommand{\etc}{etc.\@\xspace}
\newcommand{\etal}{et al.\@\xspace}

\newcommand{\st}{{{\rm s.t.}\!:}\xspace}

\newcommand{\argmax}{\operatornamewithlimits{argmax}}

\newcommand{\trace}{\operatornamewithlimits{Tr}}
\newcommand{\half}{{\frac{1}{2}}}

\newcommand{\crf}{{CRF}\xspace}
\newcommand{\mrf}{{MRF}\xspace}
\newcommand{\cnn}{{CNN}\xspace}
\newcommand{\dcnf}{{DCNF}\xspace}
\newcommand{\dcnfsp}{{DCNF-FCSP}\xspace}

\def\Real{\mathbb{R}}

\def\regress{{z}}
\def\bregress{{\bf z}}
\def\pws{{R}}
\def\bpws{{\bf R}}

\def\x{{\bf x}}
\def\y{{\bf y}}
\def\h{{\bf h}}

\def\I{{\bf I}}
\def\S{{\bf S}}
\def\D{{\bf D}}
\def\A{{\bf A}}

\def\J{{\bf J}}
\def\C{{\bf C}}
\def\W{{\bf W}}
\def\bbeta{{\boldsymbol \beta}}
\def\btheta{{\boldsymbol \theta}}

\def\T{{\mathrm{T}}}
\def\Pr{{\mathrm{Pr}}}

\def\zeros{{\boldsymbol  0}}

\def\T{{\!\top}}

\begin{document}

\title{Learning Depth from Single Monocular Images Using Deep Convolutional Neural Fields}

\author{
Fayao Liu,
Chunhua Shen,
Guosheng Lin,
Ian Reid
\IEEEcompsocitemizethanks{
\IEEEcompsocthanksitem
Authors are with  the School of Computer Science, The University of Adelaide,
Australia.
C. Shen, G. Lin and I. Reid are also with
ARC Centre of Excellence for Robotic Vision.
}
\thanks{}
}

\markboth{}{}

\IEEEcompsoctitleabstractindextext{%
\begin{abstract}

  In this article, we tackle the problem of depth estimation from single monocular images.
  Compared with depth estimation using multiple images such as stereo depth perception, depth from monocular images is
  much more challenging.
  Prior work typically focuses on exploiting geometric priors or
  additional sources of information, most using hand-crafted features.
  Recently, there is mounting evidence that features from deep convolutional neural networks (\cnn)
  set new records for various vision applications.
  On the other hand, considering the continuous characteristic of the depth values,
  depth  estimation can be naturally formulated as a continuous conditional random field (\crf) learning problem.
  Therefore, here we present a deep convolutional neural field model for estimating depths from  single
  monocular images, aiming to jointly explore the capacity of deep \cnn and continuous \crf.
  In particular, we propose a deep structured learning scheme which learns the unary and pairwise potentials
  of continuous \crf in a unified deep \cnn framework.
  We then further propose an  equally effective model based on fully convolutional networks
  and a novel superpixel pooling method, which is about 10  times faster, to speedup the patch-wise convolutions in the deep model.
  With this more efficient model, we are able to design deeper networks to pursue better performance.

Our proposed method can be used for depth estimation of general scenes
with no geometric priors nor any extra information injected.
In our case, the integral of the partition function
can be  calculated in a closed form such that we can exactly solve the log-likelihood maximization.
Moreover, solving the inference problem for predicting depths of a test image is highly efficient as closed-form solutions exist.
Experiments on both indoor and outdoor scene datasets demonstrate that the proposed method outperforms
state-of-the-art depth estimation approaches.

\end{abstract}


}

\maketitle

%

\section{Introduction}

Estimating depth information from single monocular images depicting general scenes is an important problem in computer vision.
Many challenging computer vision problems have proven to  benefit from the incorporation of depth information,
to name a few, semantic labellings \cite{Ladicky_cvpr14}, pose estimations \cite{Shotton_pami13}.
Although the highly developed depth sensors such as Microsoft Kinect nowadays have made the acquisition of RGBD
images affordable, most of the vision datasets commonly evaluated among the vision community are still RGB images.
Moreover, outdoor applications still rely on LiDAR or other laser sensors due to the fact that strong sunlight can
cause infrared interference and make depth information extremely noisy.
This has led to wide research interest on the topic of estimating depths from single RGB images.
Unfortunately, it is a notoriously ill-posed problem,
as one captured image scene may correspond to numerous real world scenarios \cite{dcnn_nips14}.

\begin{figure} \centering
\resizebox{.98\linewidth}{!} {
\begin{tabular}{ccc}
	\includegraphics[width=0.32\textwidth]{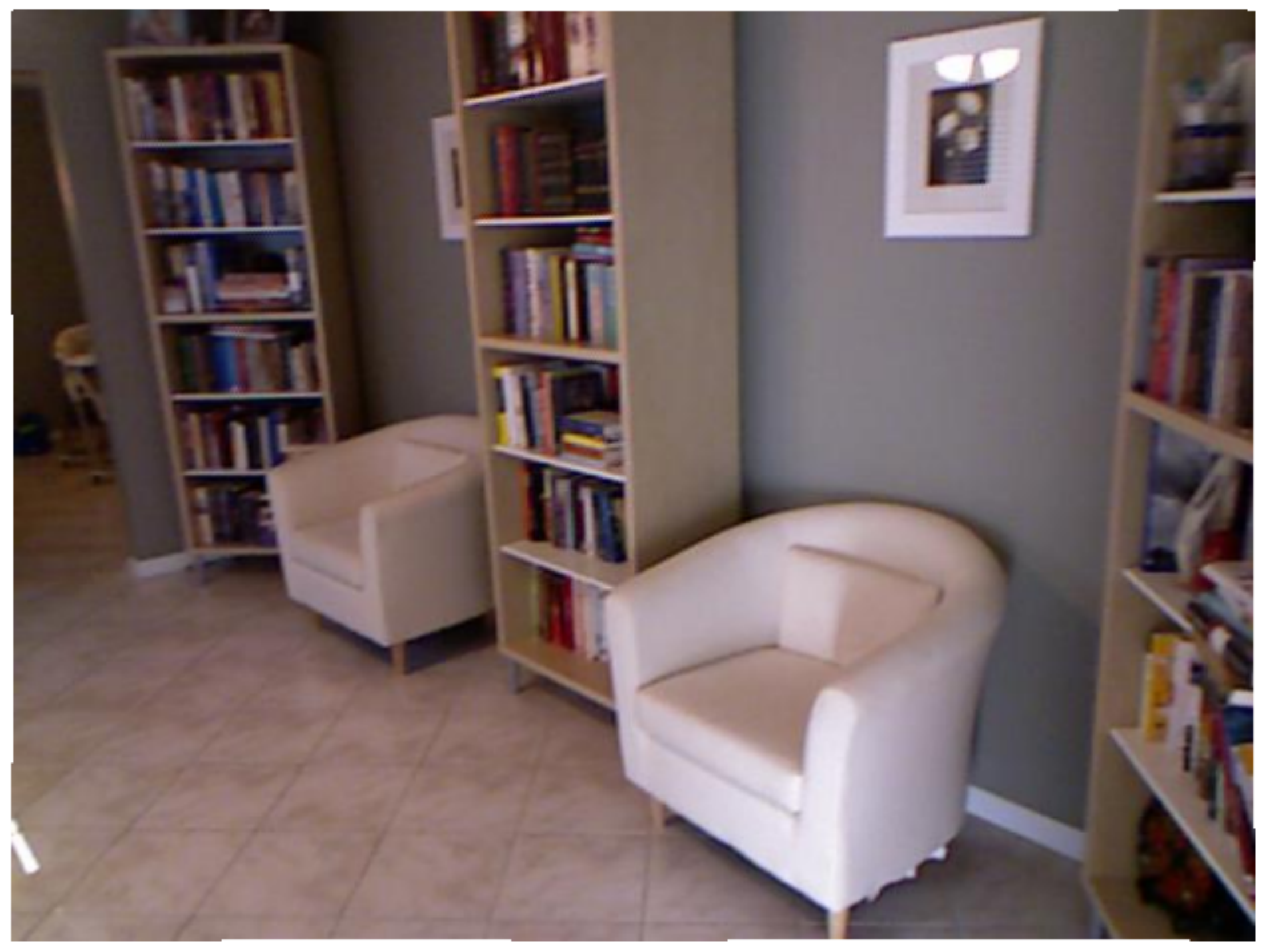}
	\includegraphics[width=0.32\textwidth]{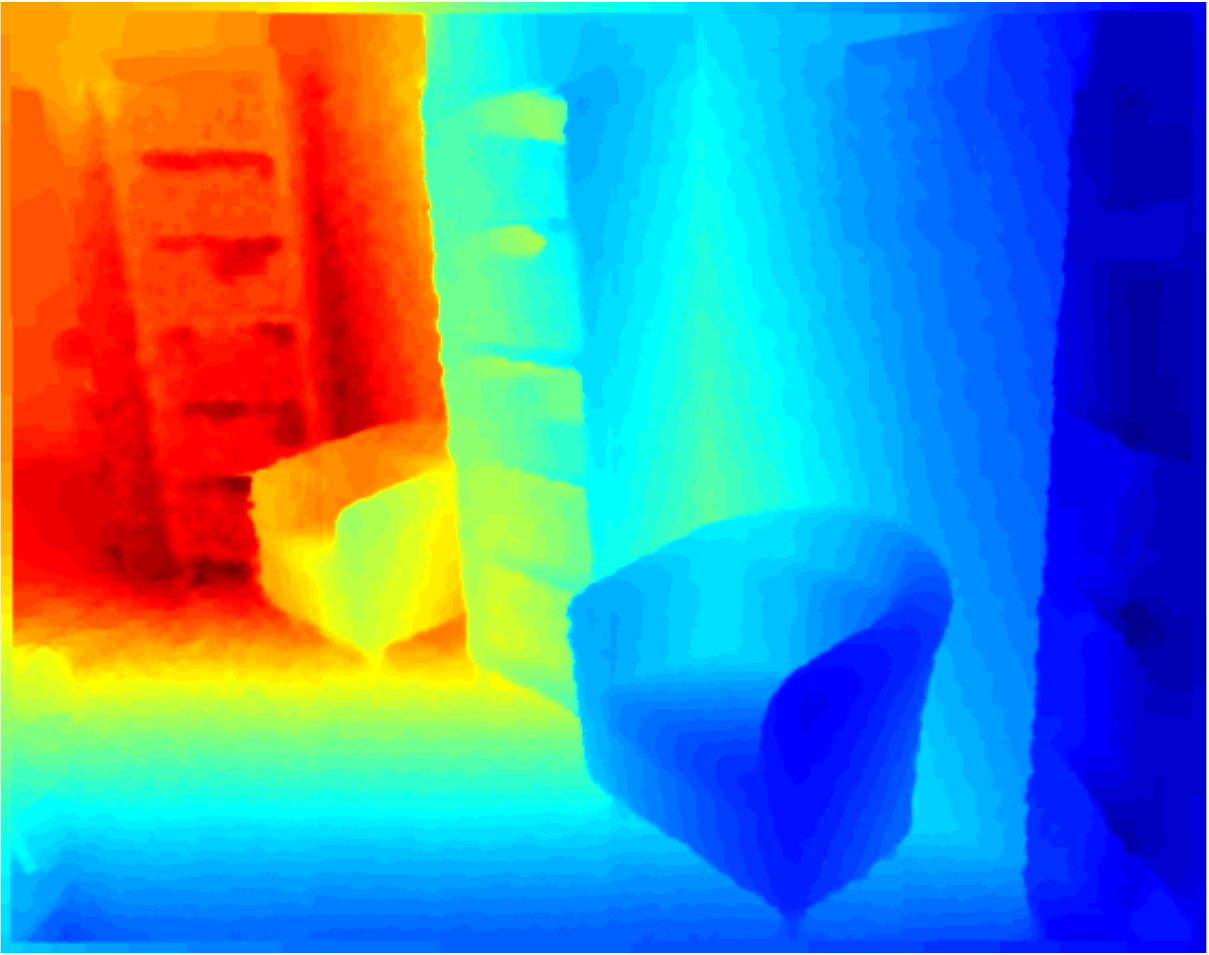}
	\includegraphics[width=0.32\textwidth]{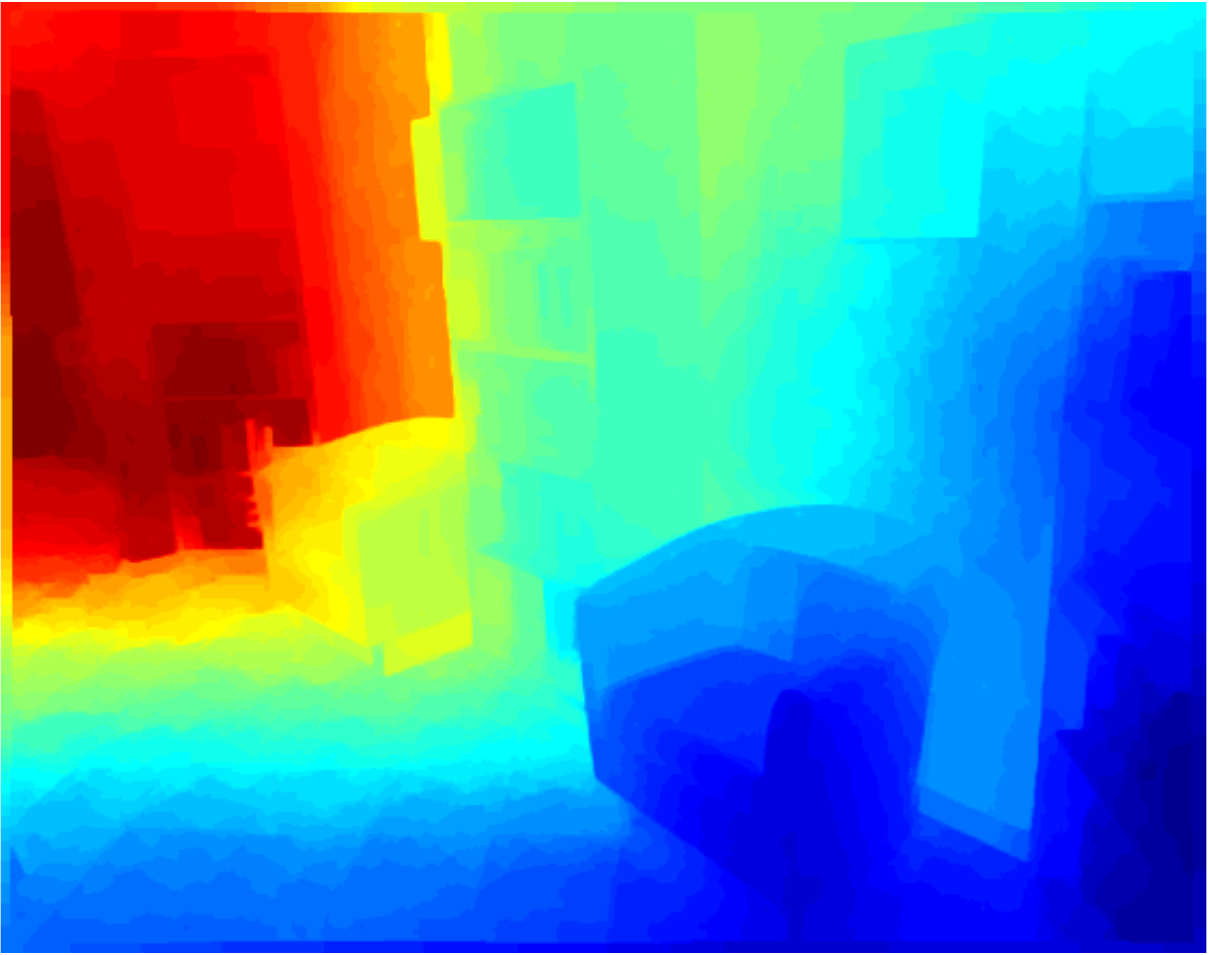} \\

	\includegraphics[width=0.32\textwidth]{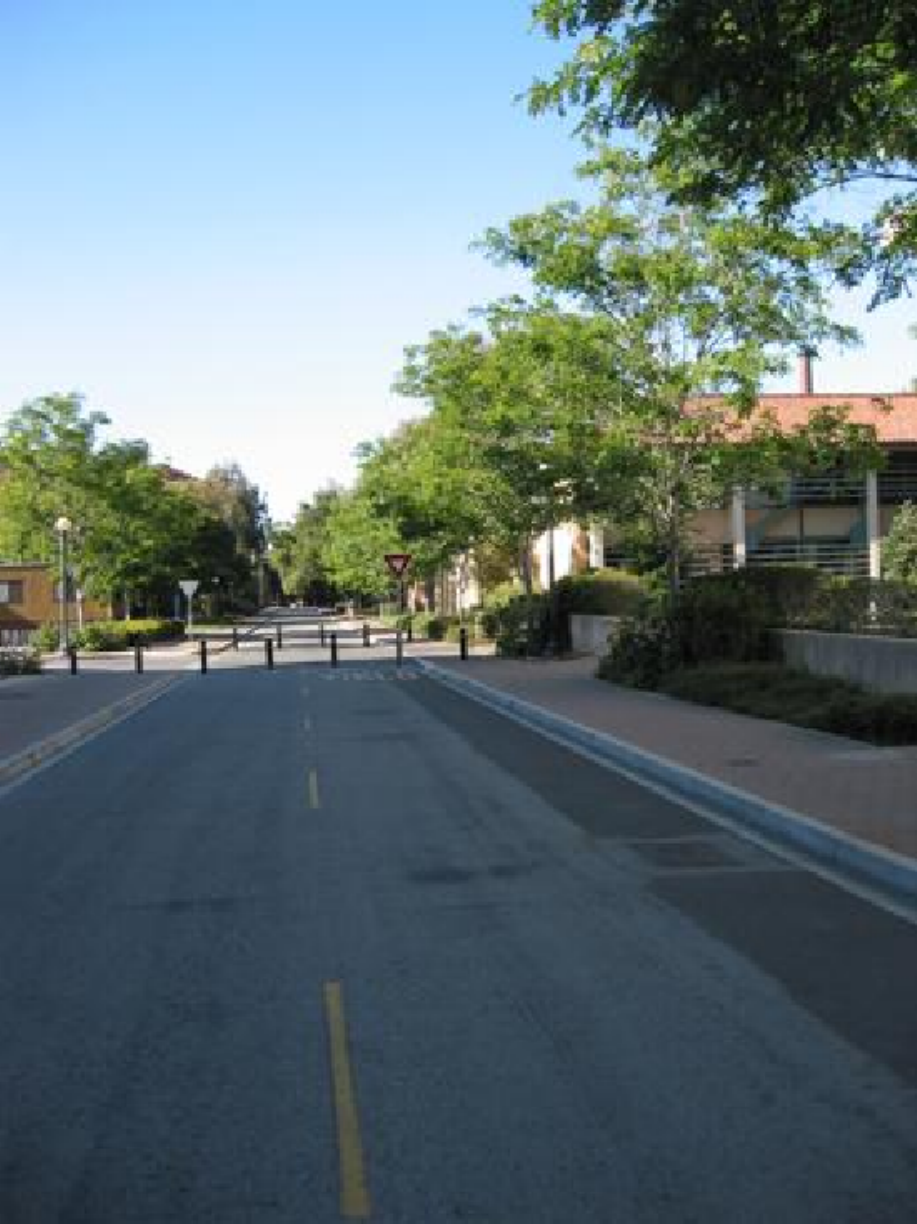}
	\includegraphics[width=0.32\textwidth]{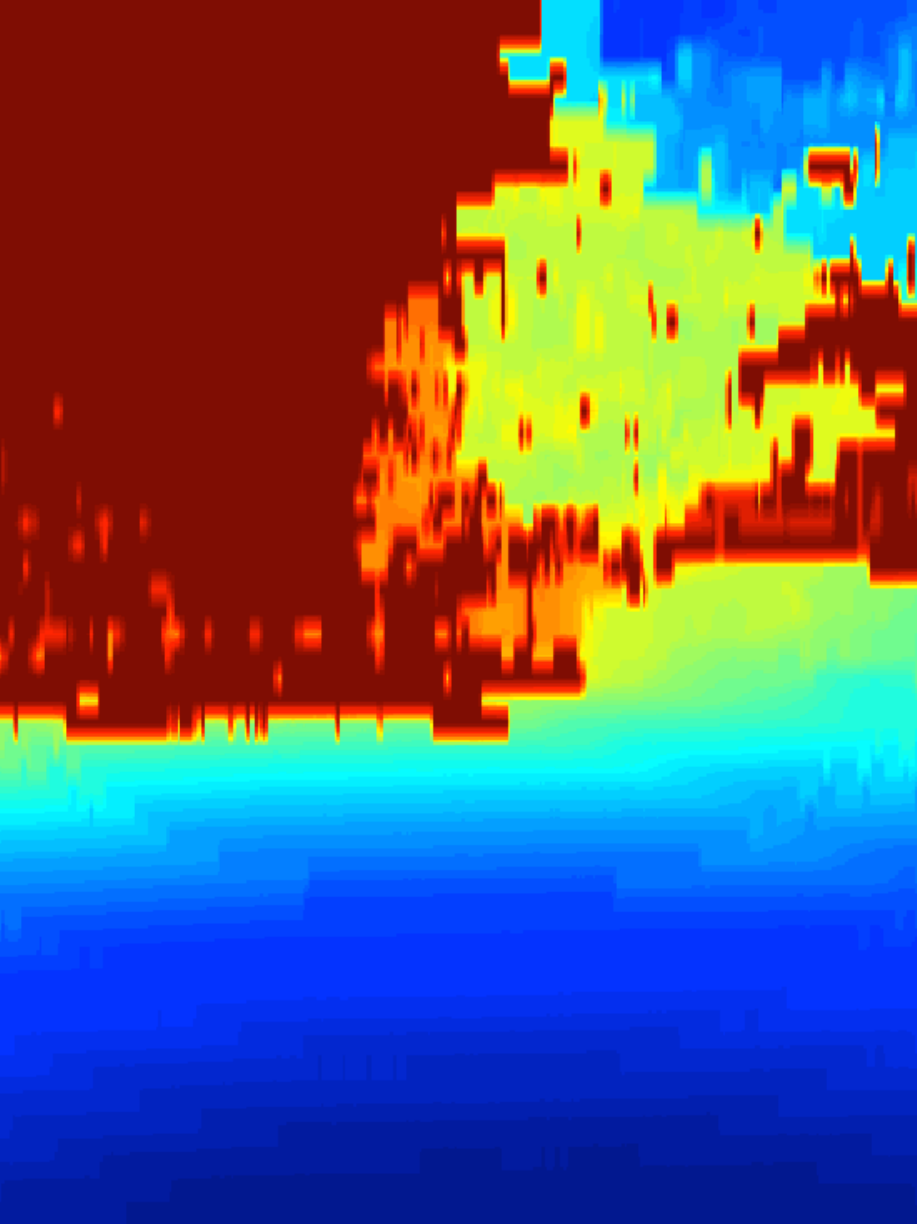}
	\includegraphics[width=0.32\textwidth]{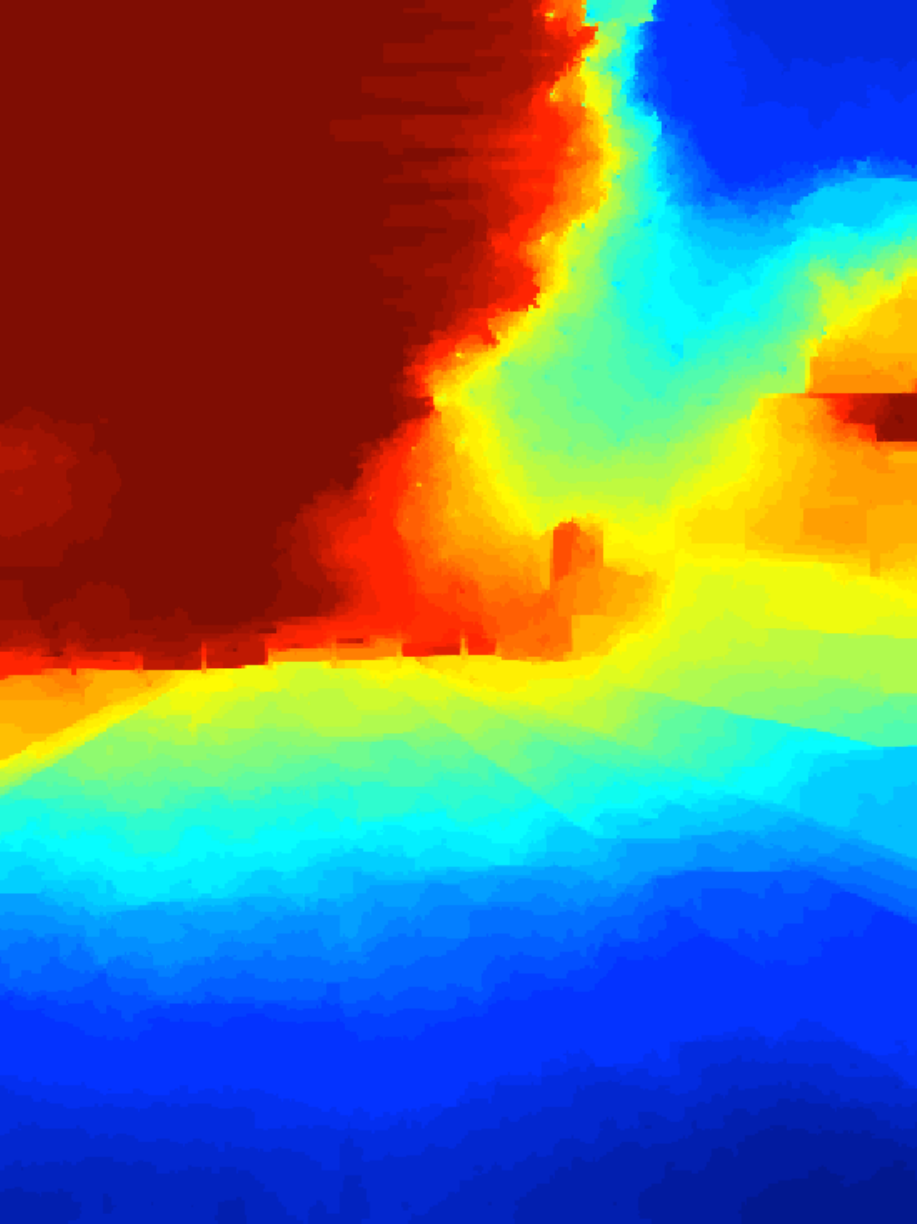}
\end{tabular}
}
\caption{Examples of depth estimation results using the proposed deep convolutional neural fields model.
First row: NYU v2 dataset;
second row: Make3D dataset.
From left to right: input image, ground-truth, our prediction.}  \label{fig:examples}
\end{figure}

Whereas for humans, inferring the underlying 3D structure from a single image is
effortless,
it remains a challenging task for automated computer vision systems to do so
since no reliable cues can be exploited, such as temporal information in videos, stereo correspondences, etc.
Previous work mainly focuses on enforcing geometric assumptions, \eg, box models, to infer the spatial layout of a room \cite{Hedau_eccv10,Lee_nips10} or outdoor scenes \cite{Gupta_eccv10}.
These models come with innate restrictions,
which are limitations  to model only particular scene structures and therefore are not applicable for general scene depth estimations.
More recently, non-parametric methods \cite{depthTransfer_pami14} are explored, which consists of candidate images retrieval, scene alignment  and then depth inference using optimizations with smoothness constraints.
This is based on the assumption that scenes with semantically similar appearances should have similar depth distributions when densely aligned.
However, this method is prone to propagate errors through the different decoupled stages and relies heavily on building a reasonable sized image database to perform the candidate retrieval.
In recent years, efforts have been made towards incorporating additional sources of information, \eg, user annotations \cite{Russell_cvpr09}, semantic labellings \cite{Liu_cvpr12, Ladicky_cvpr14}.
In the recent work of \cite{Ladicky_cvpr14}, Ladicky \etal have shown that jointly performing depth estimation and semantic labelling can benefit each other.
Nevertheless, all these methods use hand-crafted features.

In contrast to previous efforts, here we propose to
formulate the depth estimation as a deep continuous Conditional Random Fields  (\crf)
learning problem, without relying on any geometric priors or any extra information.
\crf \cite{Lafferty_icml01} is a popular graphical model  for structured output predictions.
While extensively studied in classification (discrete) domains, \crf has been less explored for regression (continuous) problems.
One of the pioneer work on continuous \crf can be attributed to \cite{ccrf_nips08}, in which it was proposed for global ranking in document retrieval.
Under certain constraints, they can directly solve the maximum likelihood optimization as the partition function
can be analytically calculated.
Since then, continuous \crf has been successfully
applied for solving various structured regression problems, \eg, remote sensing \cite{ccrf_ecai10,ccrf_aaai13}, image denoising \cite{ccrf_aaai13}.
Motivated by these successes, we here propose to use it for depth estimation,
given the continuous nature of the depth values, and learn the potential functions in a deep convolutional neural network (\cnn).

Recent years have witnessed the prosperity of the deep
\cnn \cite{cnn_Lecun89} since the breakthrough work of Krizhevsky
\etal \cite{AlexNet12}.
CNN features have been setting new records for a wide variety of vision applications \cite{astound_cvprW14}.
Despite all the successes in classification problems,
deep \cnn has been less explored for structured learning problems, \ie, joint training of a deep \cnn and a graphical model, which is a relatively new and not well addressed problem.
To our knowledge, no such model has been successfully used for depth estimations.
Here, we bridge this gap by jointly exploring \cnn and continuous \crf, denoting this new method as a deep convolutional neural field (\dcnf).

Fully convolutional networks have recently been studied for dense prediction problems, \eg, semantic labelling \cite{Long_arxiv14,Cogswell_arxiv14}.
Models based on fully convolutional networks have the advantage of highly efficient training and prediction.
We here exploit this advance to speedup the training and prediction of our \dcnf model.
However, the feature maps produced by the fully convolutional models are typically much smaller than the input image size.
This can cause problems for both training and prediction.
During training, one needs to downsample the ground-truth maps,
which may lead to information loss since small objects might disappear.
In prediction, the upsampling operation is likely to  bring in degraded
performance at the object boundaries.
We therefore propose a novel superpixel pooling method to address this issue.
It jointly exploits the strengths of highly efficient fully convolutional networks and the benefits of superpixels at preserving object boundaries.

To sum up, we highlight the main contributions of this work as follows.
\begin{itemize}
\item
We propose a deep convolutional neural field (\dcnf) model for depth estimations by exploring \cnn and continuous \crf.
Given the continuous nature of the depth values, the partition function in the probability density function can be analytically calculated, therefore we can directly solve the log-likelihood optimization without any approximations.  The gradients can be exactly calculated in the back propagation training.
Moreover, solving the MAP problem for predicting the depth of a new image is highly efficient since closed form solutions exist.
\item
We jointly learn the unary and pairwise potentials of the \crf in a unified deep \cnn framework, which is trained using back propagation.
\item
We propose a faster model based on fully convolutional networks and a novel superpixel pooling method, which results in $\sim 10$  times speedup while producing similar prediction accuracy.
With this more efficient model, which we refer as \dcnfsp , we are able to design very deep networks for better performance.
\item
We demonstrate that the proposed method outperforms state-of-the-art results of depth estimation on both indoor and outdoor scene datasets.
\end{itemize}

Preliminary results of our work appeared in Liu \etal \cite{CVPR2015LIUDepth}.

\subsection{Related work}
Our method exploits the recent advances of deep nets in image classification \cite{AlexNet12,vgg16_arxiv},  object detection \cite{RCNN14} and semantic segmentation \cite{Long_arxiv14,Cogswell_arxiv14},
for single view image depth estimations.
In the following, we give a brief introduction to the most closely related work.

{\bf Depth estimation}
Traditional methods \cite{Saxena_nips05,make3d_pami09,Liu_cvpr12}
typically formulate the depth estimation as a Markov Random Field (\mrf) learning problem.
As exact \mrf learning and inference are intractable in general,
most of these approaches employ approximation methods, \eg, multi-conditional learning (MCL) or particle belief propagation (PBP).
Predicting the depths of a new image is inefficient, taking around 4-5s in \cite{make3d_pami09} and even longer (30s) in \cite{Liu_cvpr12}.
To make things worse, these methods suffer from lacking of flexibility in that \cite{Saxena_nips05,make3d_pami09}  rely on horizontal alignment of images and
\cite{Liu_cvpr12} requires the semantic labellings of the training data available beforehand.
More recently, Liu \etal \cite{Miaomiao_cvpr14} propose a discrete-continuous \crf model to take
into consideration the relations between adjacent superpixels, \eg, occlusions.
They also need to use approximation methods for learning and maximum a posteriori (MAP) inference.
Besides, their method relies on image retrieval to obtain a reasonable initialization.
In contrast, here we present a deep continuous \crf model in which we can
directly solve the log-likelihood optimization without any approximations, since the partition function can be analytically calculated.
Predicting the depth of a new image is highly efficient since a closed form solution exists.
Moreover, we do not inject any geometric priors nor any extra information.
On the other hand, previous methods \cite{make3d_pami09,Liu_cvpr12,depthTransfer_pami14,Miaomiao_cvpr14,Ladicky_cvpr14} all use hand-crafted features in their work, \eg, texton, GIST, SIFT, PHOG, object bank, etc.
In contrast, we learn deep \cnn for constructing unary and pairwise potentials of the \crf.

Recently, Eigen \etal \cite{dcnn_nips14} proposed a multi-scale \cnn approach for depth estimation, which bears similarity to our work here.
However, our method differs critically from theirs:
they use the \cnn as a black-box by directly regressing the depth map from an input image through convolutions; in contrast we use a \crf to explicitly model the relations of neighboring superpixels, and learn the potentials (both unary and binary) in a unified \cnn framework.

Recent work of \cite{haosu_sig14} and \cite{Kar_cvpr15} is relevant to ours in that they also perform depth estimation from a single image.
The method of Su \etal \cite{haosu_sig14} involves a continuous depth optimization step like ours, which also contains a unary regression term and a pairwise local smoothness term.
However, these two works focus on 3D reconstruction of known segmented objects while our method targets at depth estimation of general scene images.
Furthermore, the method of \cite{haosu_sig14} relies on a pre-constructed 3D shape database of input object categories, and
the work of \cite{Kar_cvpr15} relies on class-specific object keypoints and object segmentations.
In contrast, we do not inject these priors.

{\bf Combining \cnn and \crf}
In \cite{Lecun13}, Farabet \etal propose a multi-scale \cnn framework for scene labelling,
which uses \crf as a post-processing step for local refinement.
In the most recent work of \cite{Lecun_nips14}, Tompson \etal present a hybrid architecture for jointly training a deep \cnn and an \mrf for human pose estimation.
They first train a unary term and a spatial model separately, then jointly learn them as a fine tuning step.
During fine tuning of the whole model, they simply remove the partition function in the likelihood to have a loose approximation.
In contrast, our model performs continuous variable prediction.
We can directly solve the log-likelihood optimization without using approximations as the partition function is integrable and can be analytically calculated.
Moreover, during prediction, we have closed-form solutions to the MAP inference problem.
Although no convolutional layers are involved, the work of \cite{Baltrusaitis2014} shares similarity with ours in
that both continuous \crf's use neural networks to model the potentials. Note that the model in \cite{Baltrusaitis2014} is not
deep and only one hidden layer is used. It is unclear how the method of \cite{Baltrusaitis2014} performs
on the challenging depth estimation problem that we consider here.

{\bf Fully convolutional networks}
Fully convolutional networks have recently been actively studied for dense prediction problems, \eg, semantic segmentation \cite{Long_arxiv14,Cogswell_arxiv14}, image restoration \cite{Eigen_iccv13}, image super-resolution \cite{Dong_eccv14}, depth estimations \cite{dcnn_nips14}.
To deal with the downsampled output issue,
interpolations are generally applied  \cite{dcnn_nips14,Cogswell_arxiv14}.
In \cite{overfeat}, Sermanet \etal propose an input shifting and output interlacing trick to produce dense predictions from coarse outputs without interpolations.
Later on, Long \etal \cite{Long_arxiv14} present a deconvolution approach to put the upsampling into the training regime instead of applying it as a post-processing step.
The \cnn model presented in Eigen \etal \cite{dcnn_nips14} for depth estimation also suffers from this
upsampling problem---the predicted depth maps of \cite{dcnn_nips14} is 1/4-resolution of the original input image with some border areas lost.
They simply use bilinear interpolations to upsample the predictions to the input image size.
Unlike these existing methods, we propose a novel superpixel pooling method to address this issue.
It jointly exploits the strengths of highly efficient fully convolutional networks and the benefits of superpixels at preserving object boundaries.

\begin{figure*}[!t] \center
	\includegraphics[width=0.88\textwidth]{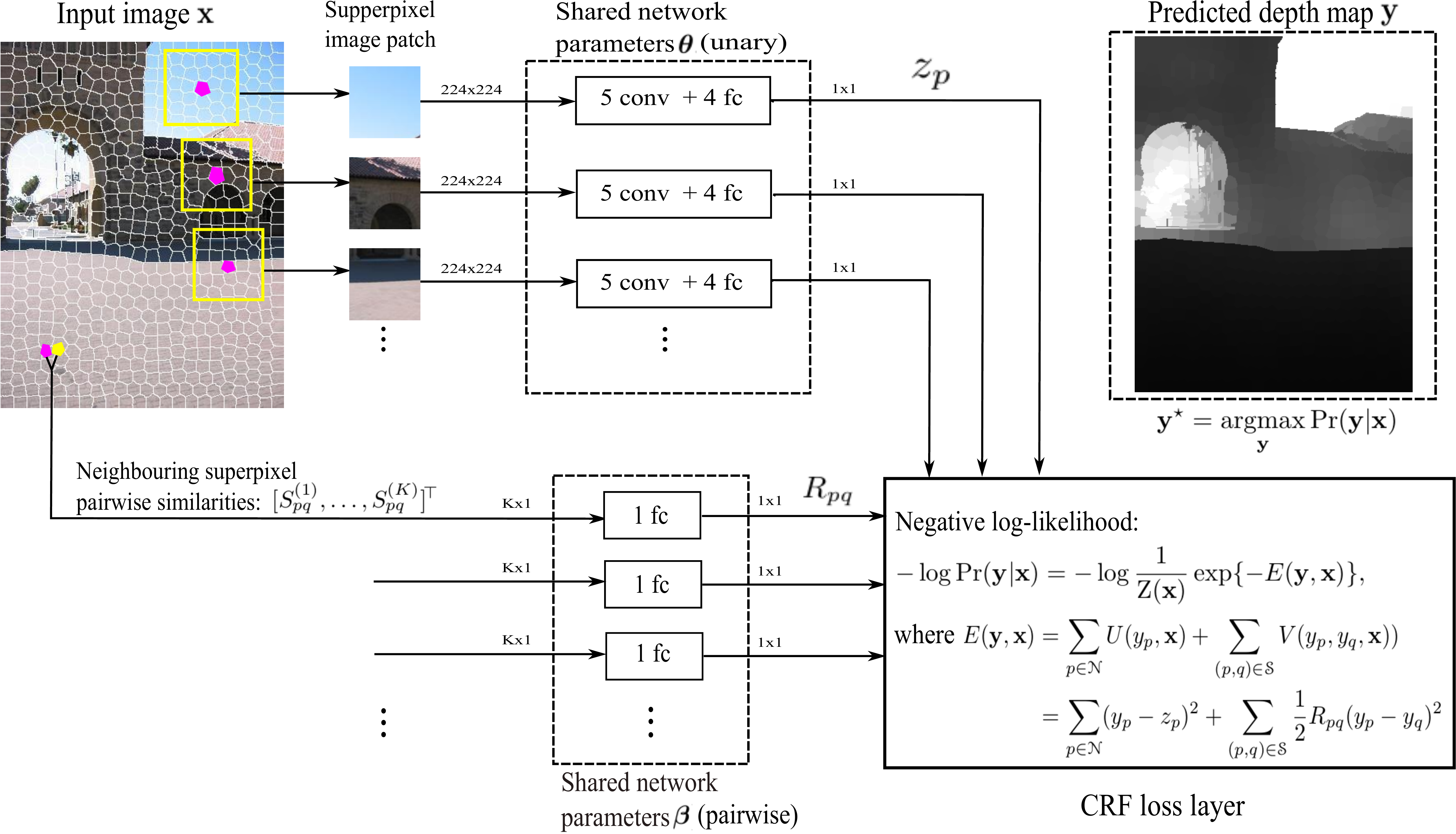}
\caption{An illustration of our \dcnf model for depth estimation.
The input image is first over-segmented into superpixels.
In the unary part, for a superpixel $p$, we crop the image patch centred around its centroid,
then resize and feed it to a \cnn which is composed of 5 convolutional and 4 fully-connected layers (details refer to Fig. \ref{fig:dcnf_unary}).
In the pairwise part,
for a pair of neighboring superpixels $(p, q)$, we consider $K$ types of similarities, and feed them into a fully-connected layer.
The outputs of unary part and the pairwise part are then fed to the \crf structured loss layer, which minimizes the negative log-likelihood.
Predicting the depths of a new image $\x$ is to maximize the conditional probability $\Pr(\y|\x)$, which has closed-form solutions (see Sec. \ref{sec:learning} for details).
}  \label{fig:dcnf}
\end{figure*}

\begin{figure*}[!t] \center
	\includegraphics[width=0.85\textwidth]{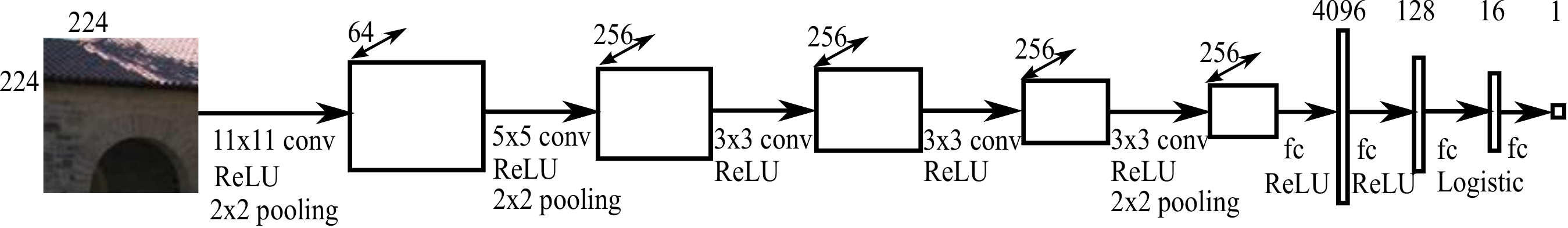}
\caption{Detailed network architecture of the unary part in Fig. \ref{fig:dcnf}. }  \label{fig:dcnf_unary}
\end{figure*}

\section{Deep convolutional neural fields}
We present the details of our deep convolutional neural field (\dcnf) model for depth estimation in this section.
Unless otherwise stated, we use boldfaced uppercase and lowercase letters to denote matrices and column vectors respectively;
$\zeros$ represents a column vector with all elements being $0$.

\subsection{Overview}
The goal here is to infer the depth of each pixel in a single image depicting a general scene.
Following the work of \cite{make3d_pami09,Liu_cvpr12,Miaomiao_cvpr14},
we make the common assumption that an image is composed of small homogeneous regions (superpixels)
and consider the graphical model composed of nodes defined on superpixels.
Each superpixel is portrayed by the depth of its centroid.
Let $\x$ be an image and $\y=[y_1,\ldots, y_n]^{\T} \in \Real^n$ be
a vector of continuous depth values corresponding to
 all $n$ superpixels in $\x$.
 Similar to conventional CRFs,
 we model the conditional probability distribution of the data with the following density function:
\begin{equation}\label{eq:prob}
\begin{aligned}
\Pr(\y|\x) = \frac{1}{\mathrm{Z(\x)}} \exp (- E(\y, \x)),
\end{aligned}
\end{equation}
where $E$ is the energy function; $\mathrm{Z}$ is the partition function  defined as:
\begin{equation}\label{eq:partition}
\mathrm{Z}(\x) = \int_{\y} \exp \{ -E(\y, \x) \}\mathrm{d}\y.
\end{equation}
Here, because $\y$ is continuous, the integral in Eq. \eqref{eq:prob} can be analytically calculated
under certain circumstances, which we will show in Sec. \ref{sec:learning}.
This is different from the discrete case, in which approximation methods need to be applied.
To predict the depth of a new image, we solve the following MAP inference problem:
\begin{align} \label{eq:inference}
\y^{\star}&=\argmax_{\y} \Pr(\y|\x).
\end{align}

We formulate the energy function as a typical combination of unary potentials $U$ and pairwise potentials $V$
over the nodes (superpixels) $\cal N$ and edges $\cal S$ of the image $\x$:
\begin{equation}\label{eq:energy}
E(\y, \x) = \sum_{p \in {\cal N} } U(y_{p}, \x)
	 + \sum_{(p,q) \in {\cal S}} V(y_{p}, y_{q}, \x).
\end{equation}
The unary term $U$ aims to regress the depth value for a single superpixel.
The pairwise term $V$ encourages neighboring superpixels with similar appearances to take similar depths.
We aim to jointly learn $U$ and $V$ in a unified \cnn framework.

Fig. \ref{fig:dcnf} sketches our
deep convolutional neural field model
for depth estimation.
The whole network comprises a unary part, a pairwise part and a continuous \crf loss layer.
For an input image, which has been over-segmented into $n$ superpixels,
we consider image patches centred around each superpixel centroid.
The unary part then takes an image patch as input and feeds it to a \cnn
whose output is a single number, the regressed depth value of the superpixel.

The network for the unary part is composed of $5$ convolutional and $4$ fully-connected layers with details in Fig. \ref{fig:dcnf_unary}.  Note that the \cnn parameters are shared across all the superpixels.
The pairwise part takes similarity vectors (each with $K$ components)  of all neighboring superpixel pairs as input and feeds
each of them to a fully-connected layer (parameters are shared among different pairs),
then outputs a vector containing all the $1$-dimensional similarities for each of the neighboring superpixel pairs.
The continuous
\crf loss layer takes the outputs from the unary and the pairwise terms to minimize the negative log-likelihood.
Compared to the direct regression method in \cite{dcnn_nips14},
our model possesses two potential advantages:
\begin{itemize}
  \item
We achieve translation invariance as we construct unary potentials irrespective of the superpixel's coordinate (shown in Sec. \ref{sec:potentials});
 \item
   We explicitly model the relations of neighboring superpixels through pairwise potentials.
\end{itemize}

In the following, we describe the details of potential functions involved in the energy function in Eq. \eqref{eq:energy}.

\subsection{Potential functions}
\label{sec:potentials}

\subsubsection{Unary potential}
The unary potential is constructed from the output of a \cnn by considering the least square loss:
\begin{equation}\label{eq:unary}
U(y_{p}, \x;\btheta) = (y_p - \regress_p (\btheta))^2, \;\; \forall p=1,...,n.
\end{equation}
Here $\regress_p$ is the regressed depth of the superpixel $p$ parametrized by the \cnn parameters $\btheta$.

The network architecture for the unary part is depicted in Fig. \ref{fig:dcnf_unary}.
Our \cnn model in Fig. \ref{fig:dcnf_unary} is mainly inspired by
the well-known network architecture of Krizhevsky \etal \cite{AlexNet12} with modifications.
It is composed of $5$ convolutional layers and $4$ fully-connected layers.
The input image is first over-segmented into superpixels, then for each superpixel, we consider the image patch centred around its centroid. Each of the image patches is resized to $ 224 \times 224$ pixels
(other resolutions also work) and then fed to the convolutional neural network.
Note that the convolutional and the fully-connected layers are shared across all the image patches of different superpixels.
Rectified linear units (ReLU) are used as activation functions for the five convolutional layers and the first two fully connected layers.
For the third fully-connected layer, we use the logistic function $f(x)=(1+e^{-x})^{-1}$ as the activation function.
The last fully-connected layer plays the role of model ensemble with no activation function followed.
The output is an $1$-dimensional real-valued depth for a single superpixel.

\subsubsection{Pairwise potential}
We construct the pairwise potential from $K$ types of similarity observations, each of which enforces smoothness by exploiting consistency information of neighboring superpixels:
\begin{align} \label{eq:pairwise}
&V(y_{p}, y_{q}, \x; \bbeta) = \half \pws_{pq} (y_p - y_q)^2, \;\forall p,q=1,...,n.
\end{align}
Here $\pws_{pq}$ is the output of the network in the pairwise part (see  Fig. \ref{fig:dcnf}) from a neighboring superpixel pair $(p,q)$.
We use a fully-connected layer here:
\begin{align}  \label{eq:def_R}
\pws_{pq} = \bbeta^{\T} [S_{pq}^{(1)},\ldots,S_{pq}^{(K)}]^{\T} = \sum_{k=1}^K \beta_k S_{pq}^{(k)},
\end{align}
where $\S^{(k)}$ is the $k$-th similarity matrix whose elements are $S_{pq}^{(k)}$ ($\S^{(k)}$ is symmetric);
$\bbeta=[\beta_1, \ldots, \beta_k]^{\T}$
are the network parameters.
From Eq. \eqref{eq:def_R}, we can see that we do not use any activation function.
However, as our framework is general, more complicated networks may be seamlessly incorporated for the pairwise part.
In Sec.~\ref{sec:learning}, we will show that we can derive a general
form for calculating the gradients with respect to $\bbeta$ (see Eq.~\eqref{eq:derive_beta_final}).
To ensure that
$ {\rm Z}(x)$ in Eq. \eqref{eq:partition} is integrable, we require $\beta_k \geq 0$
as in \cite{ccrf_nips08}. Note that this is a sufficient but not necessary condition.

  Here we consider $3$ types of pairwise similarities,
    measured by the colour difference, colour histogram difference and texture disparity
    in terms of local binary patterns (LBP) \cite{LBP_icpr94}, which take the conventional form:
\[
    S_{pq}^{(k)} = e^{-\gamma \norm{s_p^{(k)} - s_q^{(k)}}}, \; k=1,2,3;
\]
    where $s_p^{(k)}$, $s_q^{(k)}$ are the observation values of the superpixel $p$,
    $q$ calculated from colour, colour histogram and LBP;
    $\norm{\cdot}$ denotes the $\ell_2$ norm of a vector and $\gamma$ is a constant.
    It may be possible to learn features for the pairwise term too.
    For example,  the pairwise term can be a deep CNN with raw pixels as the input.
    A more sophisticated pairwise energy may further improve the estimation,
    especially for complex discrete labelling problems,
    with the price of increased computation complexity.
    For depth estimation, we find that our current pairwise energy already works very well.

\subsection{Learning}
\label{sec:learning}
With the unary and the pairwise potentials defined in Eq. \eqref{eq:unary}, \eqref{eq:pairwise}, we can now write the energy function as:
\begin{equation}\label{eq:feature}
E(\y, \x) = \sum_{p \in {\cal N} } (y_p - \regress_p)^2
+ \sum_{(p,q) \in {\cal S}} \half \pws_{pq} (y_p - y_q)^2.
\end{equation}
For ease of expression, we introduce the following notation:
\begin{align} \label{eq:def_Amat}
 \A = \I+ \D - \bpws,
\end{align}
where $\I$ is the $ n \times n$ identity matrix; $\bpws$ is the affinity
matrix composed of $\pws_{pq}$;  $\D$ is a diagonal matrix with $ D_{pp} = \sum_q \pws_{pq}$.
We see that $  \D - \bpws  $ is the graph Laplacian matrix. Thus $  \A  $ is the regularized Laplacian matrix.
Expanding Eq. \eqref{eq:feature}, we have:
\begin{align} \label{eq:feature_expand}
E(\y, \x) = \y^\T \A \y - 2\bregress^\T \y + \bregress^\T \bregress.
\end{align}
Due to the quadratic terms of $\y$ in the energy function in Eq. \eqref{eq:feature_expand} and the positive definiteness of $\A$
(with all positive edges of the graph, the Laplacian matrix must be positive
semidefinite and therefore $ \A $ must be positive definite), we can analytically calculate the integral in the partition function (Eq. \eqref{eq:partition}) as:
\begin{align} \label{eq:partition_expand}
  {\rm Z}(\x) &=\int_\y \exp \Big\{ -E(\y, \x) \Big\}\mathrm{d}\y   \notag \\
	&=\exp \{  - \bregress^\T \bregress \} \int_{\y} \exp \Big\{ - \y^\T \A \y + 2\bregress^\T \y \Big\}\mathrm{d}\y \notag  \\
	&= \frac{{(\pi)}^{\frac{n}{2}}}{{|\A|}^{\frac{1}{2}}} \exp \{{ \bregress ^\T \A^{-1} \bregress   - \bregress^\T \bregress }\} ,
\end{align}
From Eqs. \eqref{eq:prob}, \eqref{eq:feature_expand},  \eqref{eq:partition_expand},
we can now write the probability distribution function as:
\begin{align} \label{eq:prob_gaussian}
\Pr(\y|\x) &=\frac{1} { Z(\x)} \exp ( -E(\y, \x) )  \notag \\
	&= \frac{\exp \Big\{ - \y^\T \A \y + 2\bregress^\T \y - \bregress^\T \bregress \Big\}}{\frac{{(\pi)}^{\frac{n}{2}}}{{|\A|}^{\frac{1}{2}}} \exp \{{ \bregress ^\T \A^{-1} \bregress   - \bregress^\T \bregress }\}}  \notag \\
	&=\frac{ |\A|^{\half}} {{\pi^{\frac{n}{2}}}} \exp \Big\{ - \y^\T \A \y + 2\bregress^\T \y -  \bregress ^\T \A^{-1} \bregress \Big\},
\end{align}
where $\bregress=[\regress_1, \ldots, \regress_n]^{\T}$; $  | \cdot |  $ denotes the determinant of a matrix, and $\A^{-1}$ the inverse of $\A$.
Then the negative log-likelihood can be written as:
\begin{align} \label{eq:log-likelihood}
-\log\Pr(\y|\x)&= \y^\T \A \y - 2\bregress^\T \y + \bregress ^\T \A^{-1} \bregress  \\ \notag
& - \half\log(|\A|) + \frac{n}{2}\log(\pi).
\end{align}

During learning, we minimize the negative conditional log-likelihood of the training data. Adding regularization to $\btheta$, $\bbeta$, we then arrive at the final optimization:
\begin{align} \label{eq:ccrf_final}
\min_{\btheta, \bbeta \geq \zeros} &\frac{\lambda_1}{2} \fnorm \btheta + \frac{\lambda_2}{2} \fnorm \bbeta  \\ \notag
&-\sum_{i=1}^N \log\Pr(\y^{(i)} | \x^{(i)}; \btheta, \bbeta),
\end{align}
where $\x^{(i)}$, $\y^{(i)}$ denote the $i$-th training image and the corresponding depth map; $N$ is the number of training images; $\lambda_1$ and $\lambda_2$ are weight decay parameters.

\begin{figure*}[!t] \center
	\includegraphics[width=0.8\textwidth]{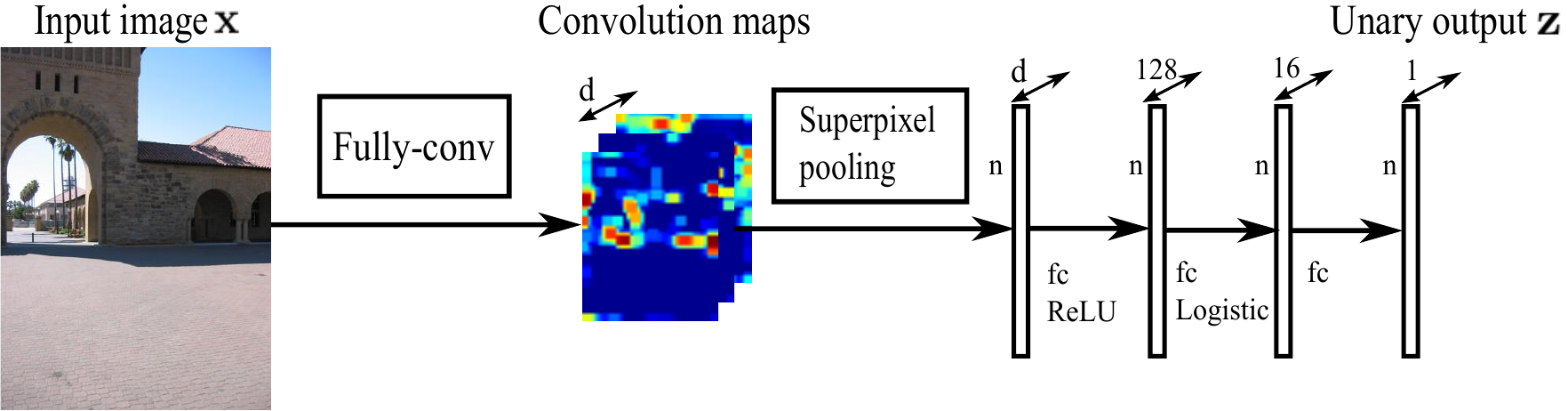}
\caption{An overview of the unary part of the \dcnfsp model.
For the unary part, the input image is fed into a fully-convolutional network to produce convolution maps
($d$ is the number of filters of the last fully-convolutional layer).
The obtained convolution maps, together with the superpixel segmentation over the original input image, are fed to a superpixel pooling layer.
The outputs are $n \times 1$ $d$ dimensional feature vectors for each of the $n$ superpixels, which are then followed by $3$ fully-connected layers to produce the unary output $\bregress$.
The pairwise part are omitted here since we use the same network architecture as in the \dcnf model (Fig. \ref{fig:dcnf}).
The unary output $\bregress$ and the pairwise output $\bpws$ are used as input to the \crf loss layer, which minimizes the negative log-likelihood (See Sec. \ref{sec:dcnfsp} for details) .
} \label{fig:dcnfsp}
\end{figure*}

\begin{figure*}[!t] \center
	\includegraphics[width=0.9\textwidth]{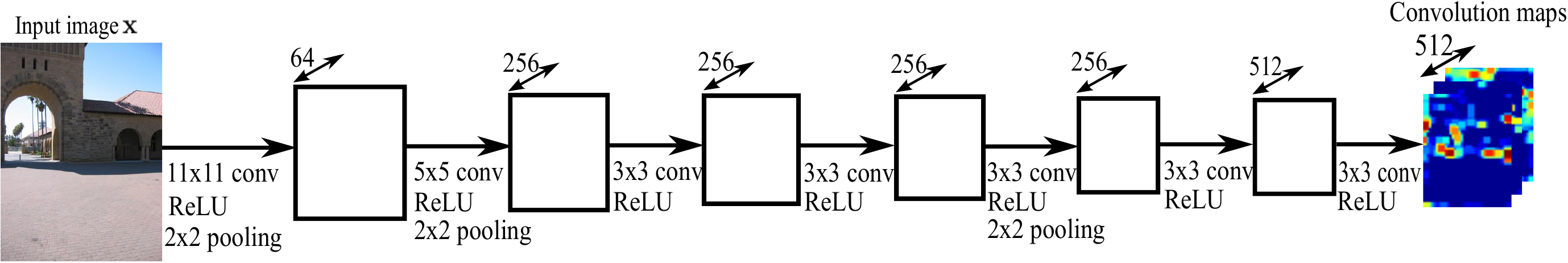}
\caption{The fully convolutional network architecture used in Fig. \ref{fig:dcnfsp}. The network takes input images of arbitrary size and output convolution maps.}  \label{fig:dcnfsp_fcnn}
\end{figure*}

\subsubsection{Optimization}
We use stochastic gradient descent (SGD) based back propagation to solve the optimization problem in Eq.  \eqref{eq:ccrf_final} for learning all parameters of the whole network.
We project the solutions to the feasible set when the bounded constraints $\beta_k \geq 0$ is violated.
In the following, we calculate the partial derivatives of $-\log\Pr(\y|\x)$ with respect to the network parameters $\btheta$ and $\bbeta$.

For the unary part,
here we calculate the partial derivatives of $-\log\Pr(\y|\x)$ with respect to $\theta_l$ (one element of $\btheta$). Recall that $\A = \I+ \D - \bpws$ (Eq. \eqref{eq:def_Amat}); $\A^{\T}=\A$; $(\A^{-1})^{\T}=\A^{-1}$; $|\A^{-1}|=\frac{1}{|\A|}$, we have:
\begin{align}  \label{eq:derive_z}
\;\;\;\;\frac{\partial \{ -\log\Pr(\y|\x)\} }{\partial \theta_l}
&= \frac{\partial \{-2\bregress^\T \y + \bregress ^\T \A^{-1} \bregress\}}{\partial \theta_l}  \\
&= 2 (\A^{-1} \bregress - \y)^{\T} \frac{\partial \bregress}{\partial \theta_l} .
\end{align}
For the pairwise part, we calculate the  partial derivatives of $-\log\Pr(\y|\x)$ with respect to $\beta_k$ as:
\begin{align}  \label{eq:derive_beta}
&\;\;\;\;\frac{\partial \{ -\log\Pr(\y|\x)\} }{\partial \beta_k}   \notag \\
&= \frac{\partial \{\y^\T \A \y + \bregress ^\T \A^{-1} \bregress - \half\log(|\A|)\}}{\partial \beta_k}   \notag \\
&=\y^\T\frac{\partial  \A }{\partial \beta_k}\y - \bregress^\T \A^{-1} \frac{\partial \A }{\partial \beta_k} \A^{-1} \bregress  - \half \frac{1}{|\A|} \frac{\partial \{|\A|\} }{\partial \beta_k}, \notag \\
&=\y^\T \frac{\partial  \A }{\partial \beta_k} \y - \bregress^\T \A^{-1} \frac{\partial \A }{\partial \beta_k} \A^{-1} \bregress  - \half \trace \Big ( \A^{-1}  \frac{\partial \A} {\partial \beta_k} \Big ).
\end{align}
where $\trace(\cdot)$ denotes the trace of a matrix.
We here introduce matrix $\J$ to denote $\frac{\partial  \A }{\partial \beta_k}$. Each element of $\J$ is:
\begin{align} \label{eq:def_J}
J_{pq} &= \frac{\partial A_{pq}}{\partial \beta_k}  \notag \\
&= \frac{\partial \{ D_{pq} - R_{pq} \} }{\partial \beta_k}  \notag \\
&= - \frac{\partial  \pws_{pq} }{\partial \beta_k} + \delta(p=q) \sum_q \frac{\partial  \pws_{pq} }{\partial \beta_k},
\end{align}
where $\delta(\cdot)$ is the indicator function, which equals 1 if $p=q$ is true and 0 otherwise.
According to Eq. \eqref{eq:derive_beta} and the definition of  $\J$ in \eqref{eq:def_J}, we can now write the partial derivative of $-\log\Pr(\y|\x)$ with respect to $\beta_k$  as:
\begin{equation}  \label{eq:derive_beta_final}
\frac{\partial \{ -\log\Pr(\y|\x)\} }{\partial \beta_k} =  \y^{\T} \J \y - \bregress^{\T} \A^{-1} \J \A^{-1}\bregress - \half \trace \Big( \A^{-1} \J \Big).
\end{equation}
From Eqs. \eqref{eq:derive_beta_final}, \eqref{eq:def_J}, we can see that our framework is general and more complicated networks for the pairwise part can be seamlessly incorporated.
Here, in our case, with the definition of $\pws_{pq}$ in Eq. \eqref{eq:def_R}, we have $\frac{\partial  \pws_{pq} }{\partial \beta_k}=S_{pq}^{(k)}$.

\subsubsection{Depth prediction}
Predicting the depths of a new image is to solve the MAP inference in Eq. \eqref{eq:inference}, which writes as:
\begin{align} \label{eq:inf_solution1}
\y^{\star}
&=\argmax_{\y} \log \Pr(\y|\x) \notag   \\
&= \argmax_{\y} -\y^\T \A \y + 2\bregress^\T \y.
\end{align}
With the definition of $\A$ in Eq. \eqref{eq:def_Amat}, $\A$ is symmetric.
Then by setting the partial derivative of $-\y^\T \A \y + 2\bregress^\T \y$ with respect to $\y$ to $\zeros$, we have
\begin{align} \label{eq:inf_deriv}
&\frac{\partial \{ -\y^\T \A \y + 2\bregress^\T \y \} } {\partial \y} =\zeros  \notag  \\
\Rightarrow \;\;&-(\A + \A^{\T} ) \y +2 \bregress=\zeros  \notag  \\
\Rightarrow \;\;&\y = \A^{-1} \bregress.
\end{align}
It shows that closed-form solutions exist for the MAP inference in Eq. \eqref{eq:inf_solution1}:
\begin{align} \label{eq:inf_solution}
\y^{\star}&= \A^{-1} \bregress.
\end{align}
If we discard the pairwise terms, namely $\pws_{pq}=0$, then Eq. \eqref{eq:inf_solution} degenerates to $\y^{\star}=\bregress$, which is a plain CNN regression model (we will report the results of this method as a baseline in the experiment).
Note that we do not need to explicitly compute the matrix inverse $ \A^{-1}$ which can be expensive (cubic in the number of nodes).
Instead we can obtain the value of $  \A^{-1} \bf z $ by solving a linear system.

\subsection{Speeding up training using fully convolutional networks and superpixel pooling}
\label{sec:dcnfsp}
Thus far, we have presented our \dcnf model for depth estimations based on image superpixels.
From Fig. \ref{fig:dcnf}, we can see that for constructing the unary potentials, we are essentially performing patchwise convolutions (similar operations are performed in the R-CNN \cite{RCNN14}).
A major concern of the proposed method is its computational efficiency and memory consumption, since we need to perform convolutions over hundreds or even thousands (number of superpixels) of image patches for a single input image.
Many of those convolutions are redundant due to significant image patch overlaps.

Naturally, a promising direction for reducing the computation burden is to perform convolutions over the entire image once, and then obtain convolutional features for each superpixel.
However, to find the convolutional features of the image superpixels from the obtained convolution maps, one needs to establish associations between these two.
Therefore, we here propose an improved model, which we term as \dcnfsp (\dcnf with Fully Convolutional networks and Superpixel Pooling), based on fully convolutional networks and a novel superpixel pooling method,  to address this issue.
As we will show in Sec.~\ref{sec:exp_time}, this new model significantly speeds up the training and  prediction while producing almost the same prediction accuracy.
Most importantly, with this more efficient model, we are able to design deeper networks to achieve better performance.

\begin{figure*}
  \centering
	\includegraphics[width=0.9\textwidth]{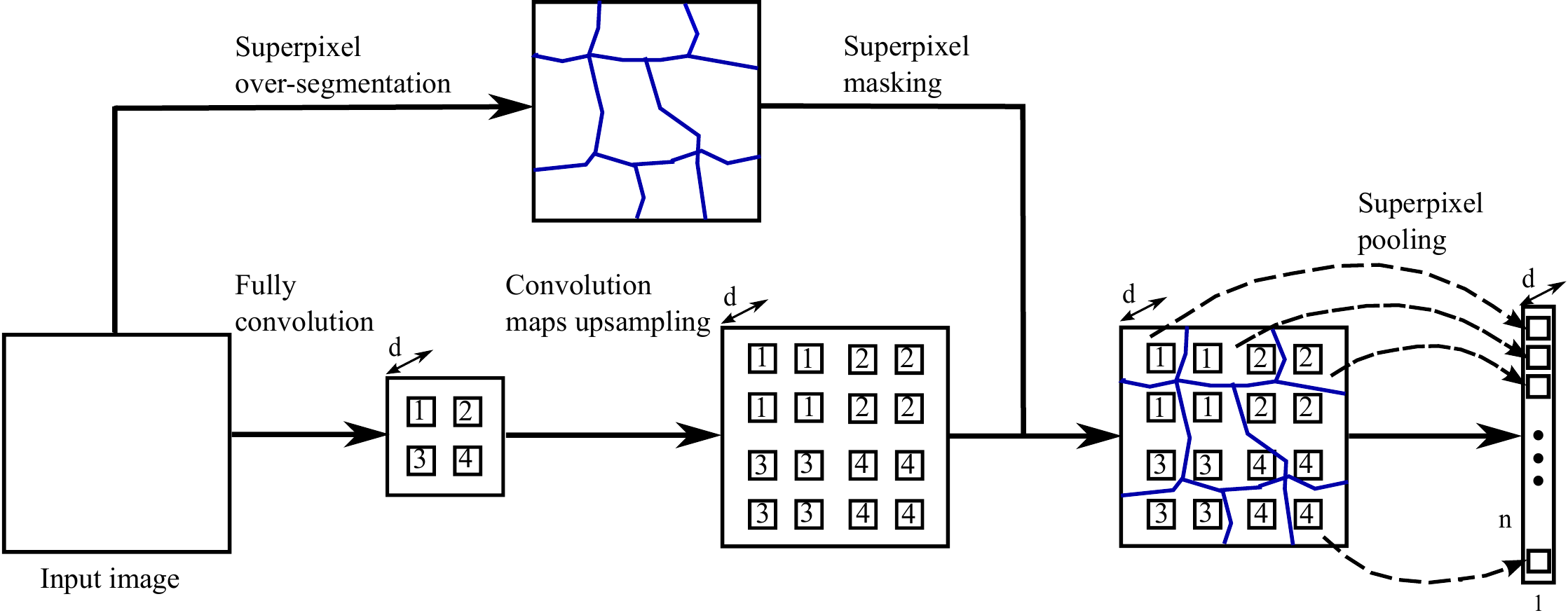}
\caption{An illustration of the superpixel pooling method,
  which mainly consists of convolution maps upsampling and superpixel pooling. The convolution maps are upsampled to the original image size by nearest neighbor interpolations,
over which the superpixel masking is applied.
Then average pooling is performed within each superpixel region,
to produce the $n$ convolution features.
$n$ is the number of superpixels in the image. $d$ is the number of channels of the convolution maps.}  \label{fig:sp_pooling}
\end{figure*}

\subsubsection{\dcnfsp overview}
In comparison to the \dcnf model, our new \dcnfsp model mainly improves the unary part while  keeping the same pairwise network architecture as in Fig. \ref{fig:dcnf}.
We show the model architecture of the unary part in Fig. \ref{fig:dcnfsp}.
Specifically, the input image is fed to a fully convolutional network (introduced in the sequel).
The outputs are convolutional feature maps of size $h \times w \times d$ ($d$ is the dimension of the obtained convolutional feature vector, \ie, the number of channels of the last convolutional layer).
The size of the convolution maps $h \times w$ are typically smaller than the input image size.
Each convolutional feature vector in the output convolution maps corresponds to a patch in the input image.
We propose a novel superpixel pooling method (described in the following) to associate these outputs back to the superpixels in the input image.
Specifically, the convolutional feature maps are used as inputs to a superpixel pooling layer
to obtain $n$ superpixel feature vectors with $d$ dimensions ($n$ is the number of superpixels).
The $n$ superpixel feature vectors are then fed to $3$ fully-connected layers to produce the unary output $\bregress$.
We use the same pairwise network architecture as depicted in Fig. \ref{fig:dcnf}, which we do not show here.
With the unary output $\bregress$ and the pairwise output $\bpws$, we construct potential
functions according to Eqs. \eqref{eq:unary} and \eqref{eq:pairwise} and optimize the negative log-likelihood.

Compared to the original proposed \dcnf method, this improved \dcnfsp model only needs to perform convolutions  over
the entire image once, rather than hundreds of superpixel image patches.
This significantly reduces the computation and GPU memory burden, bringing around $10$ times
training speedup while producing almost the same prediction accuracy,
as we demonstrate later in Sec. \ref{sec:exp_time}.
We next introduce the fully convolutional networks and the superpixel pooling method in detail.

\subsubsection{Fully convolutional networks}
Typical \cnn models, including AlexNet \cite{AlexNet12}, vggNet \cite{vgg_bmvc14,vgg16_arxiv}, \etc,  are composed of convolution layers and fully connected layers.
They usually take standard sized images as inputs, \eg, $224 \times 224$ pixels, to produce nonspatial outputs.
In contrast, a fully convolutional network can take as inputs arbitrarily sized images, and outputs convolutional spatial maps.
It has therefore been actively studied for dense prediction problems \cite{Long_arxiv14,Cogswell_arxiv14,Dong_eccv14,dcnn_nips14} very recently.
We here exploit this new development trend in \cnn to speedup the patch-wise convolutions in the \dcnf model.
We illustrate the fully convolutional network architecture that we use in Fig.~\ref{fig:dcnfsp_fcnn}.
As shown, the network is composed of $7$ convolution layers, with the first $5$ layers transferred from the AlexNet \cite{AlexNet12}.
We then add $2$ more convolution layers with $3 \times 3$ filter size and $512$ channels each.
The network takes as input images of arbitrary size and outputs convolution maps of channel $d=512$.
Note that we can design deeper networks here for pursuing better performance.
We will demonstrate in Sec.~\ref{sec:exp_stateComp} the benefits of using deeper models.

\subsubsection{Superpixel pooling}
After the input images go through the fully convolutional networks, we acquire convolution maps.
To obtain superpixel features, we need to associate these convolutional feature maps back to the image superpixels.
Thus we here propose a novel superpixel pooling method.
An illustration of this method is shown in Fig. \ref{fig:sp_pooling}, which mainly consists of the convolution maps upsampling and superpixel average pooling.
In general, this superpixel pooling layer takes the convolution maps as input and outputs superpixel features.
Specifically, the obtained convolution maps are first upsampled to the original image size by
nearest neighbor interpolation.
Note that other interpolation methods such as linear interpolation may be applicable too. However,
as discussed below, nearest neighbor interpolation makes the implementation much easier and computation faster.

Then the superpixel masking is applied and average pooling is performed within each superpixel region.
The outputs are pooled superpixel features, which are used for constructing the unary potentials.
In the sequel, we describe this method in detail.

In practice there is no need to explicitly upsample the convolution maps.
Instead, we count the frequencies of the convolutional feature vectors that fall into each  superpixel region.
We denote  the convolution maps as $\C \in \Real^{h \times w \times d}$, with each element being $C_{ijk}$ ($i=1, \ldots, h; j=1, \ldots, w; k=1, \ldots, d$).
We represent the $t$-th superpixel feature as a $d$-dimensional column vector $\h_t$ ($t=1, \ldots, n$), with elements $h_{tk}$ ($k=1, \ldots, d$).
$\W_t \in \Real^{h \times w}$  is a frequency weighting matrix associated to the $t$-th superpixel, with elements being $W_{ijt}$.
$W_{ijt}$ represents the weight of $(i, j)$-th feature vector in the convolution maps that associated to the $t$-th superpixel.
To calculate $W_{ijt}$, we simply count the occurrences of the $(i,j)$-th convolutional feature vector that appear in  the $t$-th superpixel region, and do $L_1$ normalization  for each $\W_t$.
By constructing this frequency matrix $\W_t$, we avoid the explicit upsampling operation.
Then the superpixel pooling can be represented as:
\begin{align} \label{eq:sppooling}
h_{tk} = \sum_{(i,j) \in {\cal R}_{t}} W_{ijt} \cdot C_{ijk},
\end{align}
where $(i, j) \in {\cal R}_t$ denotes the $(i, j)$-th convolutional feature vector  in the convolution maps being associated to the $t$-th superpixel.

During the network forward pass, the superpixel pooling layer performs a linear transformation in Eq. \ref{eq:sppooling} to output $\h_t$ from the input $\C$.
For the network backward, the gradients can be easily calculated since we have:
\begin{align} \label{eq:derivative_sppooling}
\frac{\partial h_{tk}}{\partial C_{ijk}}= \left\{
			\begin{array}{ll}
				W_{ijt}  &\mbox{if } (i,j) \in {\cal R}_t \\
				0 &\mbox{otherwise}.
			\end{array}
		\right.
\end{align}
Thus far, we have successfully established the associations between the convolutional feature maps and the image superpixels.
It should be noted that although simple as it is, the proposed superpixel pooling method
jointly exploits the benefits of fully convolutional networks and superpixels.
It provides an efficient yet equally effective approach to the patchwise convolutions used in the \dcnf model,
as we demonstrate in Sec. \ref{sec:exp_time}.

\subsection{Implementation details}
We implement the network training based
on the  CNN toolbox: VLFeat MatConvNet\footnote{VLFeat MatConvNet: http://www.vlfeat.org/matconvnet/}
with our own modifications.
Training is done on a standard desktop with an NVIDIA GTX 780 GPU with 6GB memory.
The DCNF model, with the network design in Fig. \ref{fig:dcnf_unary}, has approximately 40 million parameters.
The DCNF-FCSP model, with the network design in Fig. \ref{fig:dcnfsp_fcnn}, has around 5.8 million parameters.
With the very deep network architecture, the DCNF-FCSP model has around 20 million parameters.
The huge amounts of parameters for the DCNF model mainly comes from the 4096 fully connected layer in Fig. \ref{fig:dcnf_unary},  which is removed in the DCNF-FCSP model.
Next we present the implementation details of the two proposed models in the following.

{\bf \dcnf}
We initialize the first 6 layers of the unary part in Fig. \ref{fig:dcnf_unary} using a \cnn model trained on the ImageNet from \cite{vgg_bmvc14}.
First, we do not back propagate through the previous 6 layers by keeping them fixed and train the rest of the network (we refer this process as pre-train) with the following settings:
momentum is set to $0.9$, and weight decay parameters $\lambda_1, \lambda_2$ are set to $0.0005$.
During pre-train, the learning rate is initialized at $0.0001$,
and decreased by 40\% every 20 epoches.
We then run 60 epoches to report the results of pre-train (with learning rate decreased twice).
During pre-training, it takes less than $0.1$s for one network forward pass to do depth predictions.
Then we train the whole network with the same momentum and weight decay.
Training the whole network takes around $16.5$ hours on the Make3D dataset, and around $33$ hours on the NYU v2 dataset.
For this fine-tune model, it takes $\sim 1.1$s for one network forward pass to do depth predictions.

{\bf \dcnfsp}
We initialize the first $5$ layers in Fig. \ref{fig:dcnfsp_fcnn} with the same model trained on the ImageNet from \cite{vgg_bmvc14}.
The momentum and weight decay parameters are set the same as in the \dcnf model.
We also use  the same training protocol as in the \dcnf model, \ie, first pre-train and then fine-tune the whole model.

\begin{figure*} [t]  \center
\resizebox{.9\linewidth}{!} {
\begin{tabular}{cccc}
	\includegraphics[width=0.2\textwidth, height=0.15\textwidth]{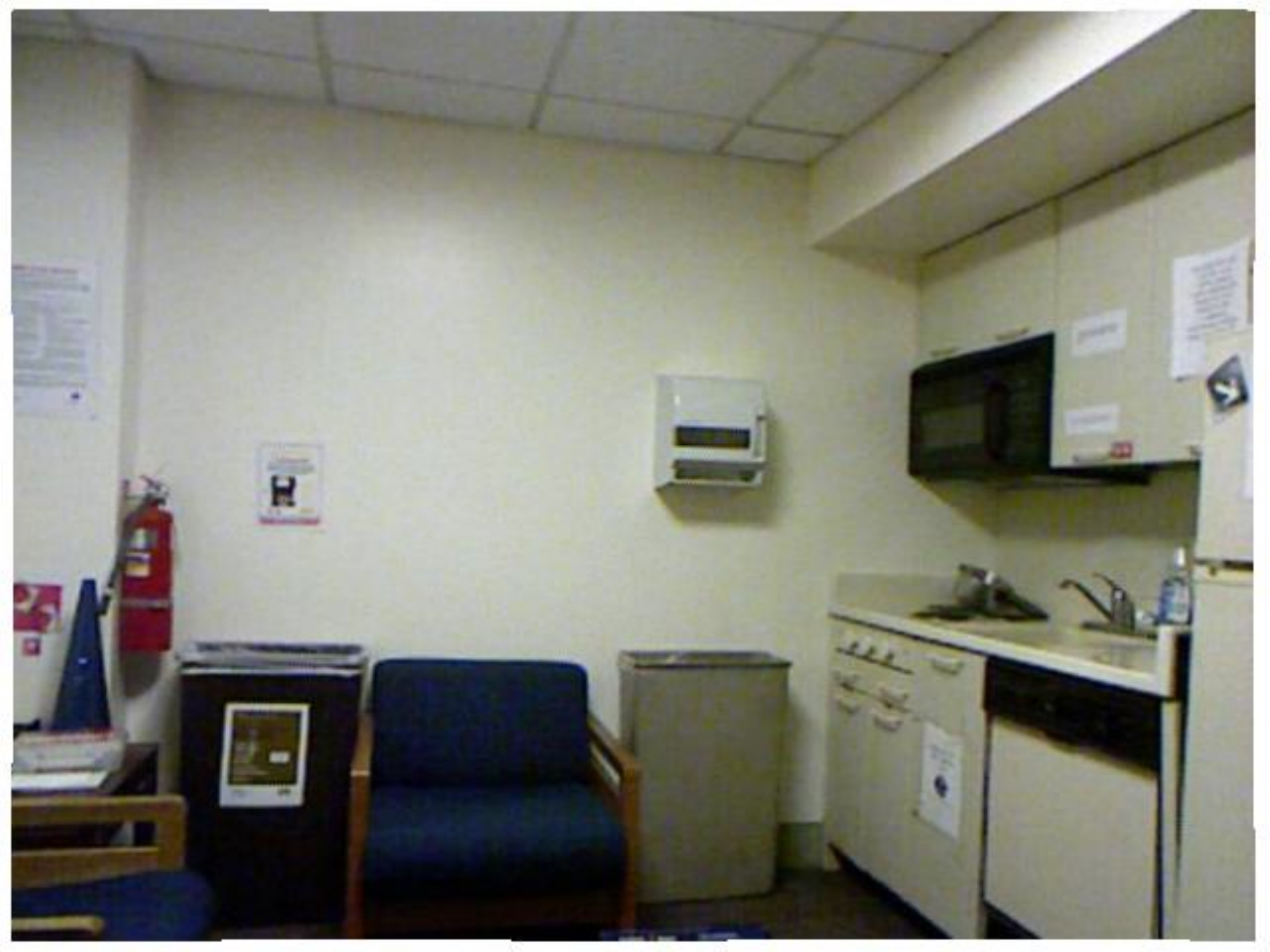}
    &\includegraphics[width=0.2\textwidth, height=0.15\textwidth]{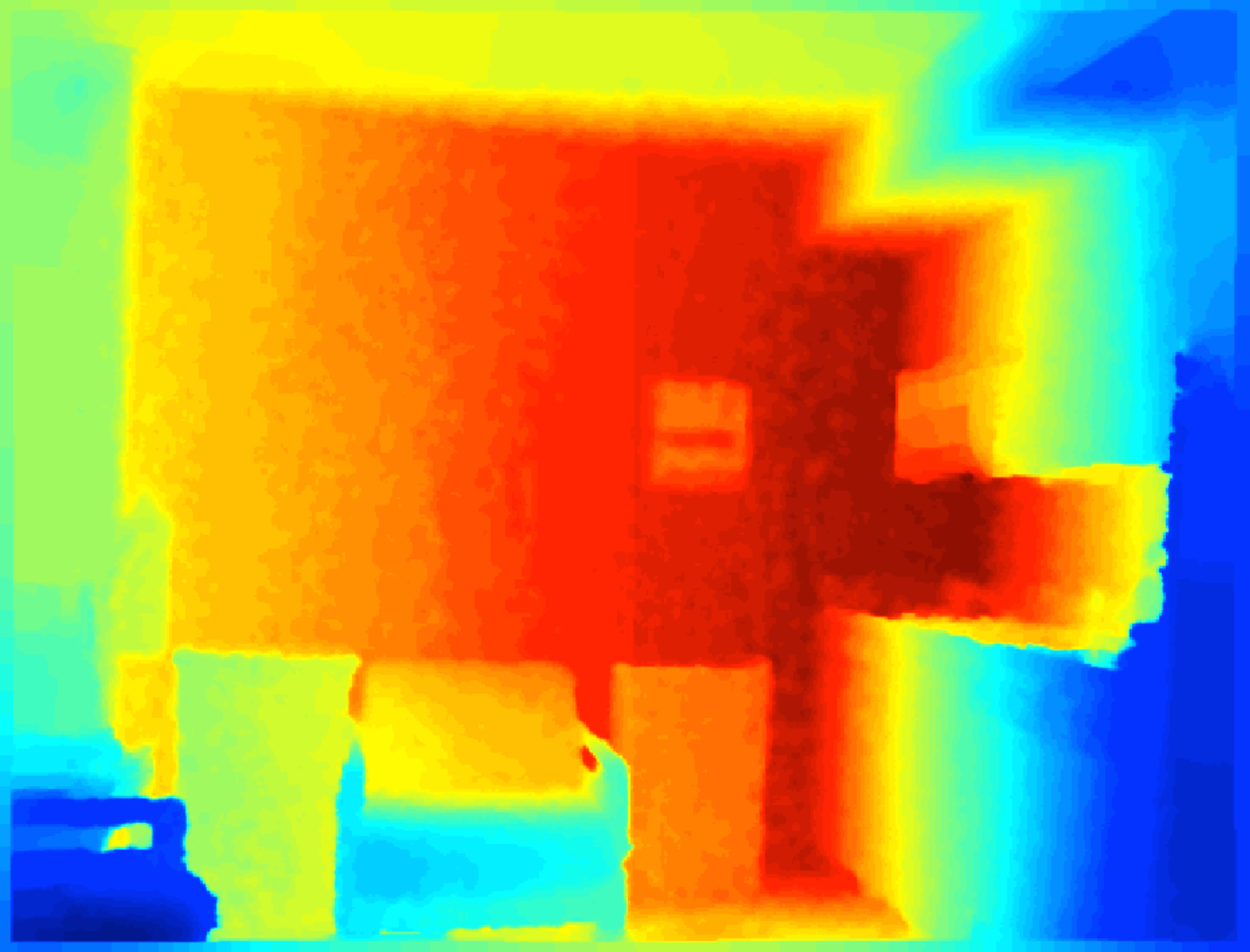}
    &\includegraphics[width=0.2\textwidth, height=0.15\textwidth]{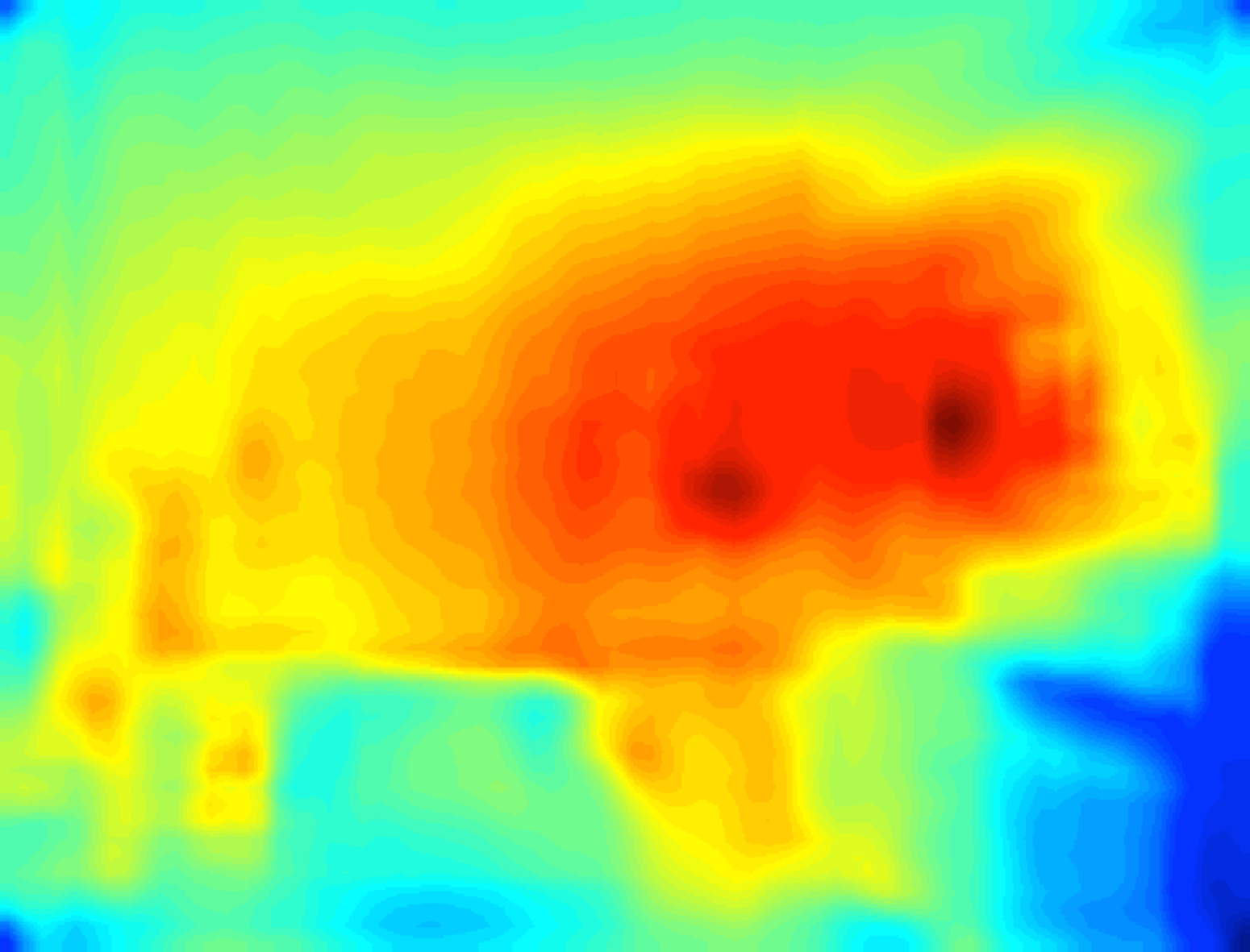}
    &\includegraphics[width=0.2\textwidth, height=0.15\textwidth]{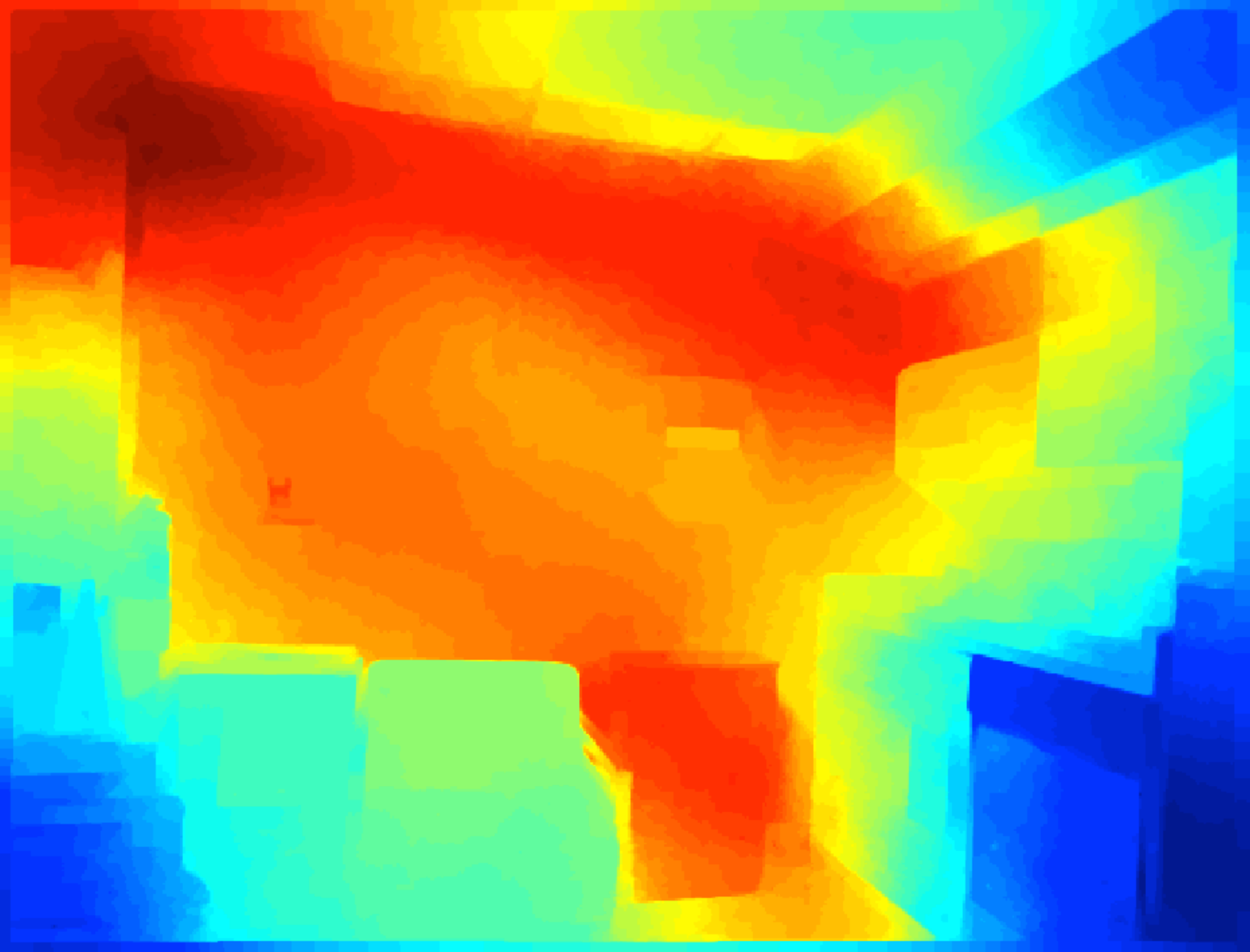}  \\

    \includegraphics[width=0.2\textwidth, height=0.15\textwidth]{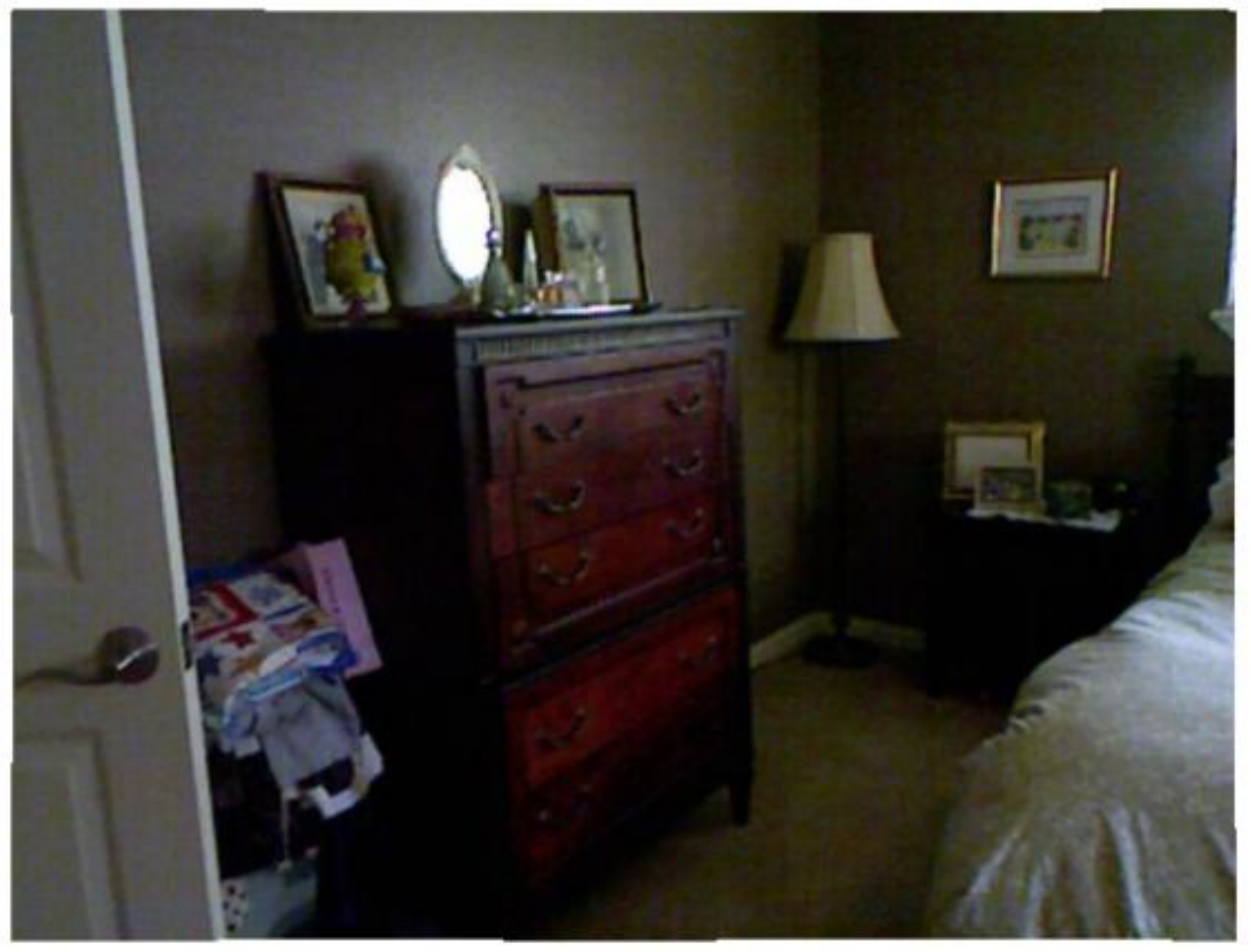}
    &\includegraphics[width=0.2\textwidth, height=0.15\textwidth]{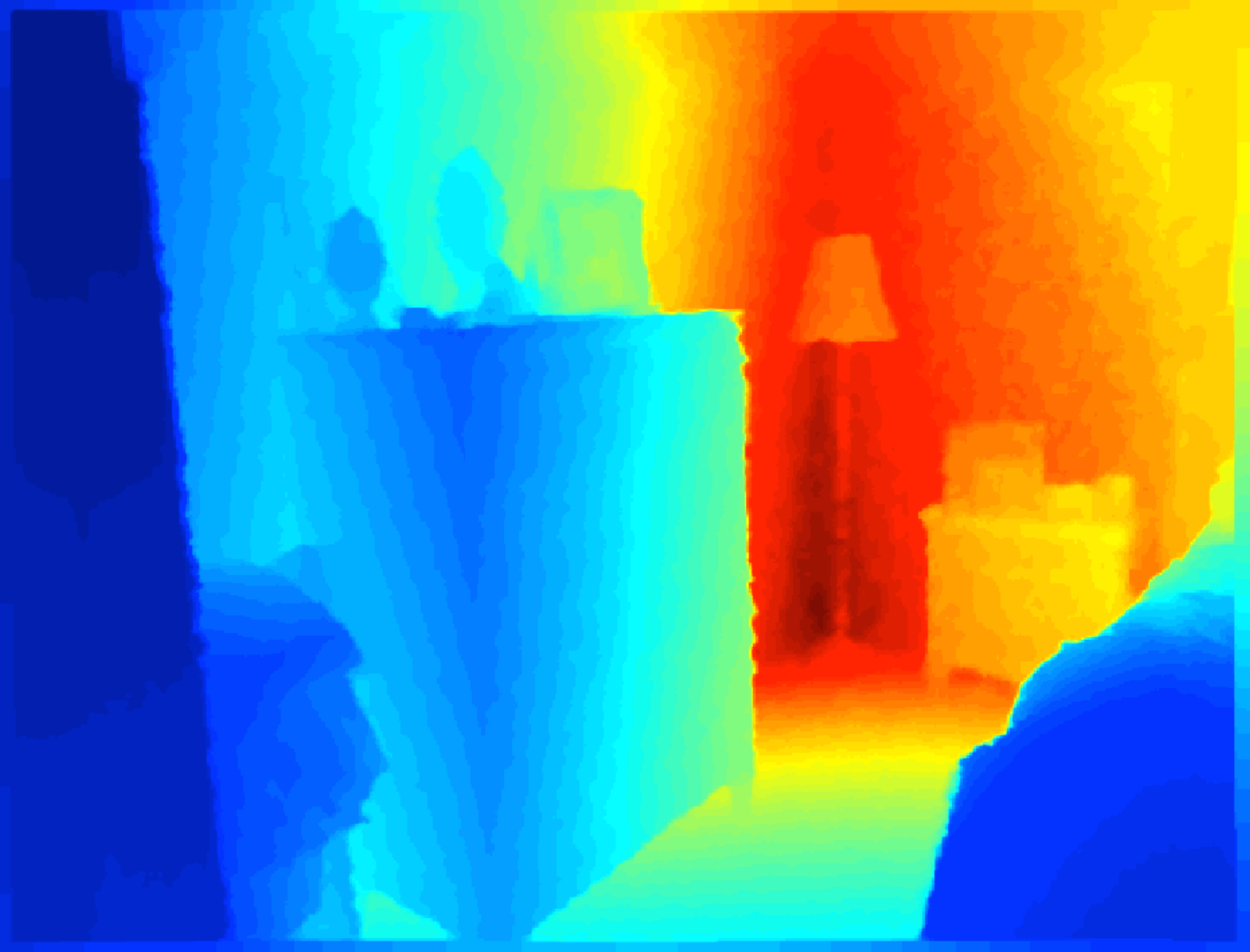}
    &\includegraphics[width=0.2\textwidth, height=0.15\textwidth]{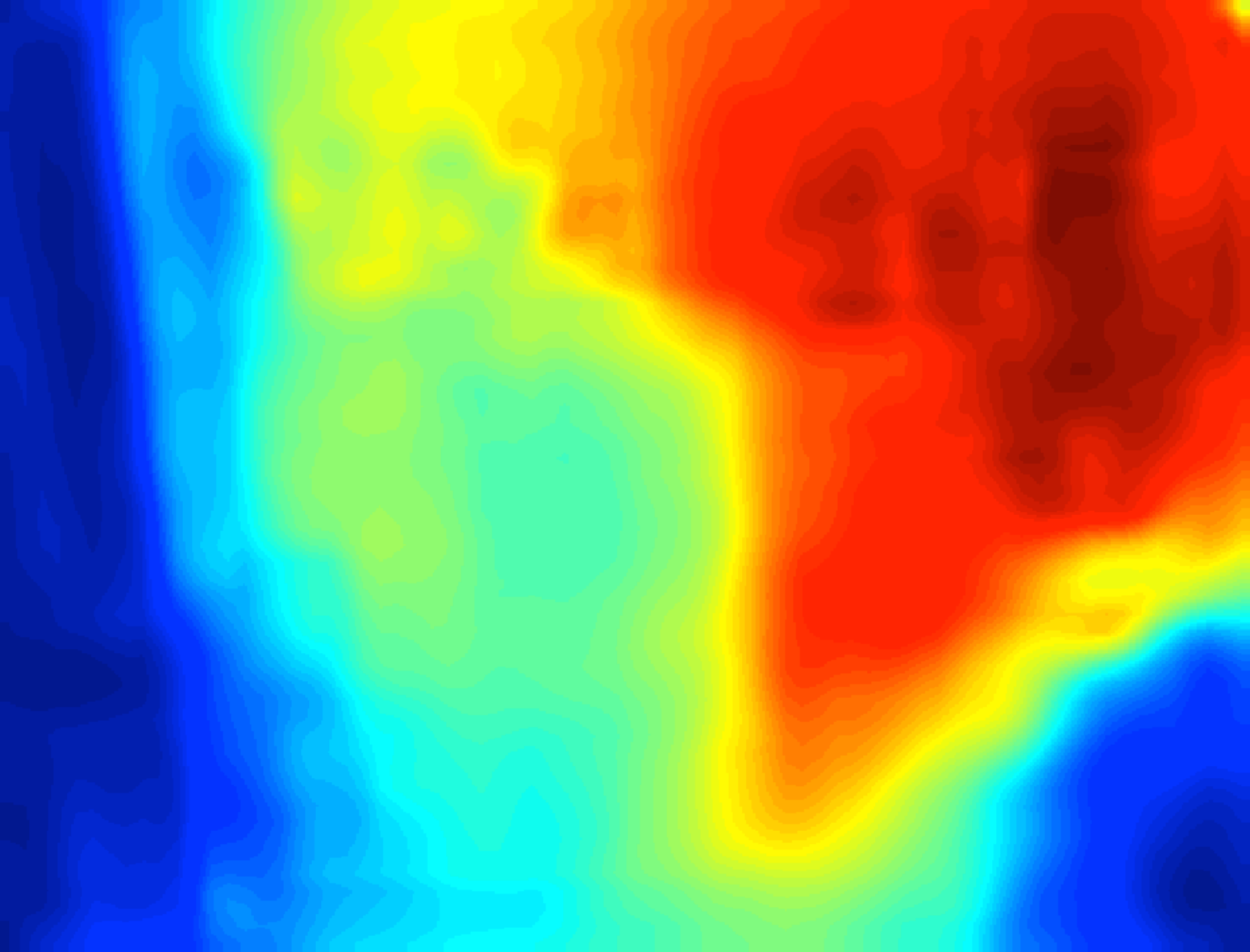}
    &\includegraphics[width=0.2\textwidth, height=0.15\textwidth]{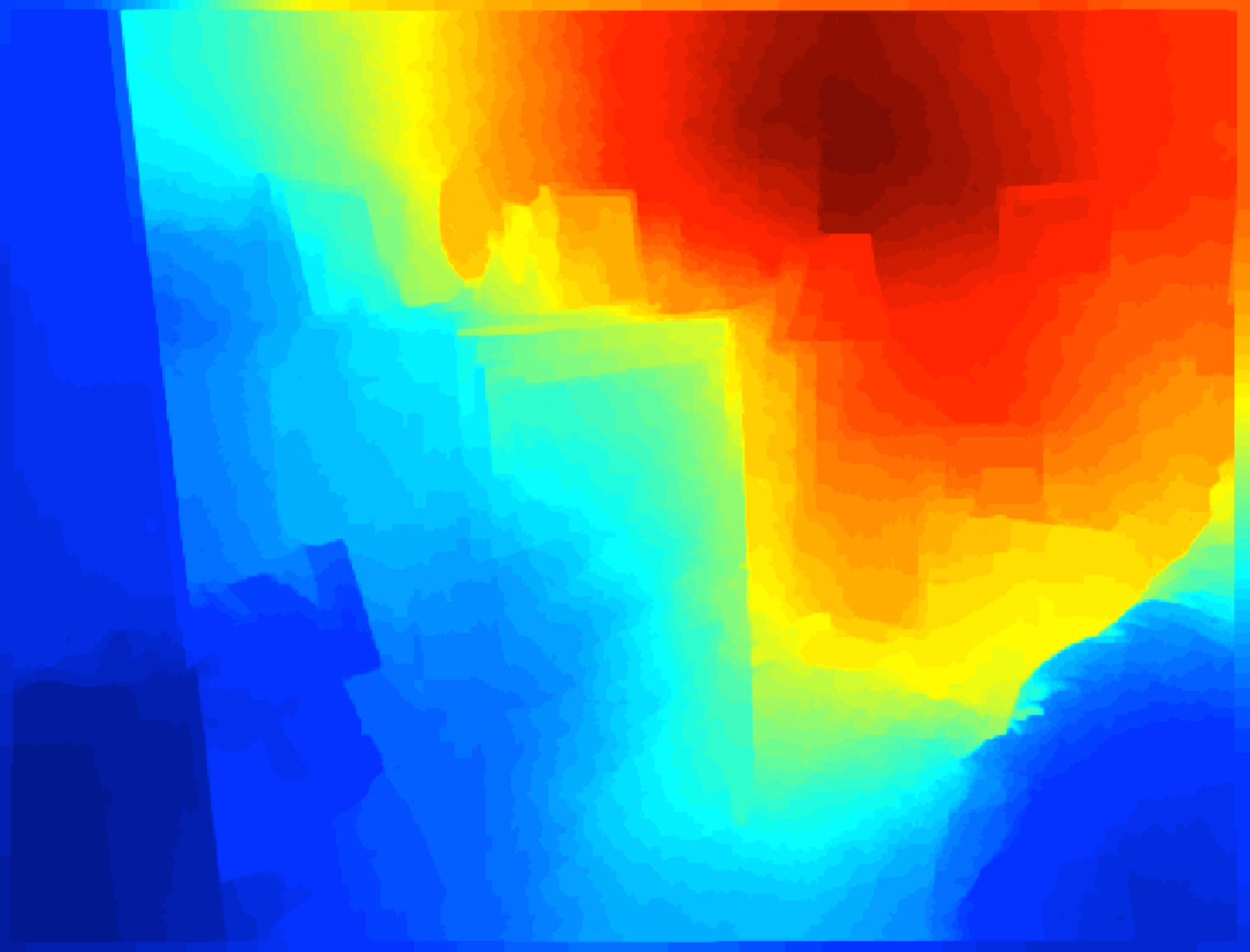}  \\

    \includegraphics[width=0.2\textwidth, height=0.15\textwidth]{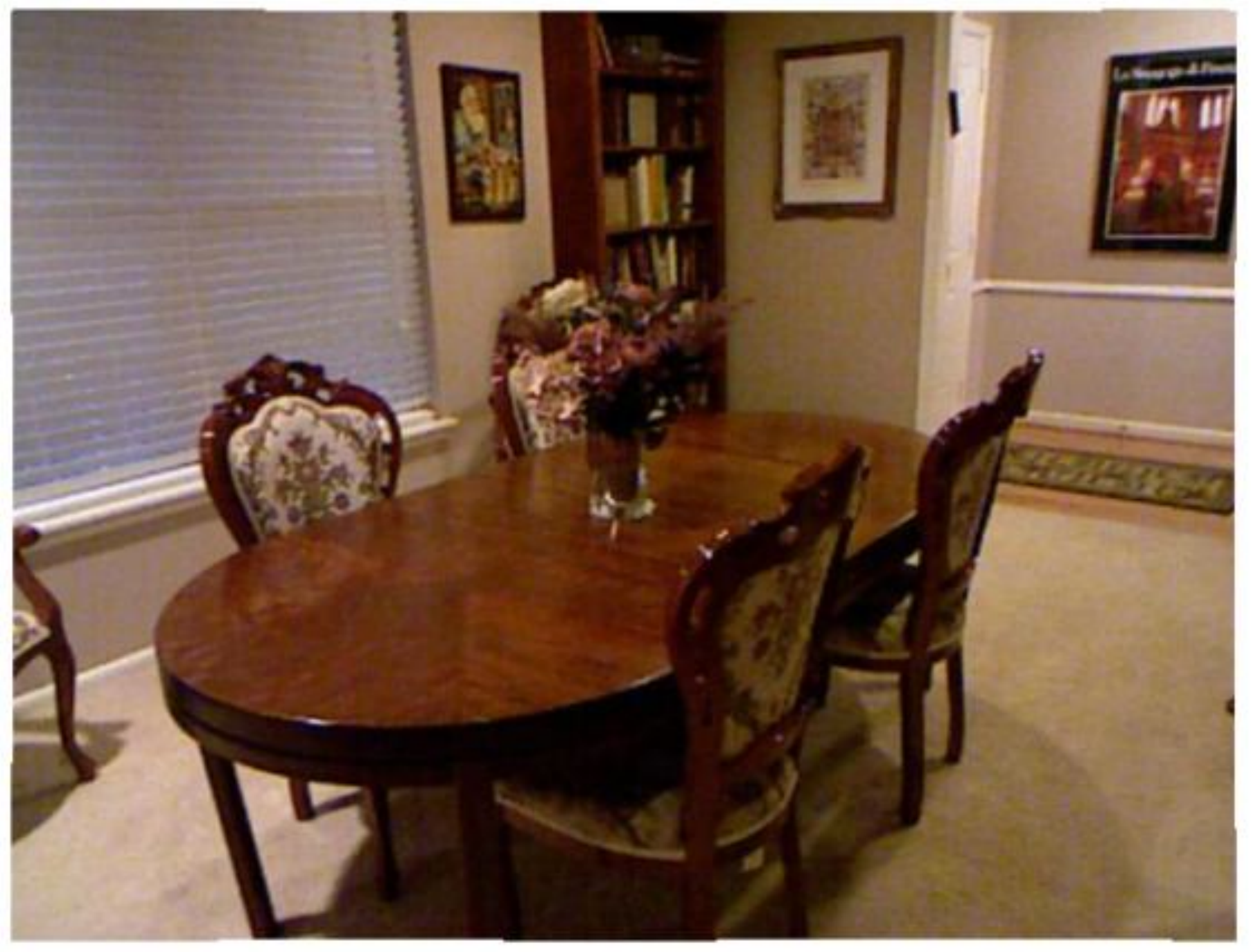}
    &\includegraphics[width=0.2\textwidth, height=0.15\textwidth]{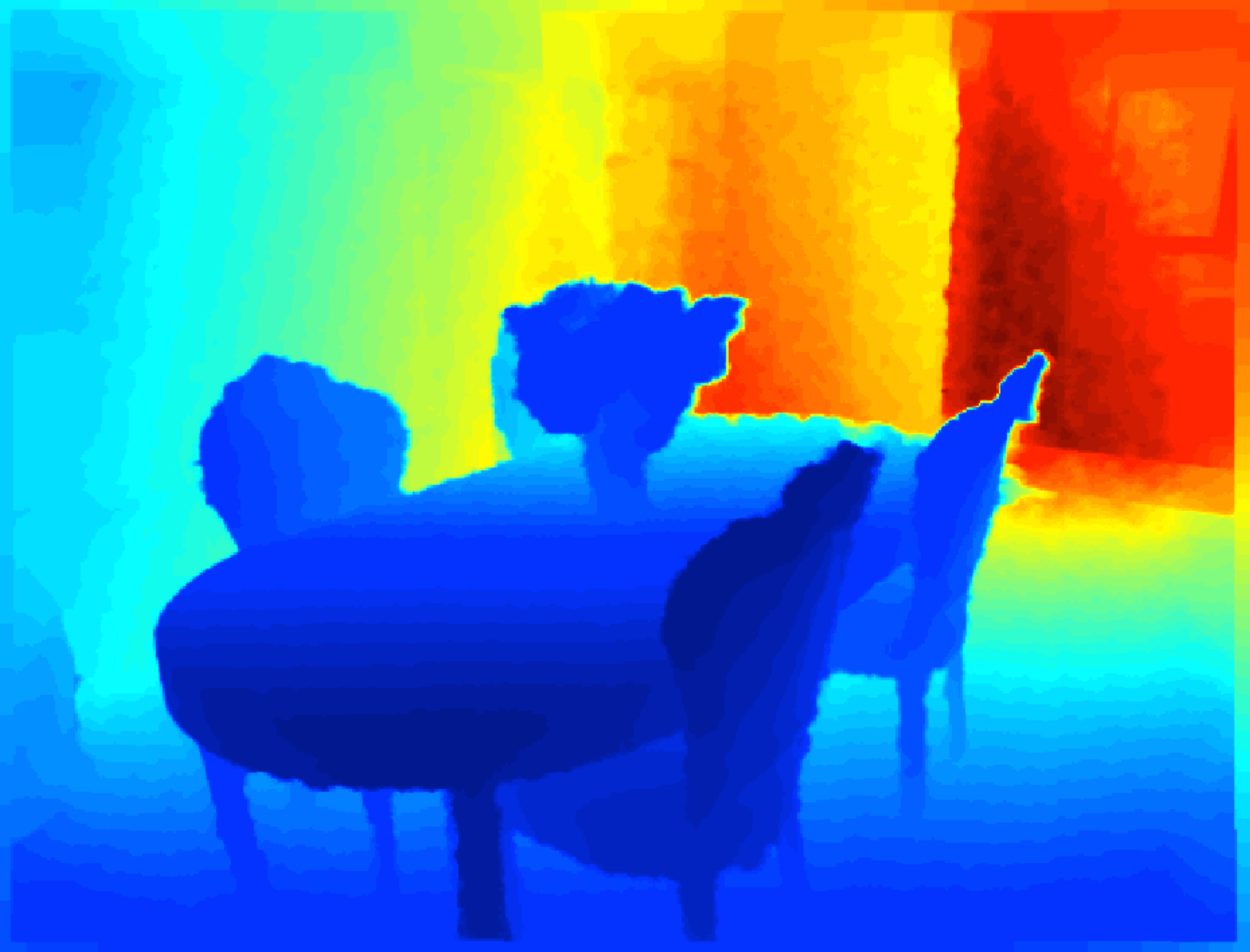}
    &\includegraphics[width=0.2\textwidth, height=0.15\textwidth]{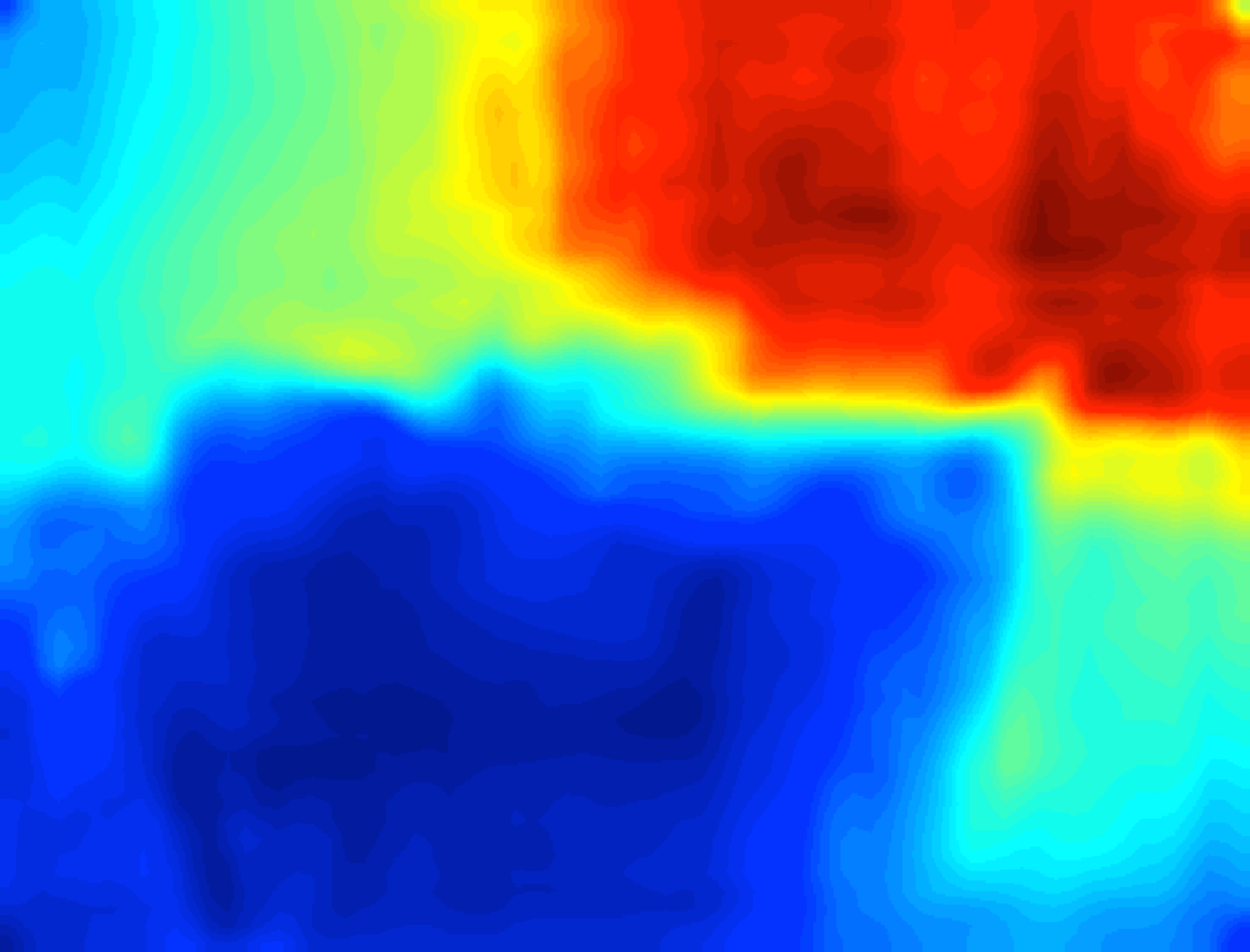}
    &\includegraphics[width=0.2\textwidth, height=0.15\textwidth]{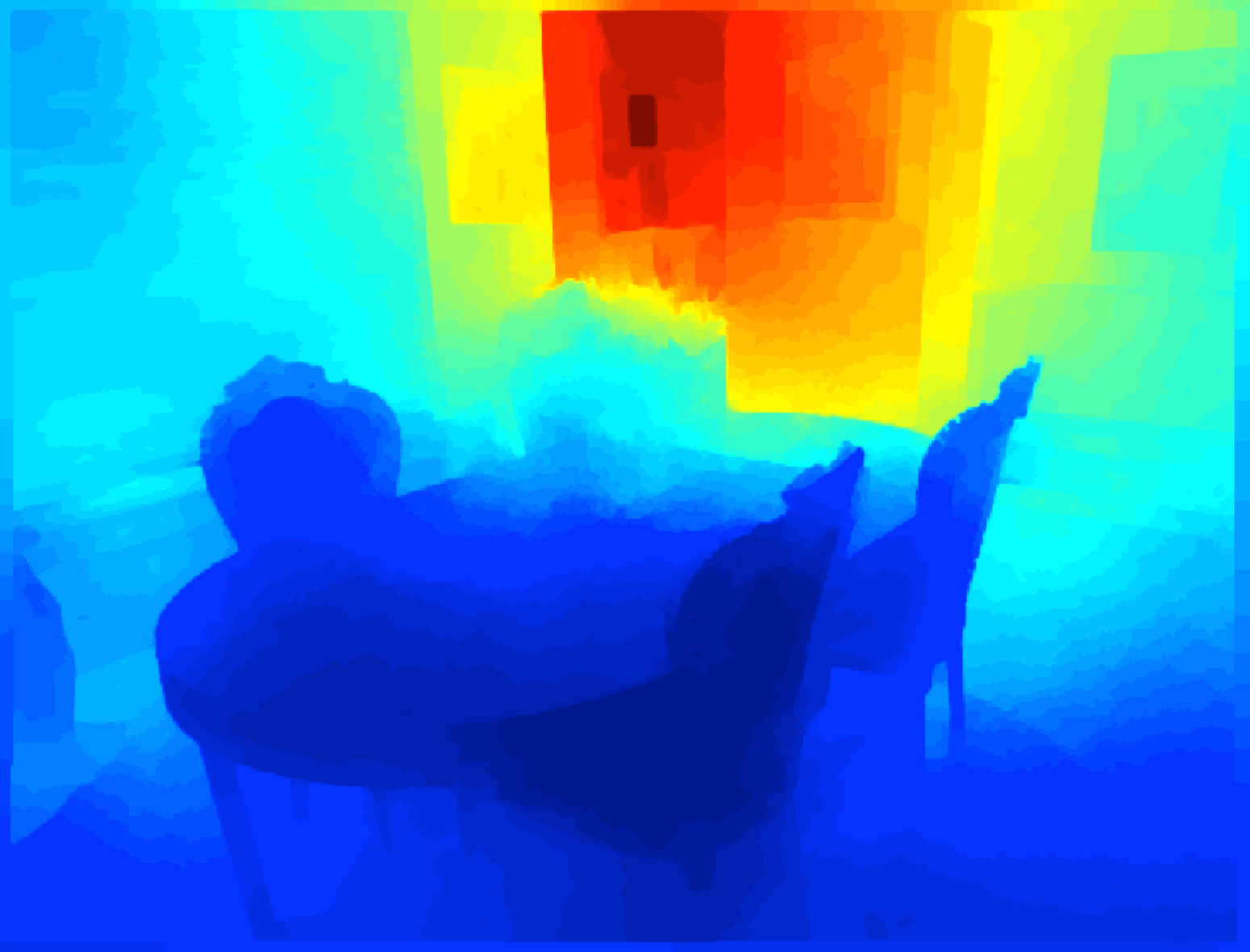}  \\

    \includegraphics[width=0.2\textwidth, height=0.15\textwidth]{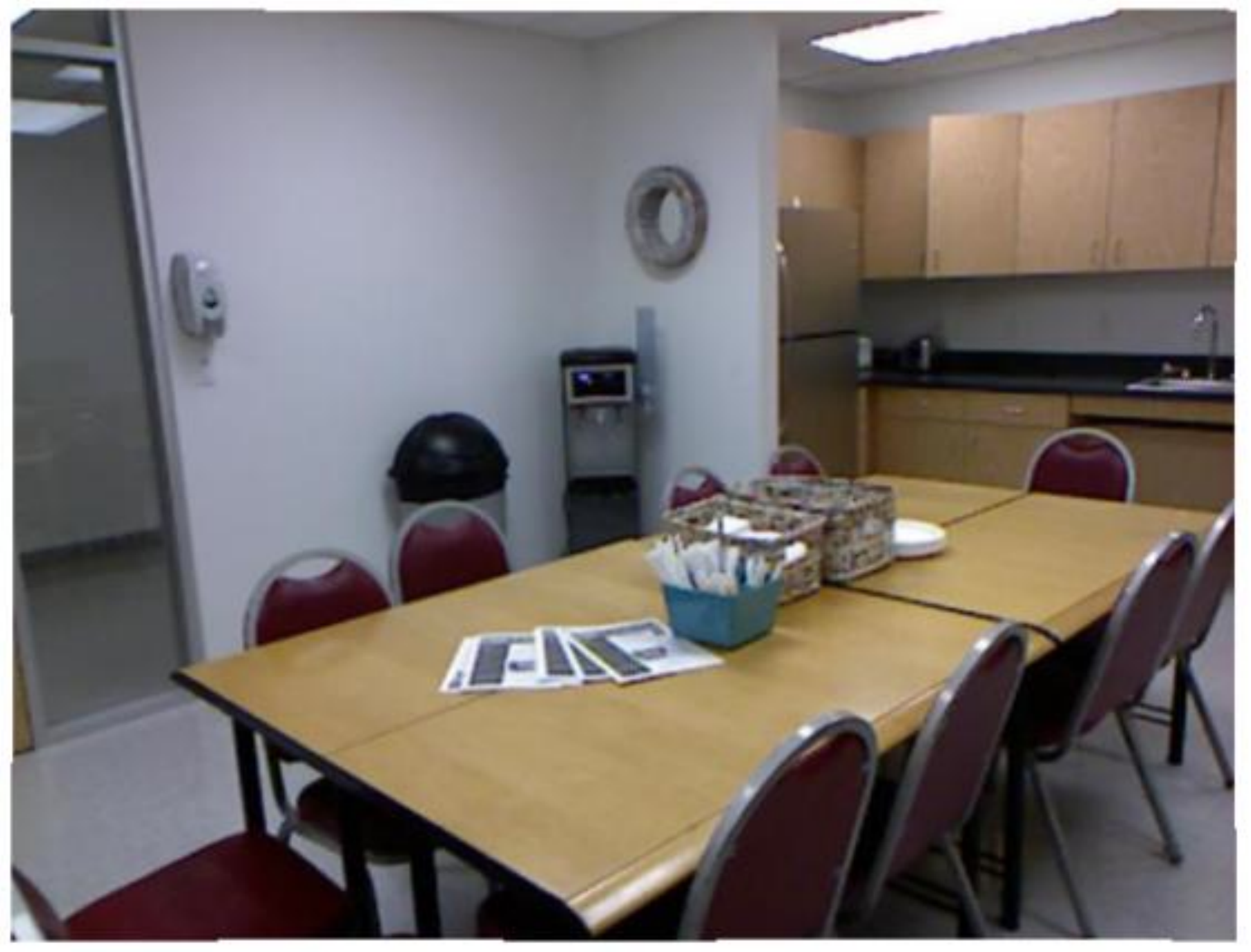}
    &\includegraphics[width=0.2\textwidth, height=0.15\textwidth]{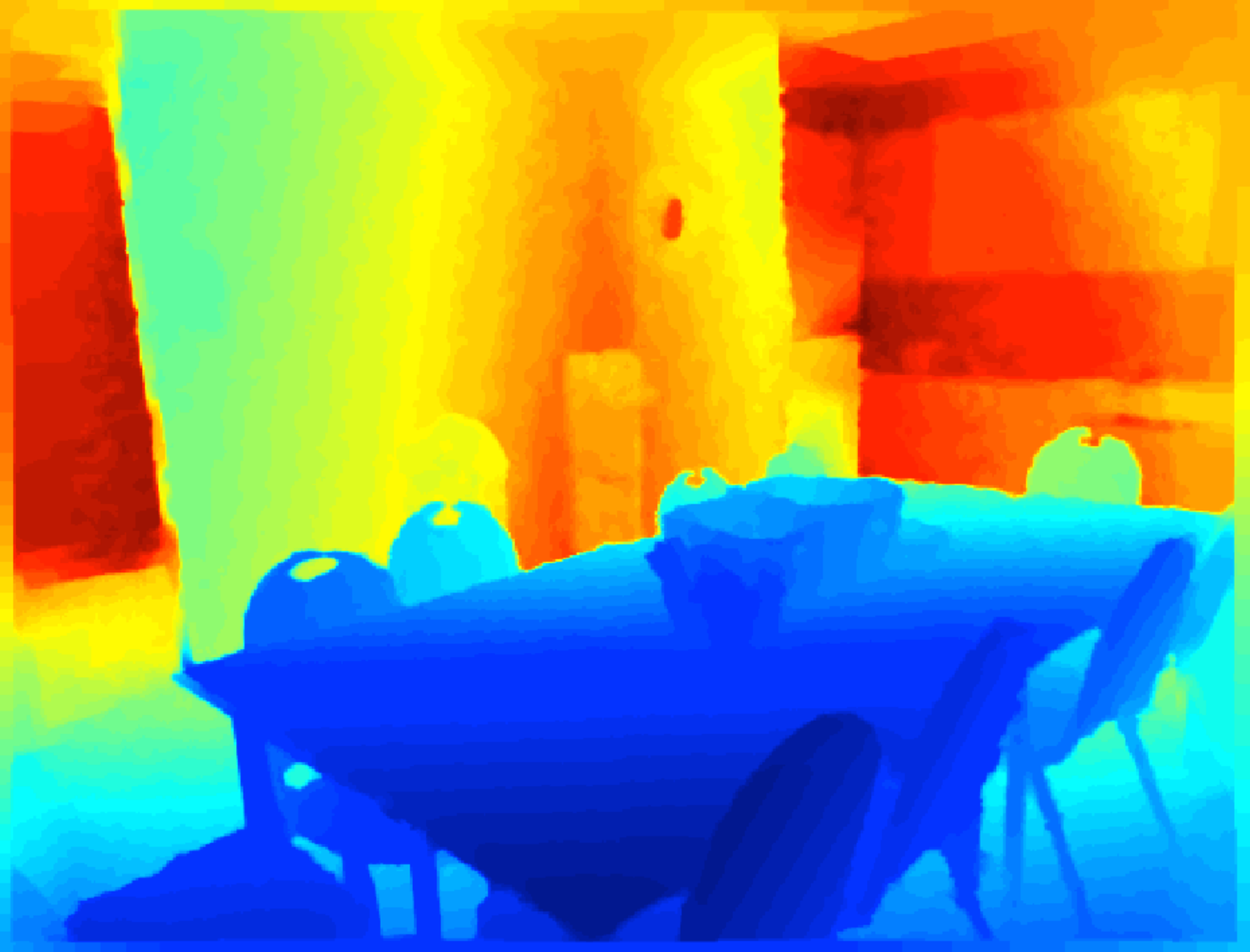}
    &\includegraphics[width=0.2\textwidth, height=0.15\textwidth]{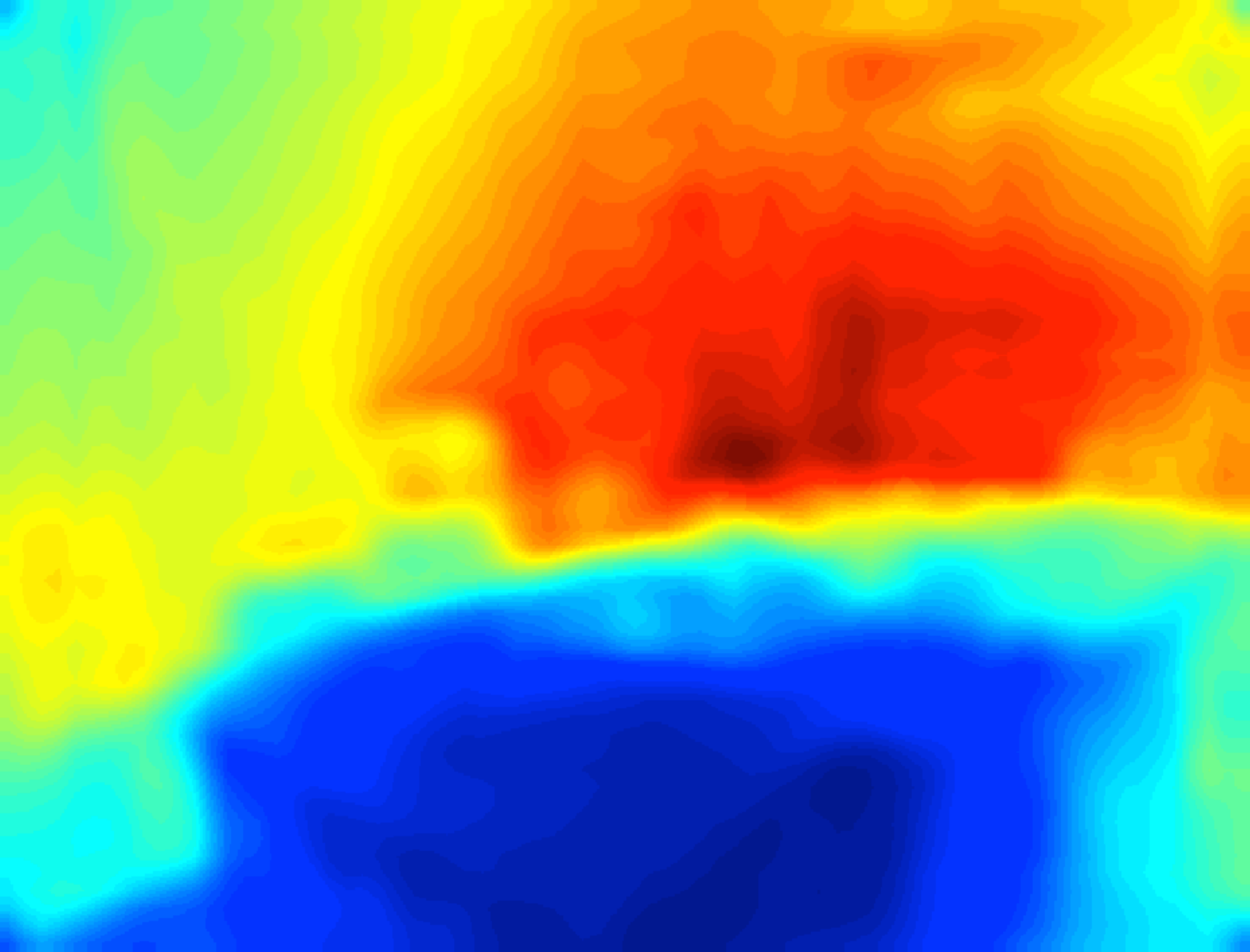}
    &\includegraphics[width=0.2\textwidth, height=0.15\textwidth]{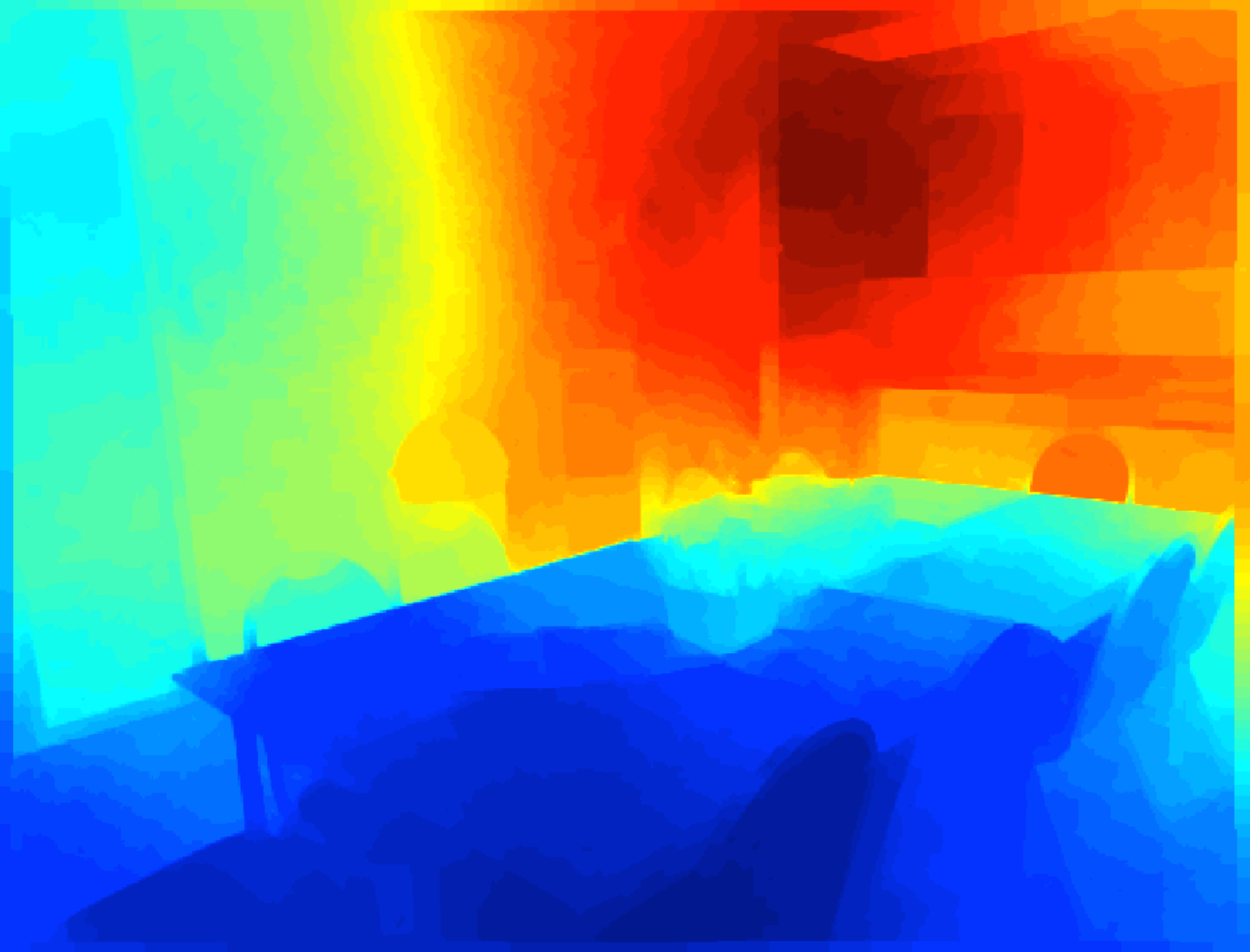}  \\

    \includegraphics[width=0.2\textwidth, height=0.15\textwidth]{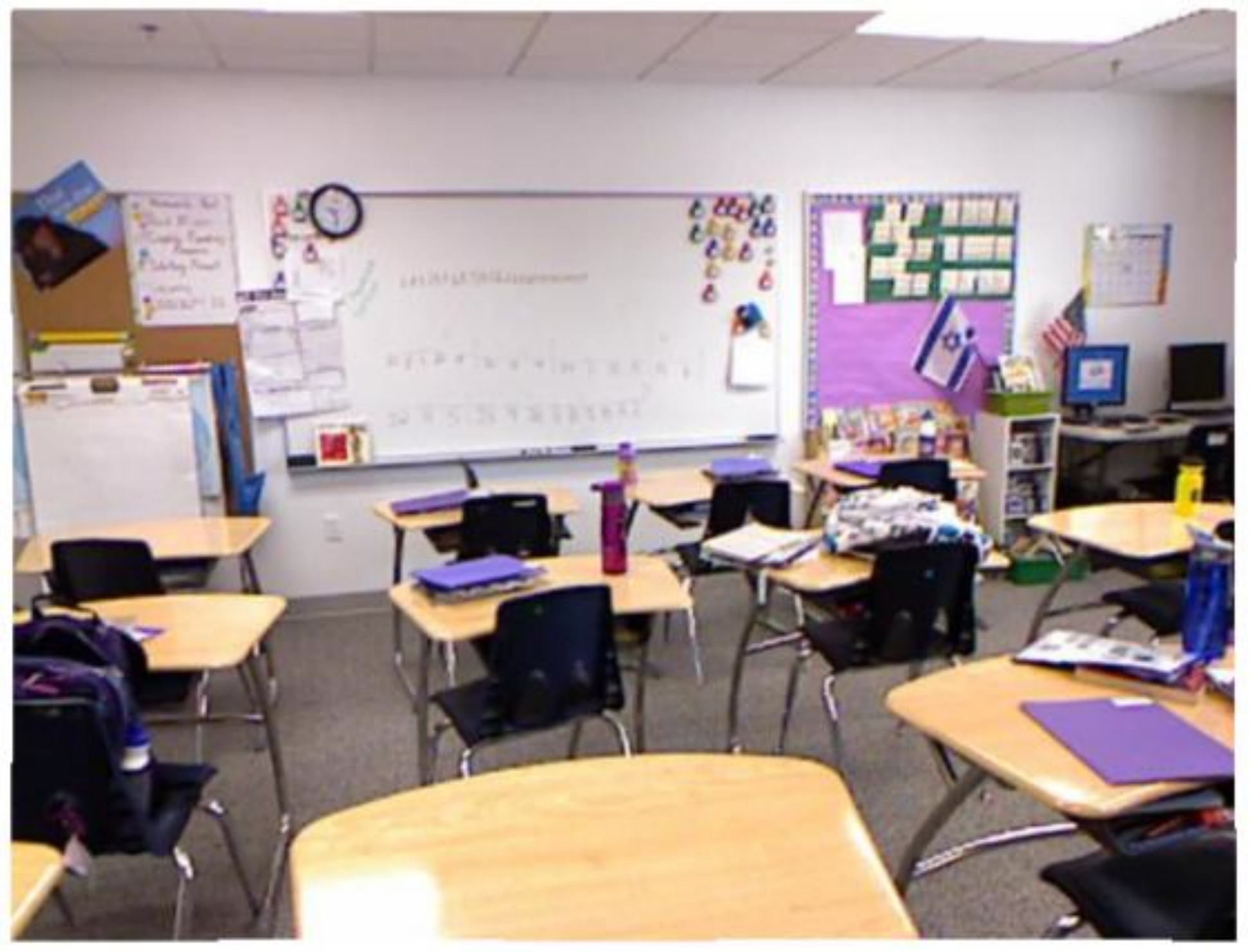}
	&\includegraphics[width=0.2\textwidth, height=0.15\textwidth]{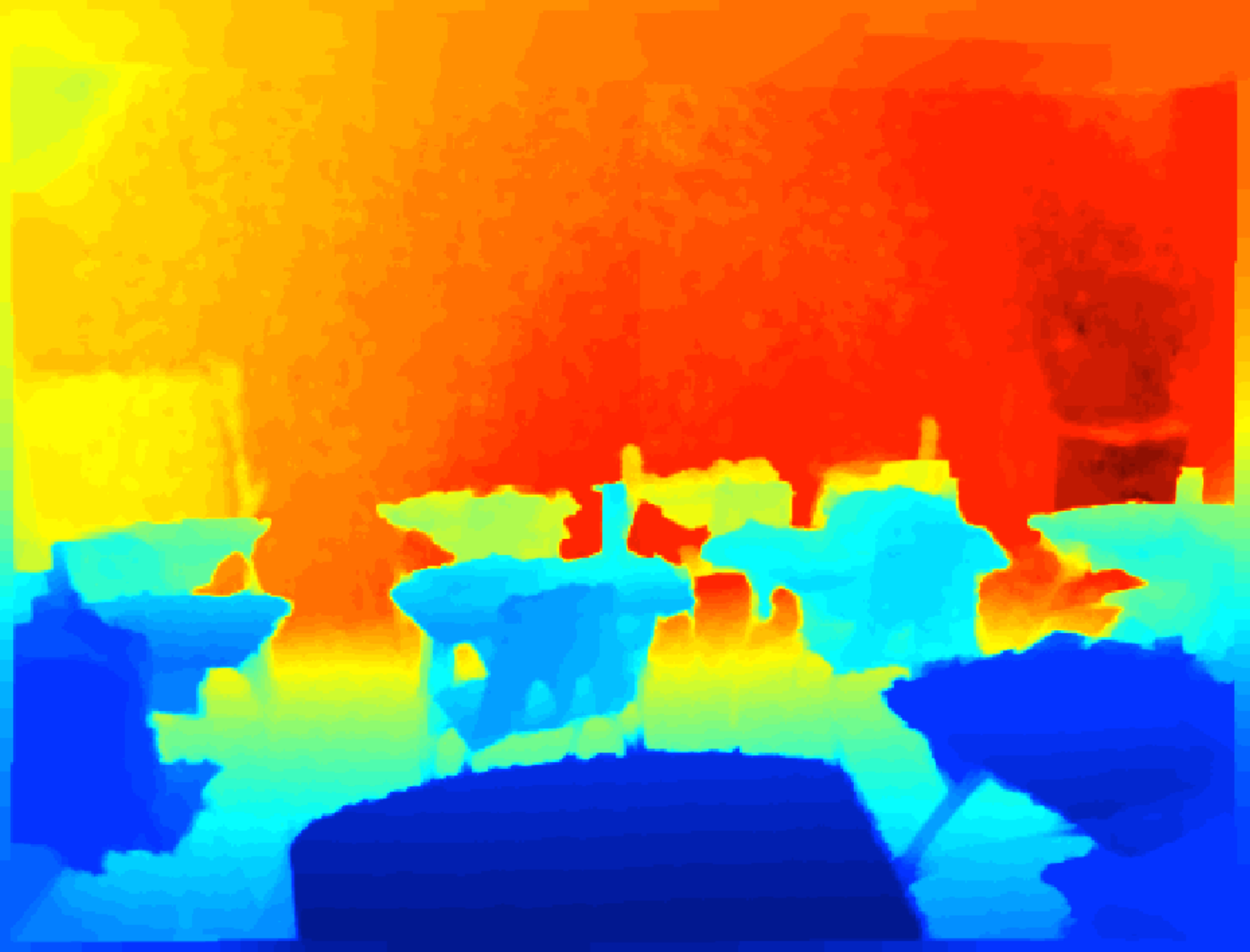}
 	&\includegraphics[width=0.2\textwidth, height=0.15\textwidth]{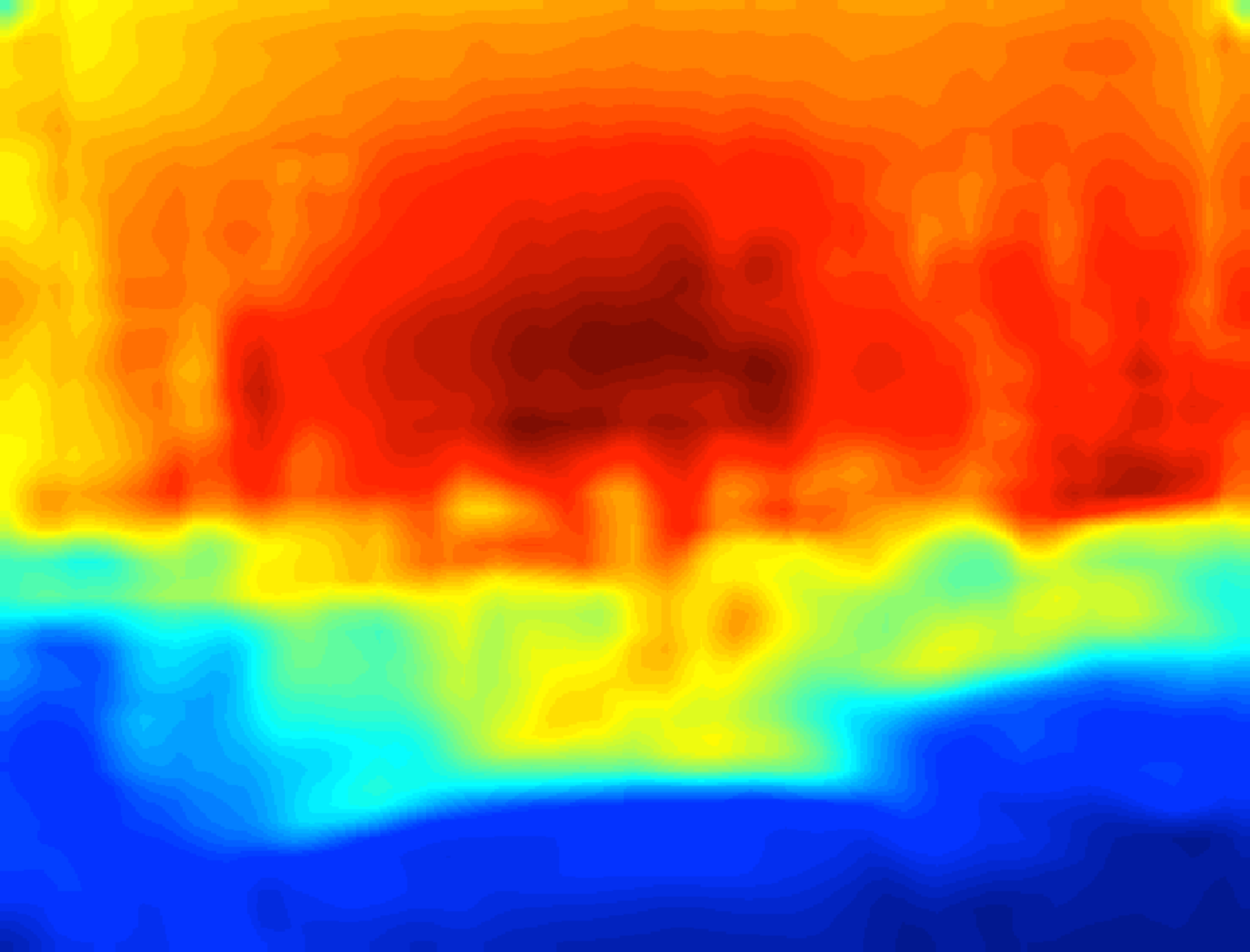}
 	&\includegraphics[width=0.2\textwidth, height=0.15\textwidth]{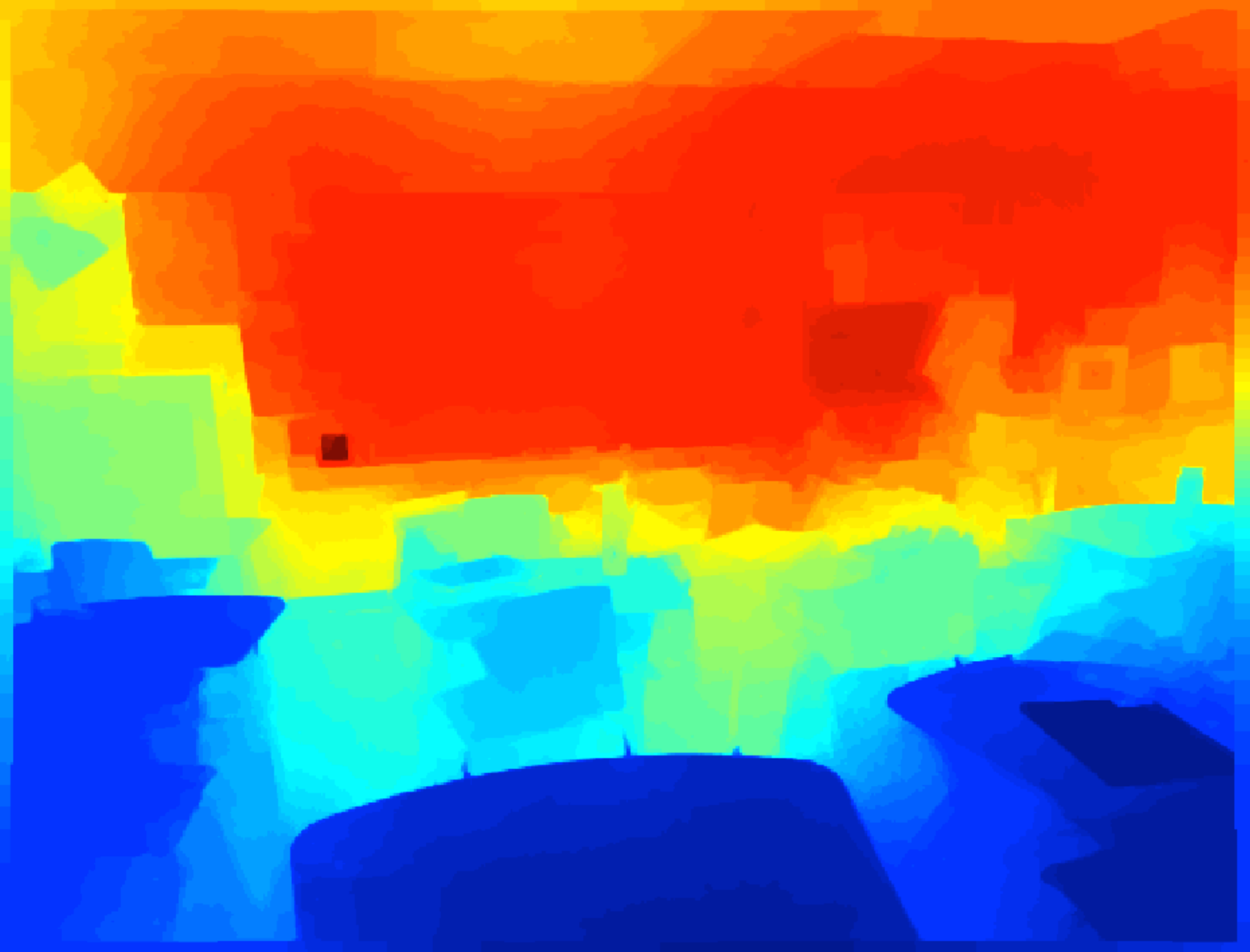} 	\\

    \includegraphics[width=0.2\textwidth, height=0.15\textwidth]{./fig/NYUD2/images/470.pdf}
    &\includegraphics[width=0.2\textwidth, height=0.15\textwidth]{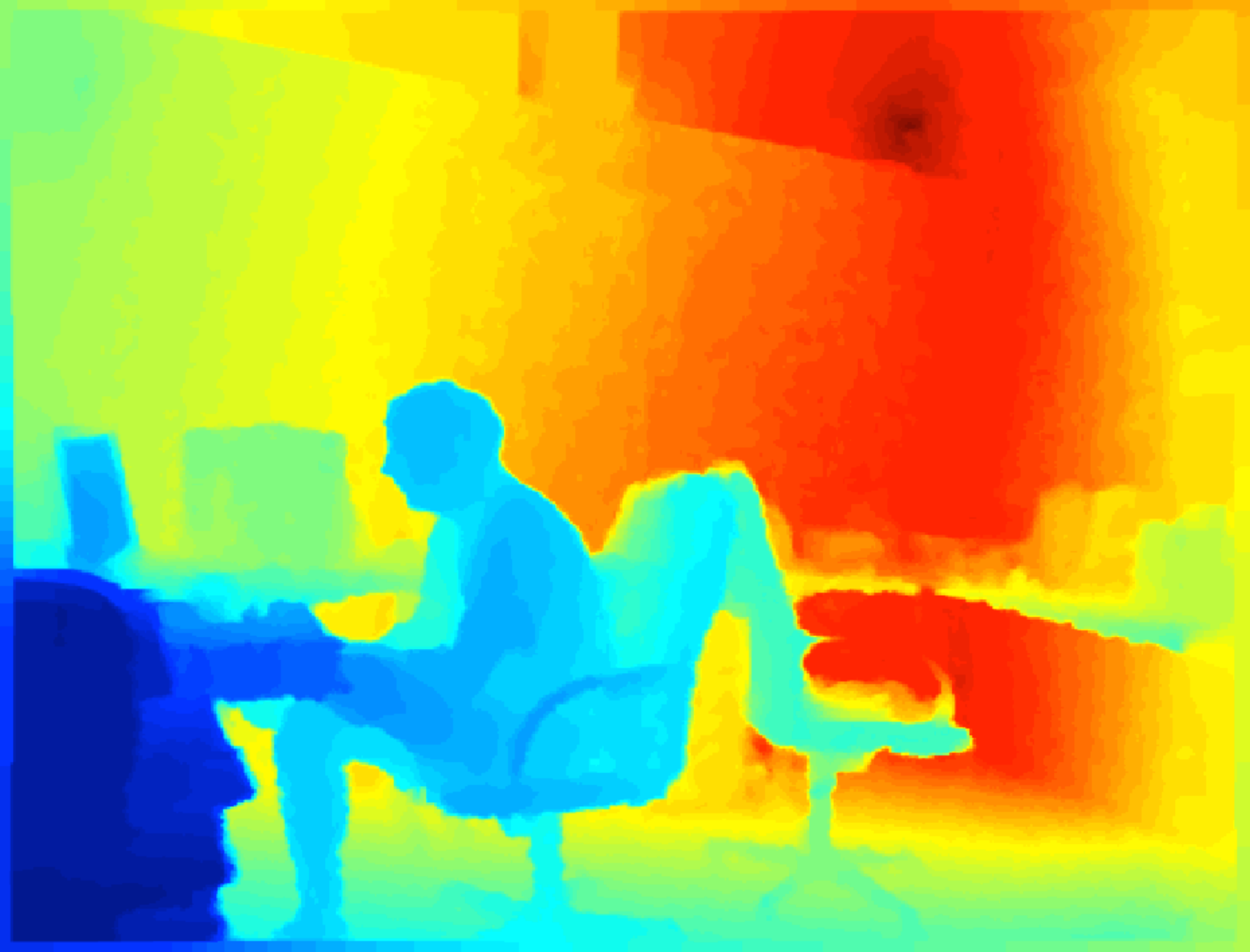}
    &\includegraphics[width=0.2\textwidth, height=0.15\textwidth]{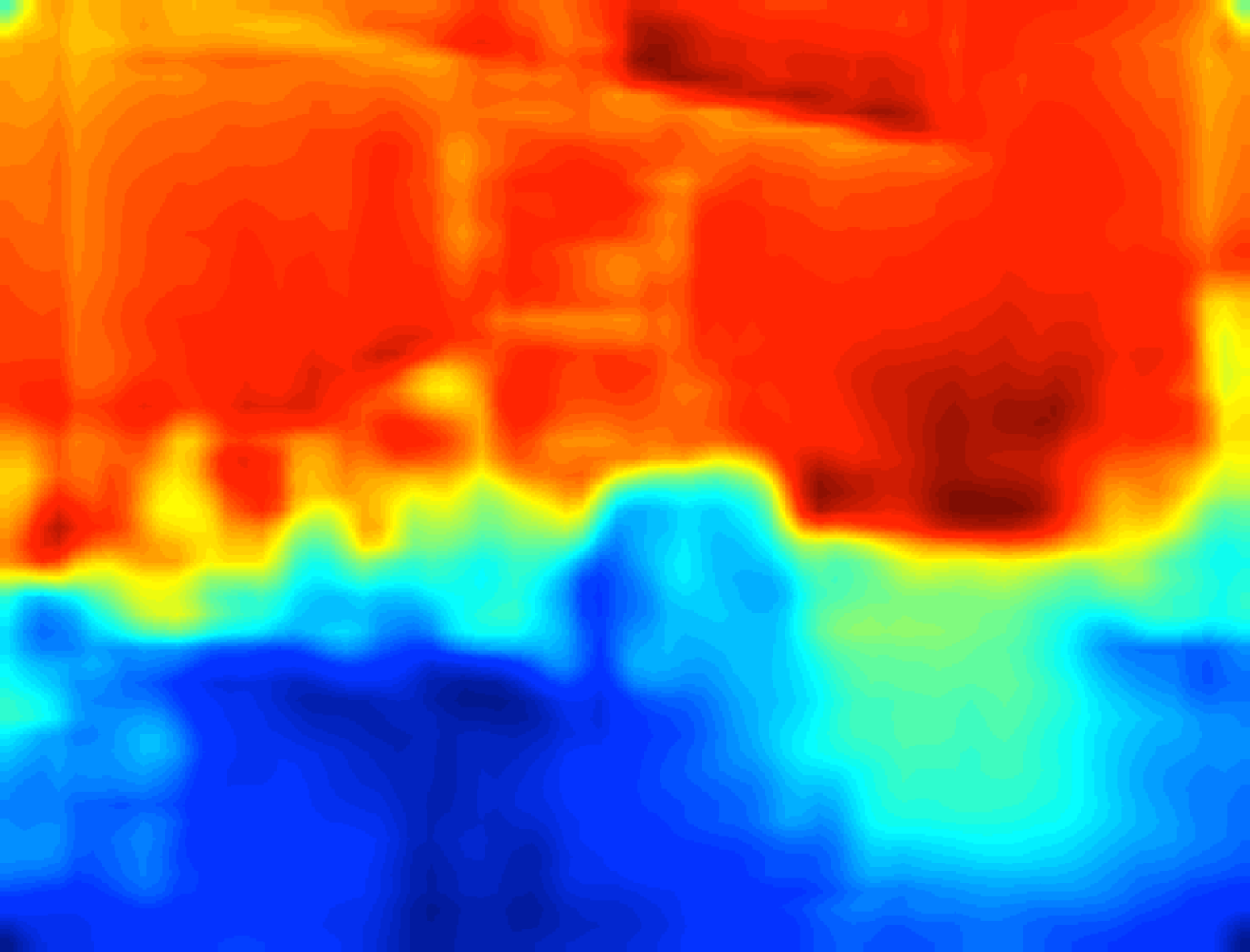}
    &\includegraphics[width=0.2\textwidth, height=0.15\textwidth]{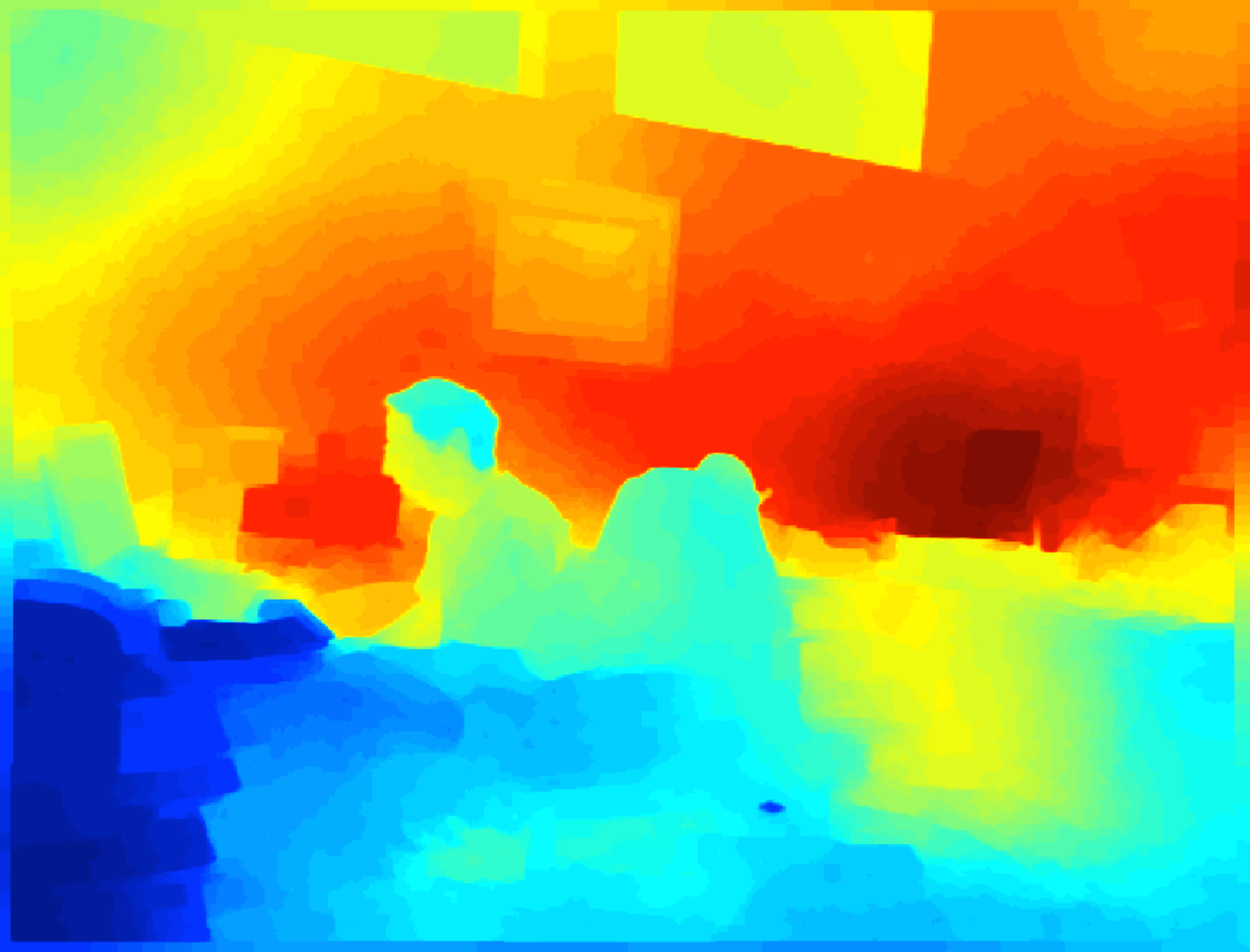}	\\

  {\footnotesize{
  Test image}} & {\footnotesize{ Ground-truth} }  & {\footnotesize{Eigen \etal  \cite{dcnn_nips14}  }}  & {\footnotesize{\dcnfsp } }
\end{tabular}
}
\caption{Examples of qualitative comparisons on the NYUD2 dataset (Best viewed on screen).
Color indicates depths (red is far, blue is close).
Our method yields visually better predictions with sharper transitions, aligning to local details.
 }  \label{fig:nyud2}
\end{figure*}

\begin{figure*} [t] \center
\resizebox{.9\linewidth}{!} {
\begin{tabular}{ccc|ccc}
	\includegraphics[width=0.165\textwidth]{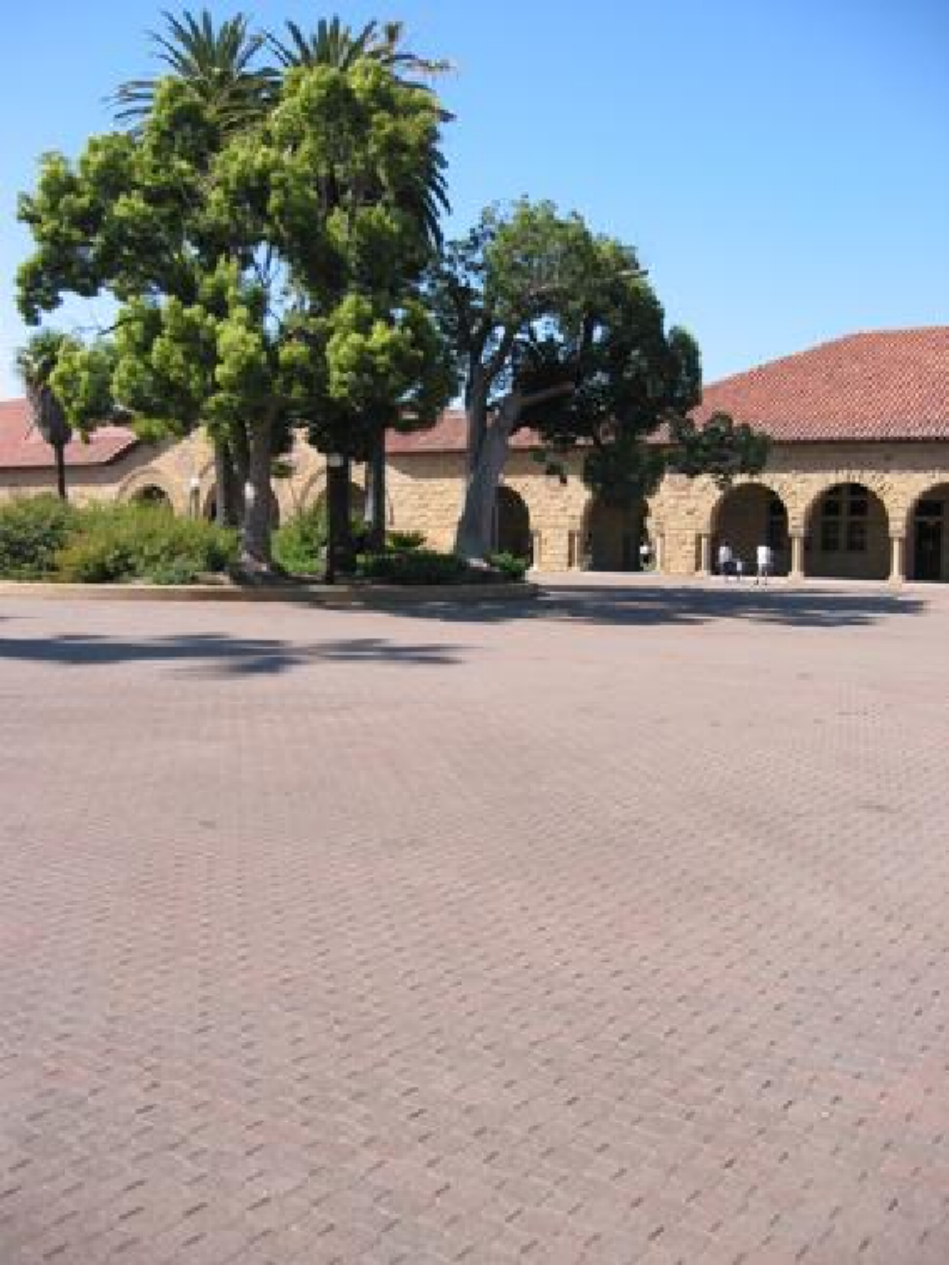}
	&\includegraphics[width=0.165\textwidth]{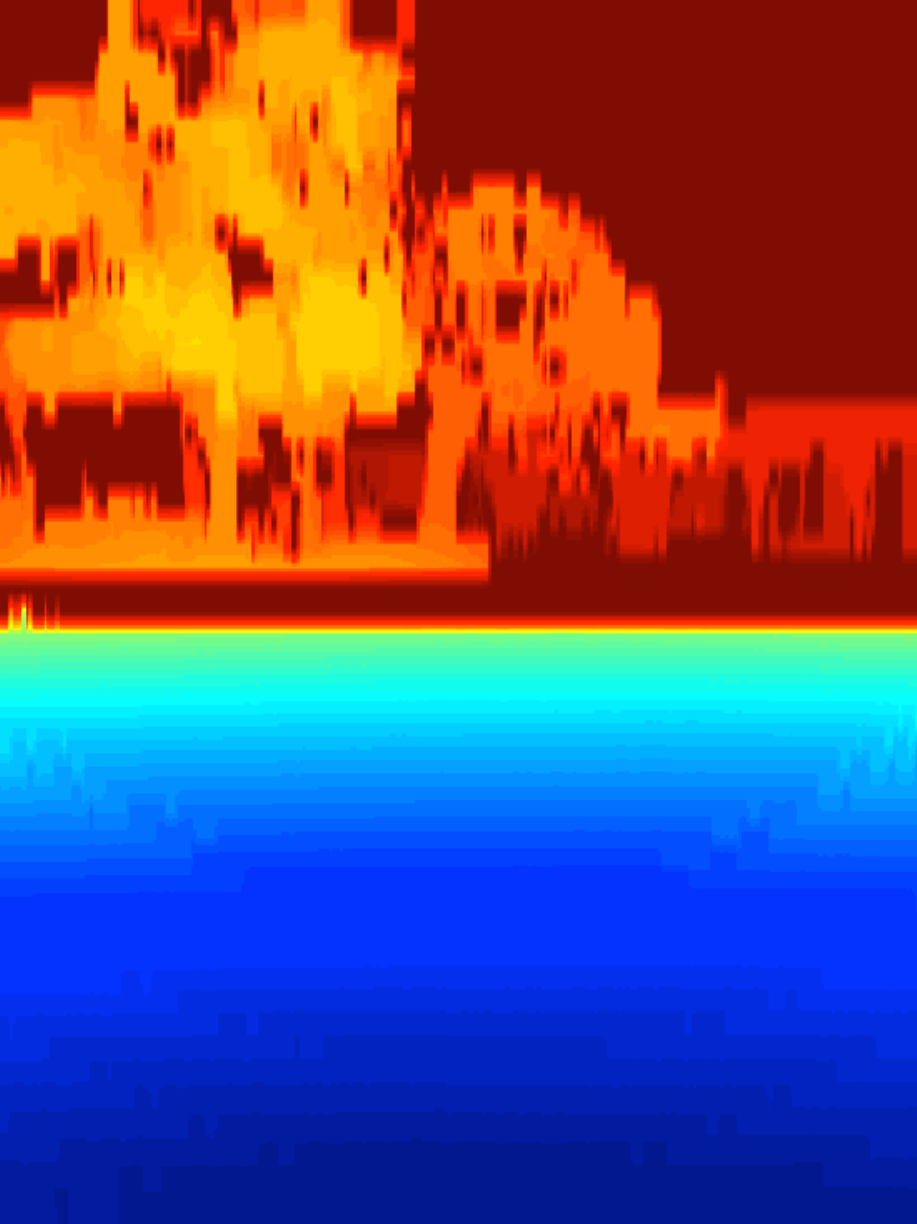}
	&\includegraphics[width=0.165\textwidth]{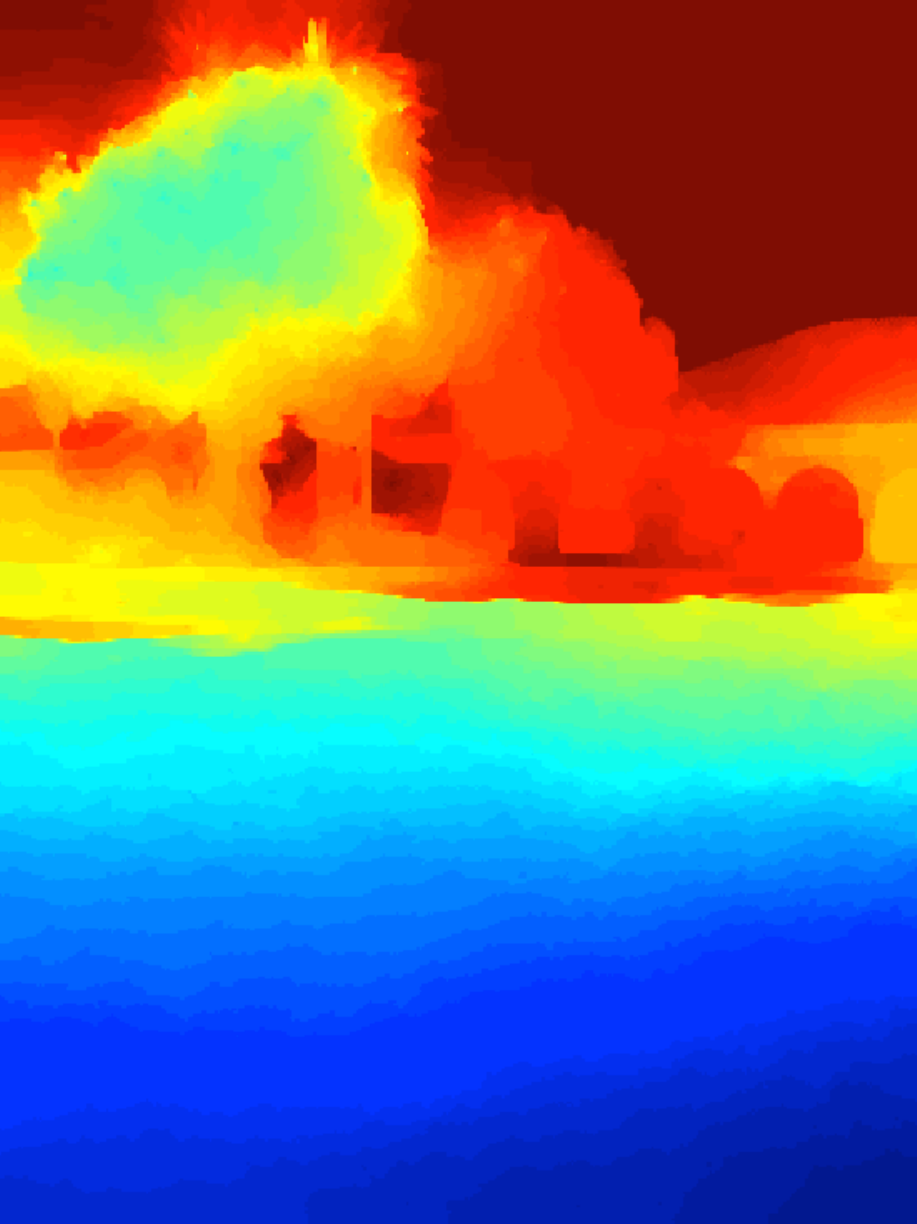}
	&\includegraphics[width=0.165\textwidth]{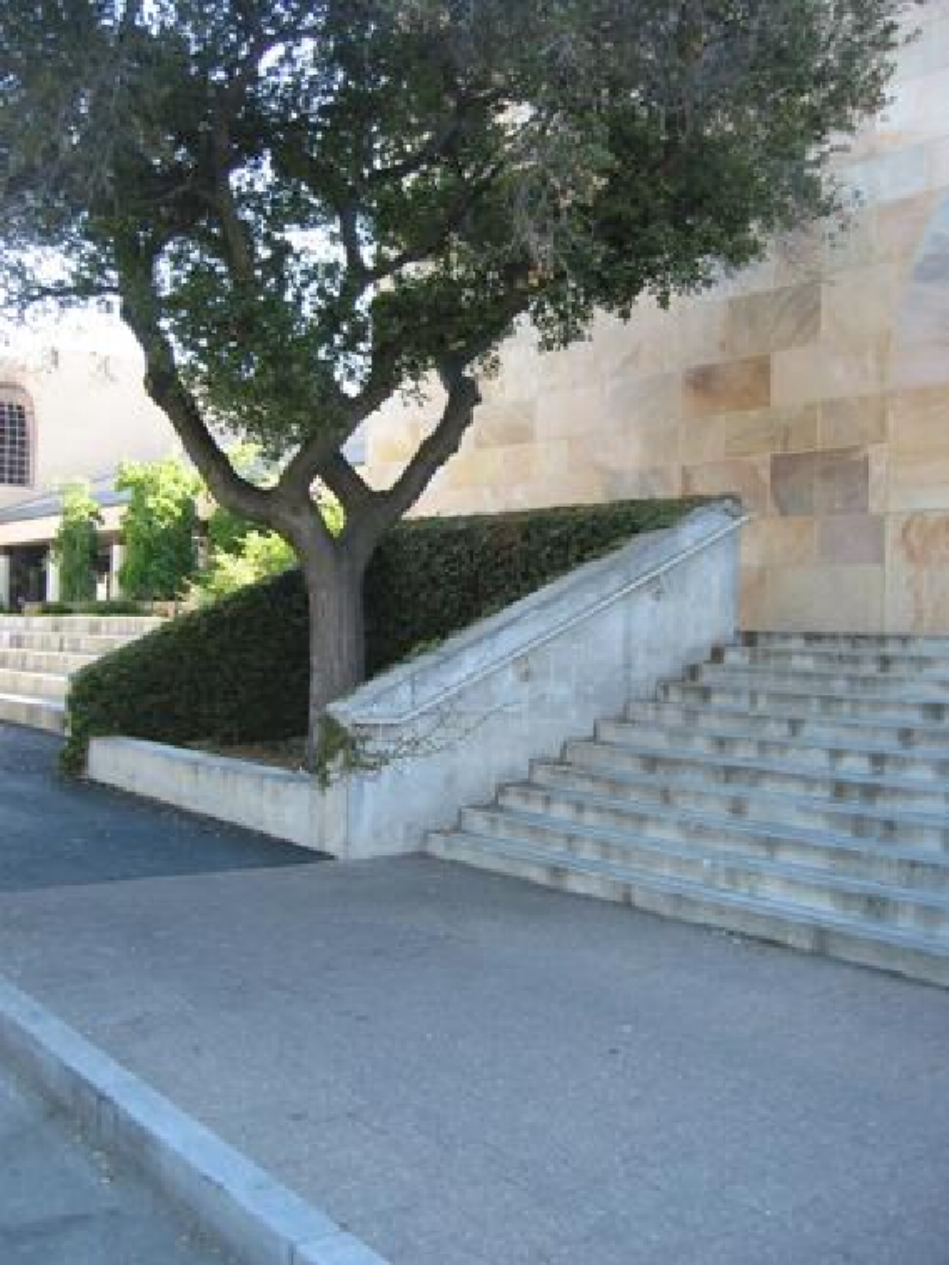}
	&\includegraphics[width=0.165\textwidth]{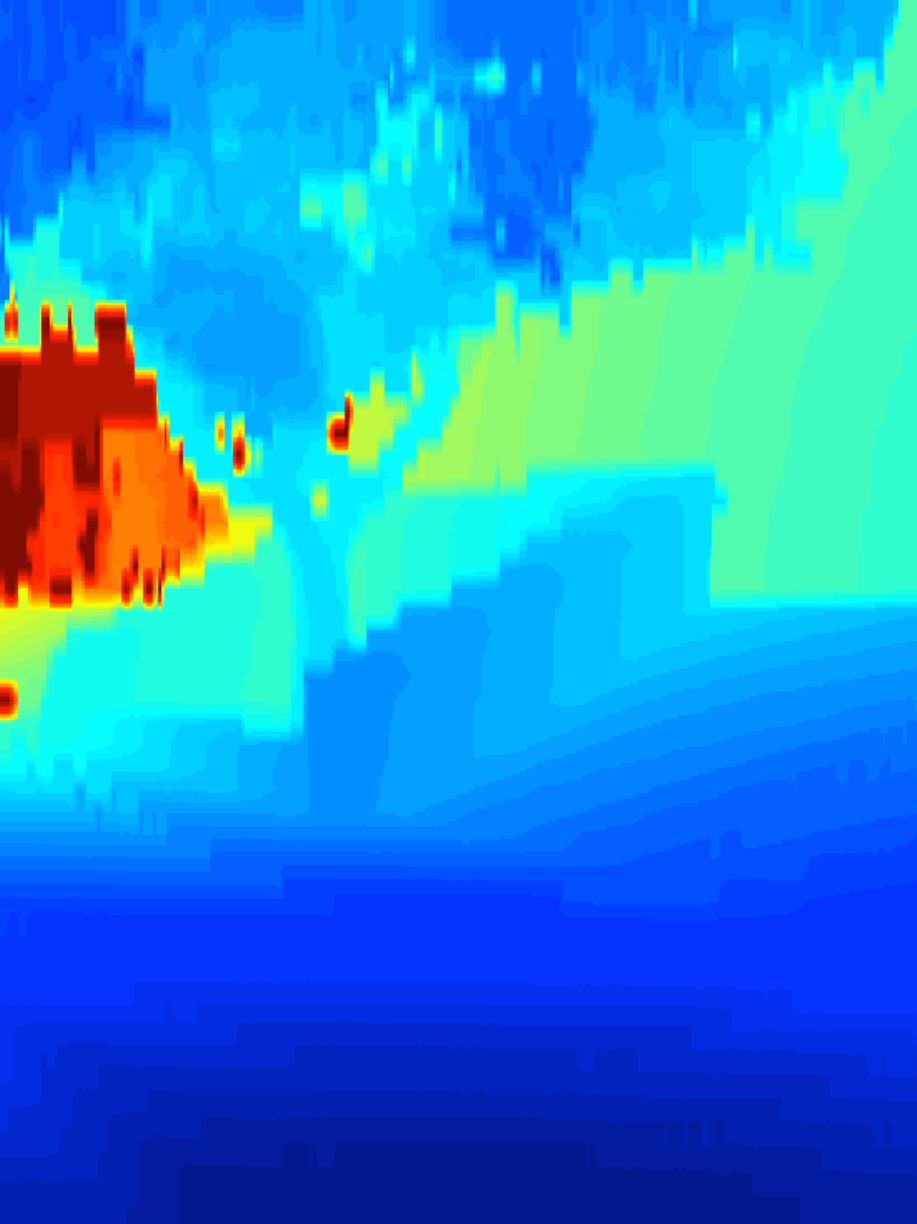}
	&\includegraphics[width=0.165\textwidth]{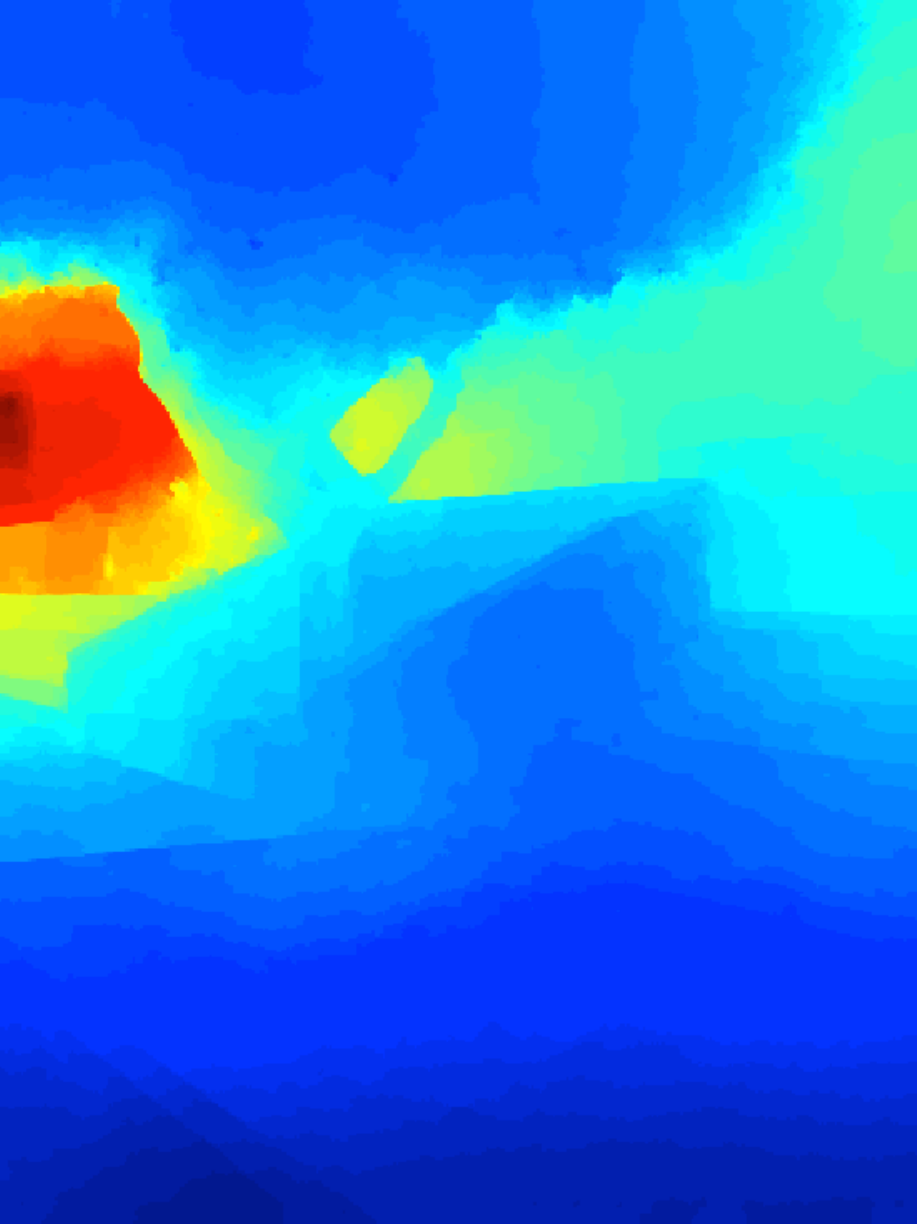} \\

	\includegraphics[width=0.165\textwidth]{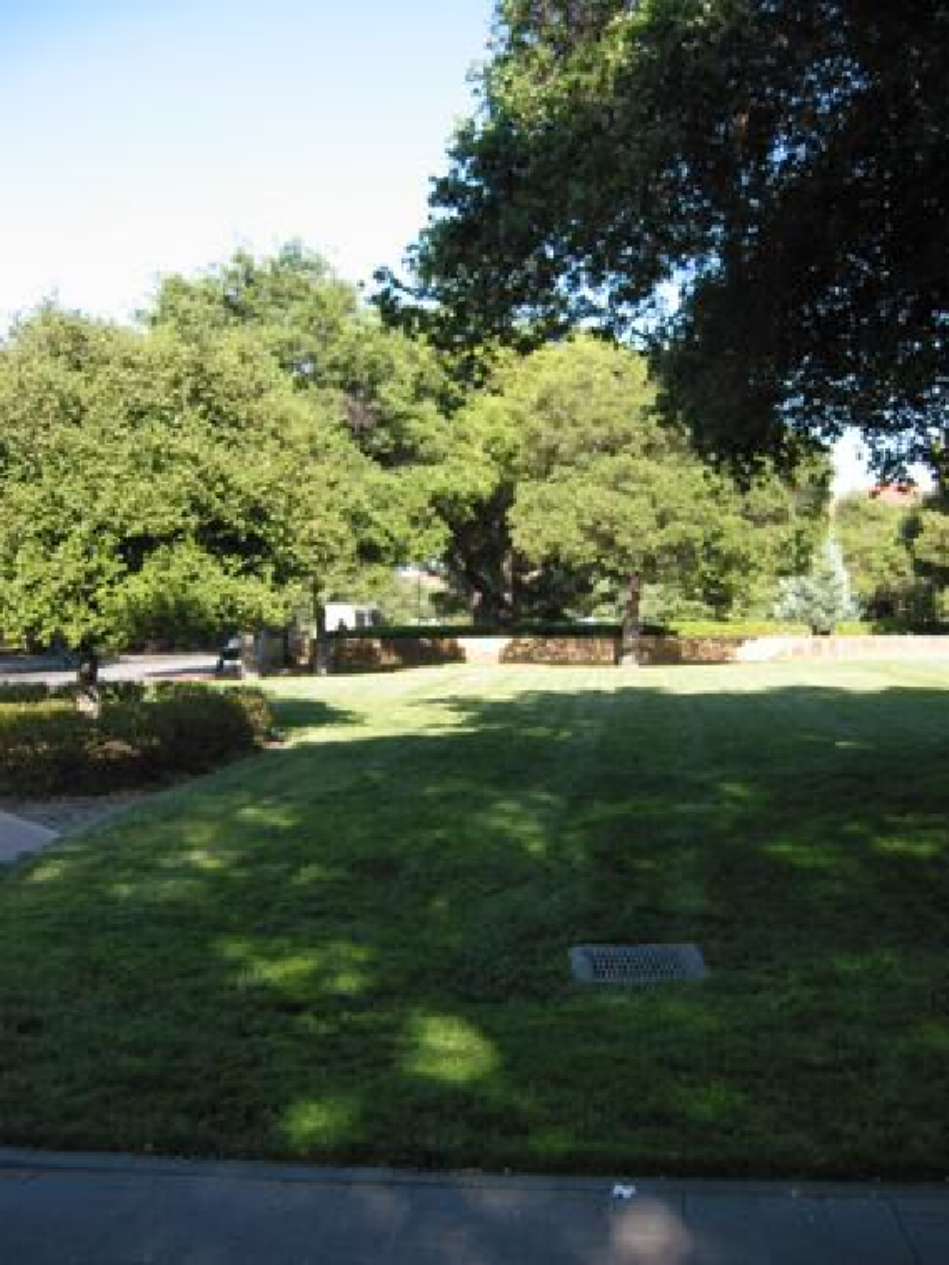}
	&\includegraphics[width=0.165\textwidth]{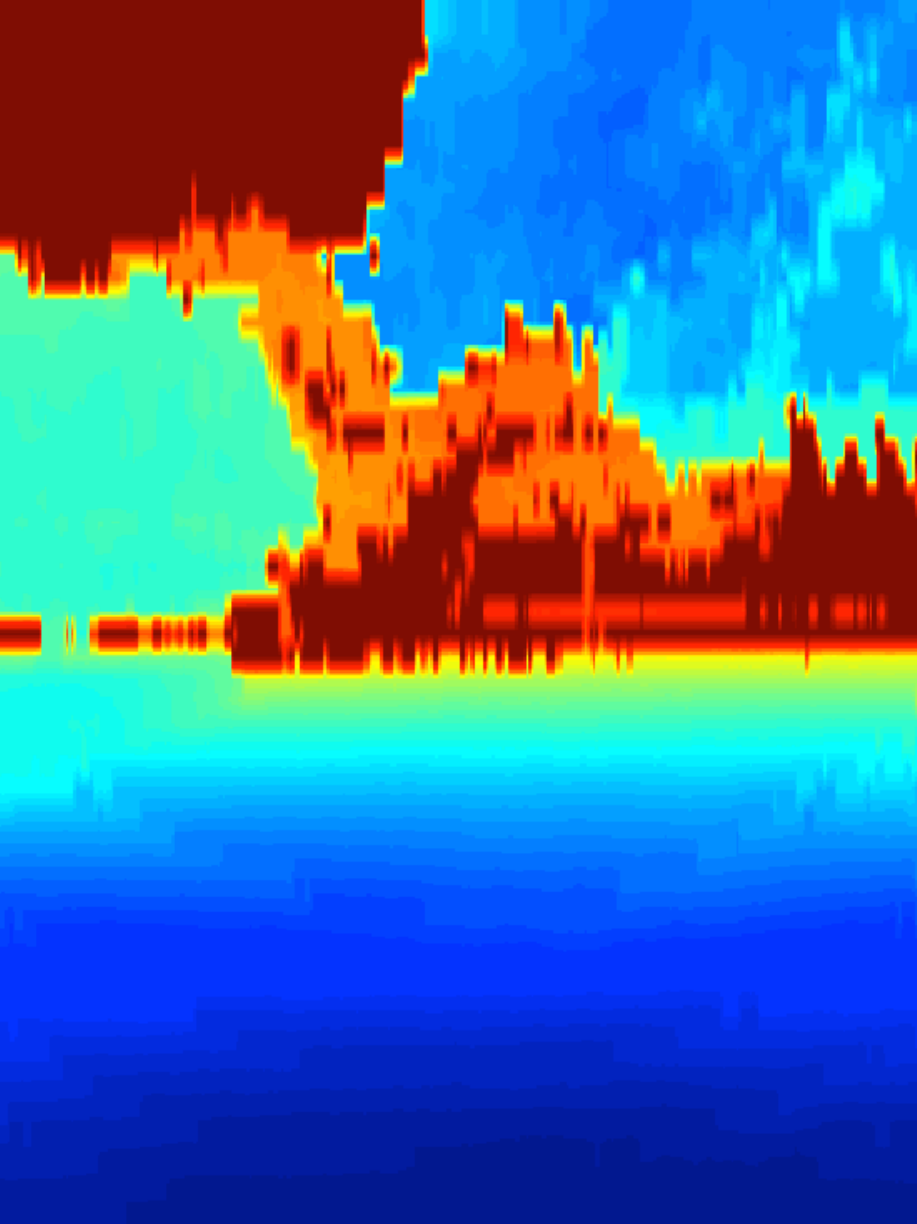}
	&\includegraphics[width=0.165\textwidth]{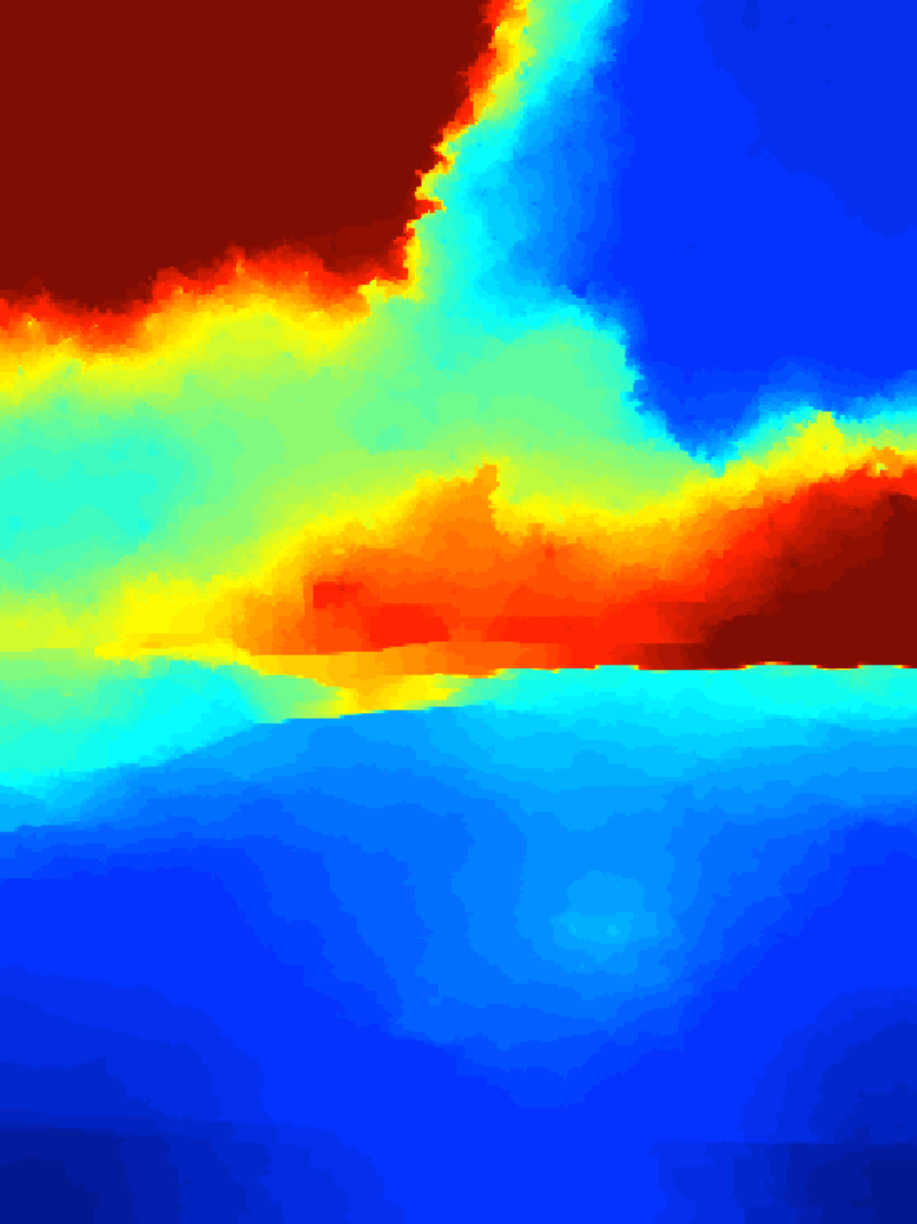}
	&\includegraphics[width=0.165\textwidth]{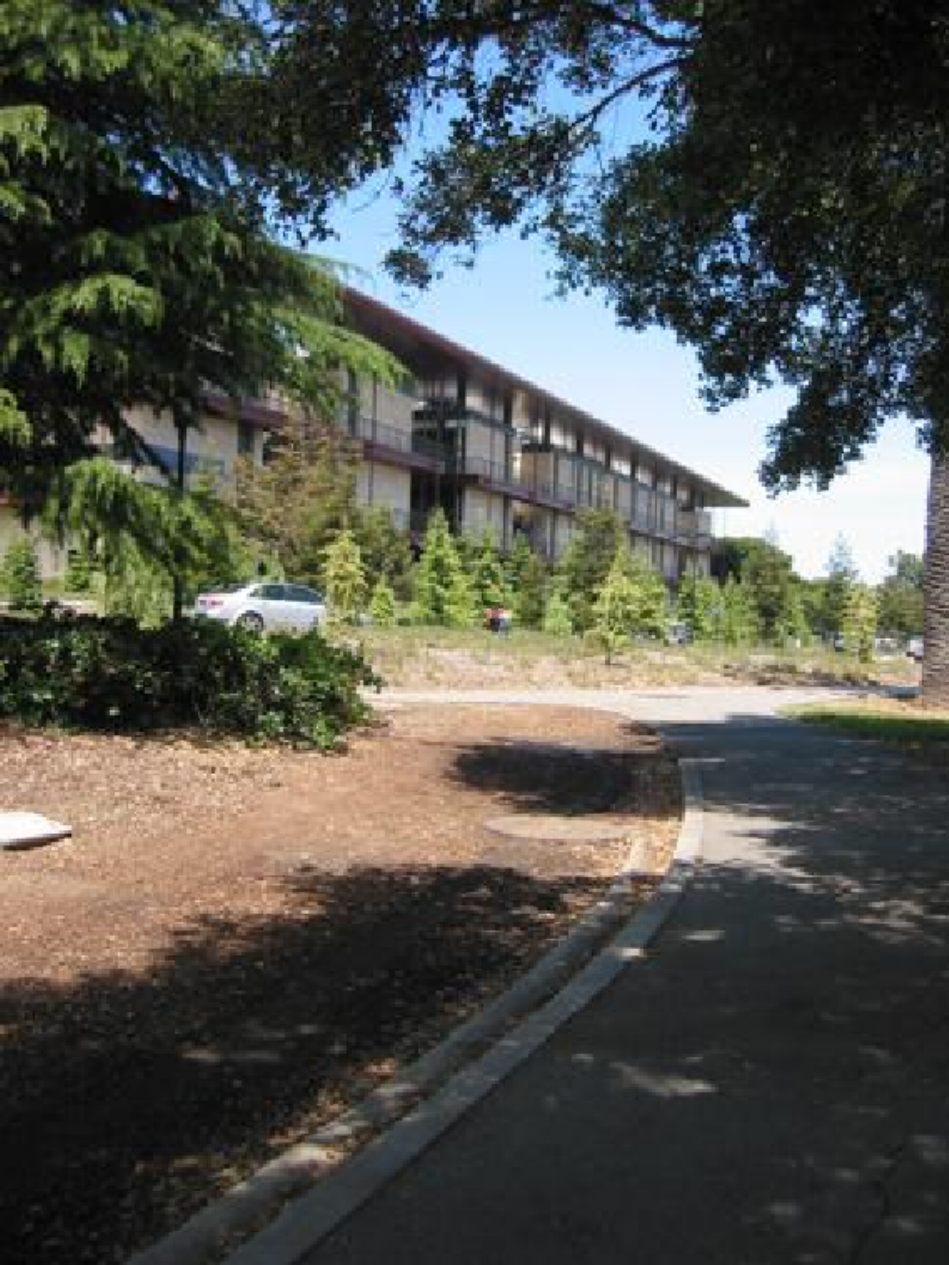}
	&\includegraphics[width=0.165\textwidth]{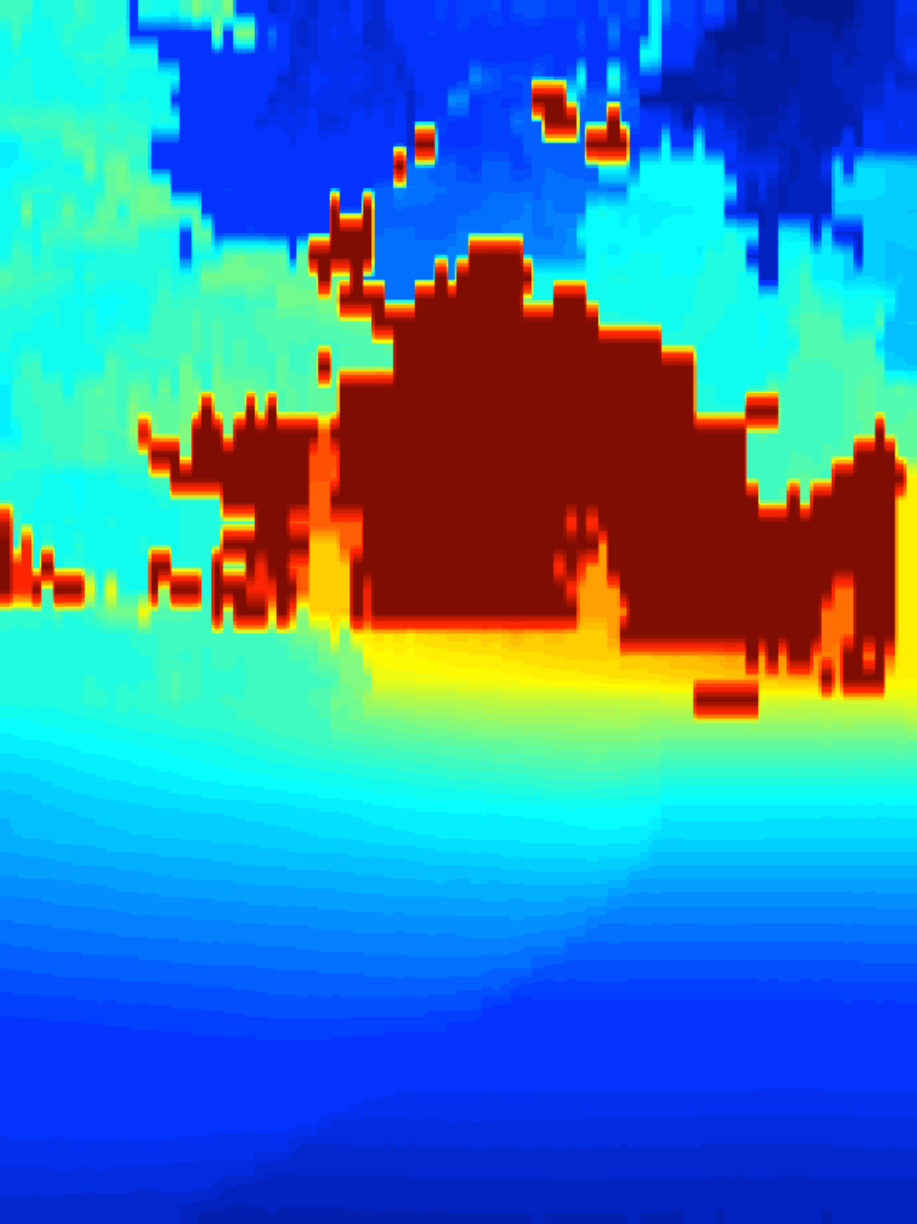}
	&\includegraphics[width=0.165\textwidth]{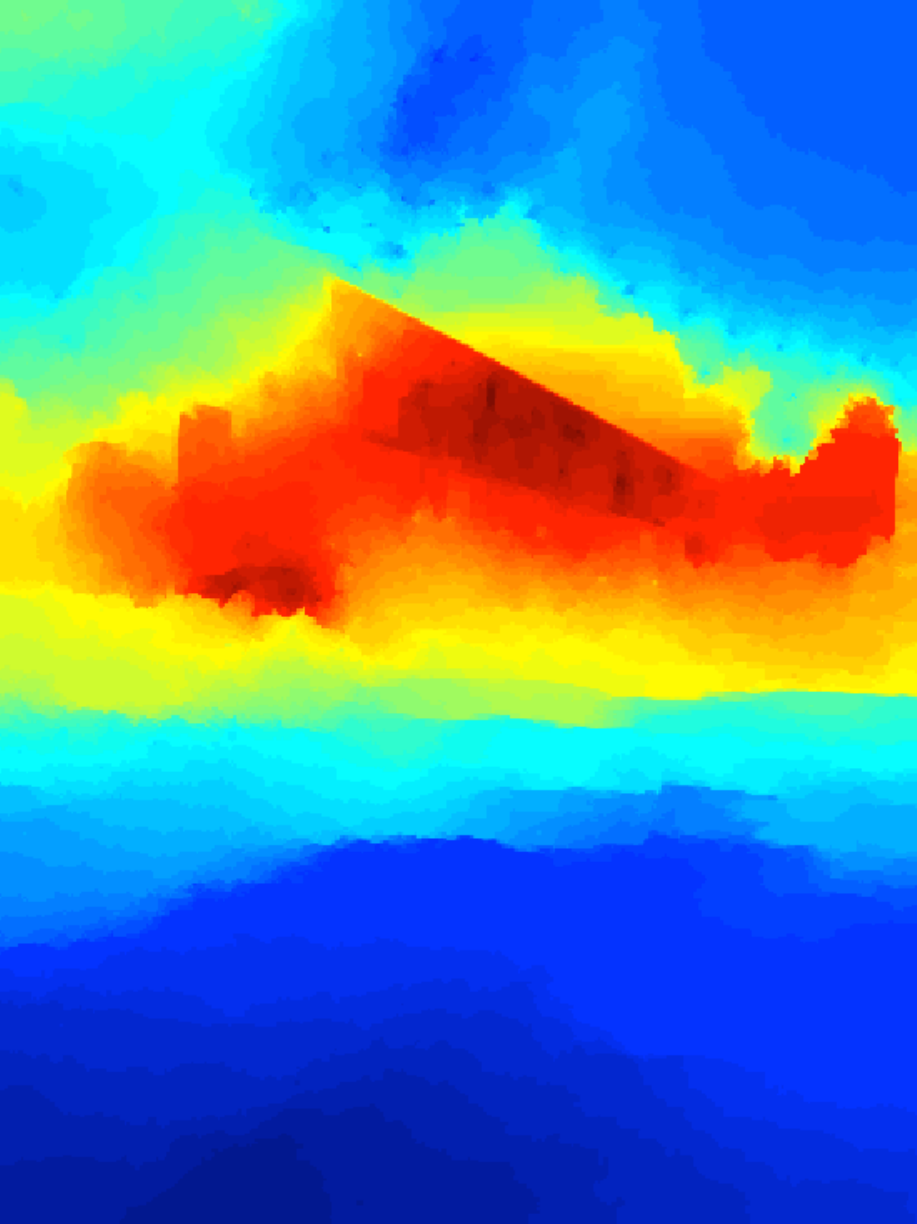} \\

	\includegraphics[width=0.165\textwidth]{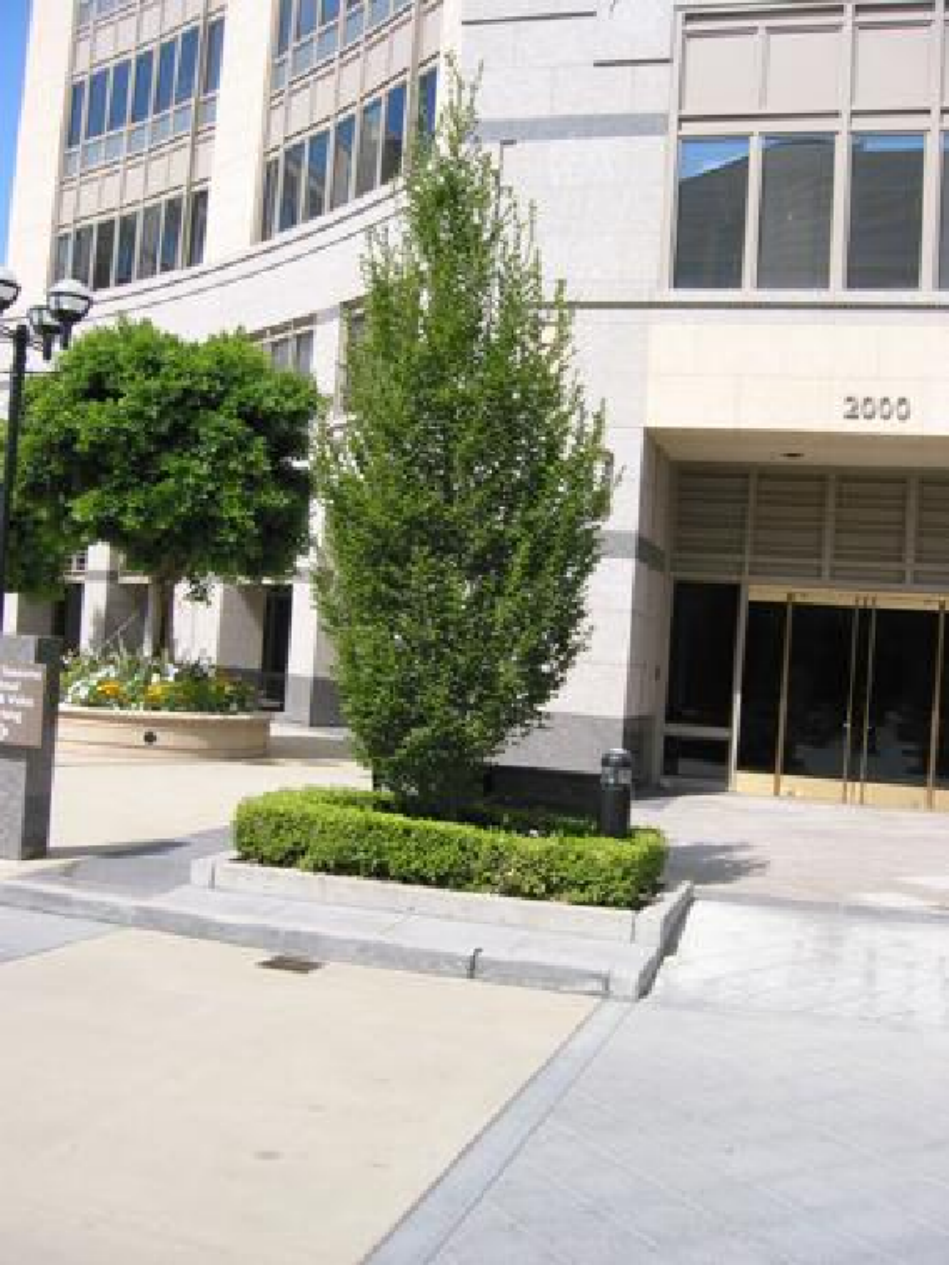}
	&\includegraphics[width=0.165\textwidth]{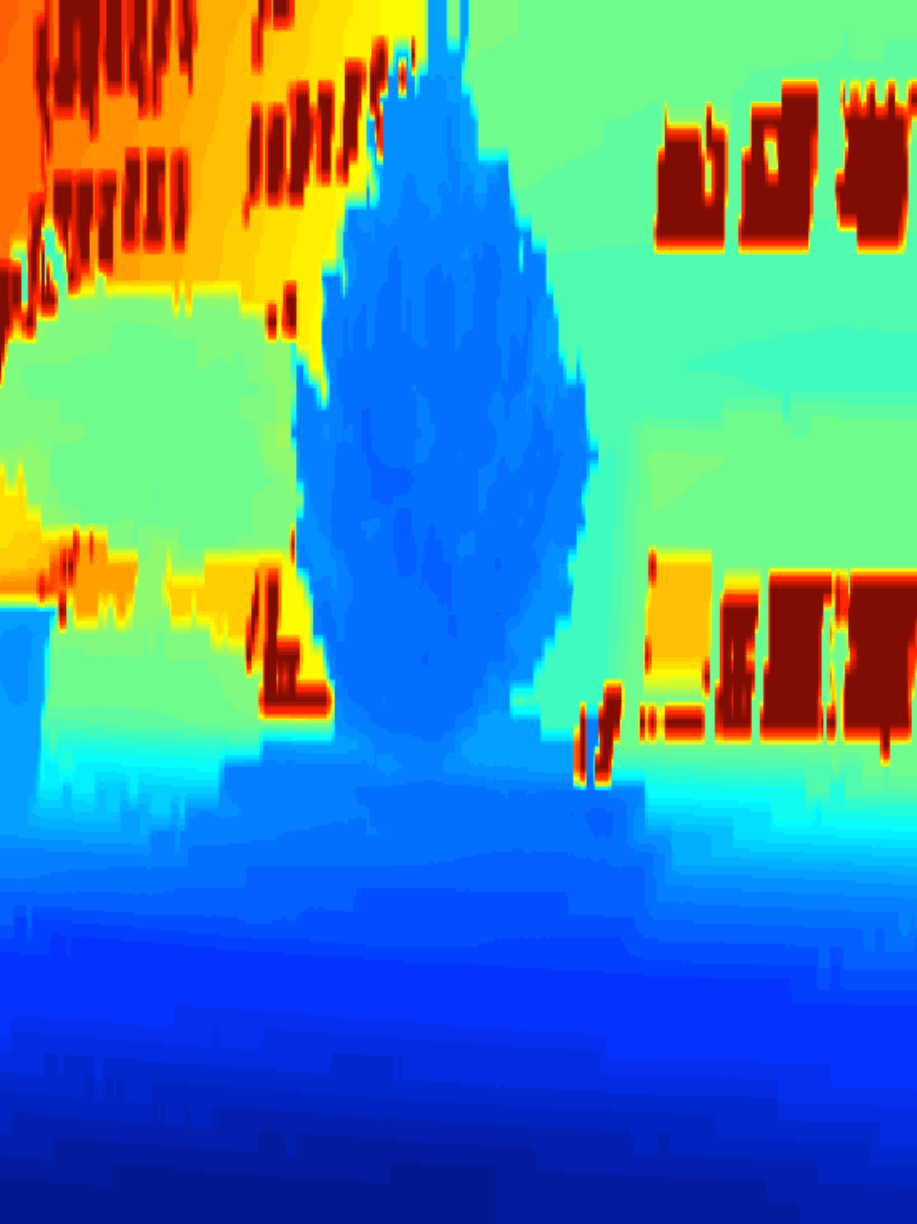}
	&\includegraphics[width=0.165\textwidth]{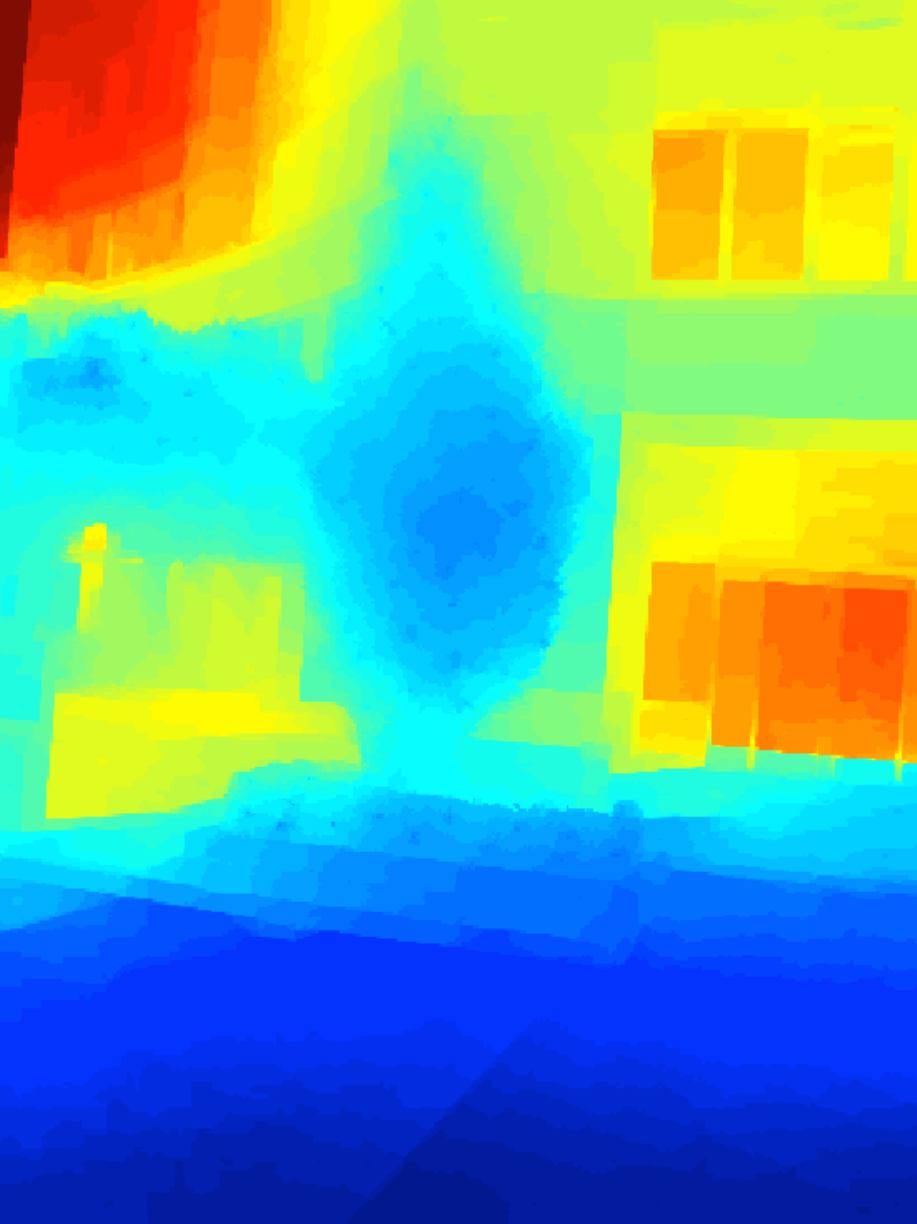}
	&\includegraphics[width=0.165\textwidth]{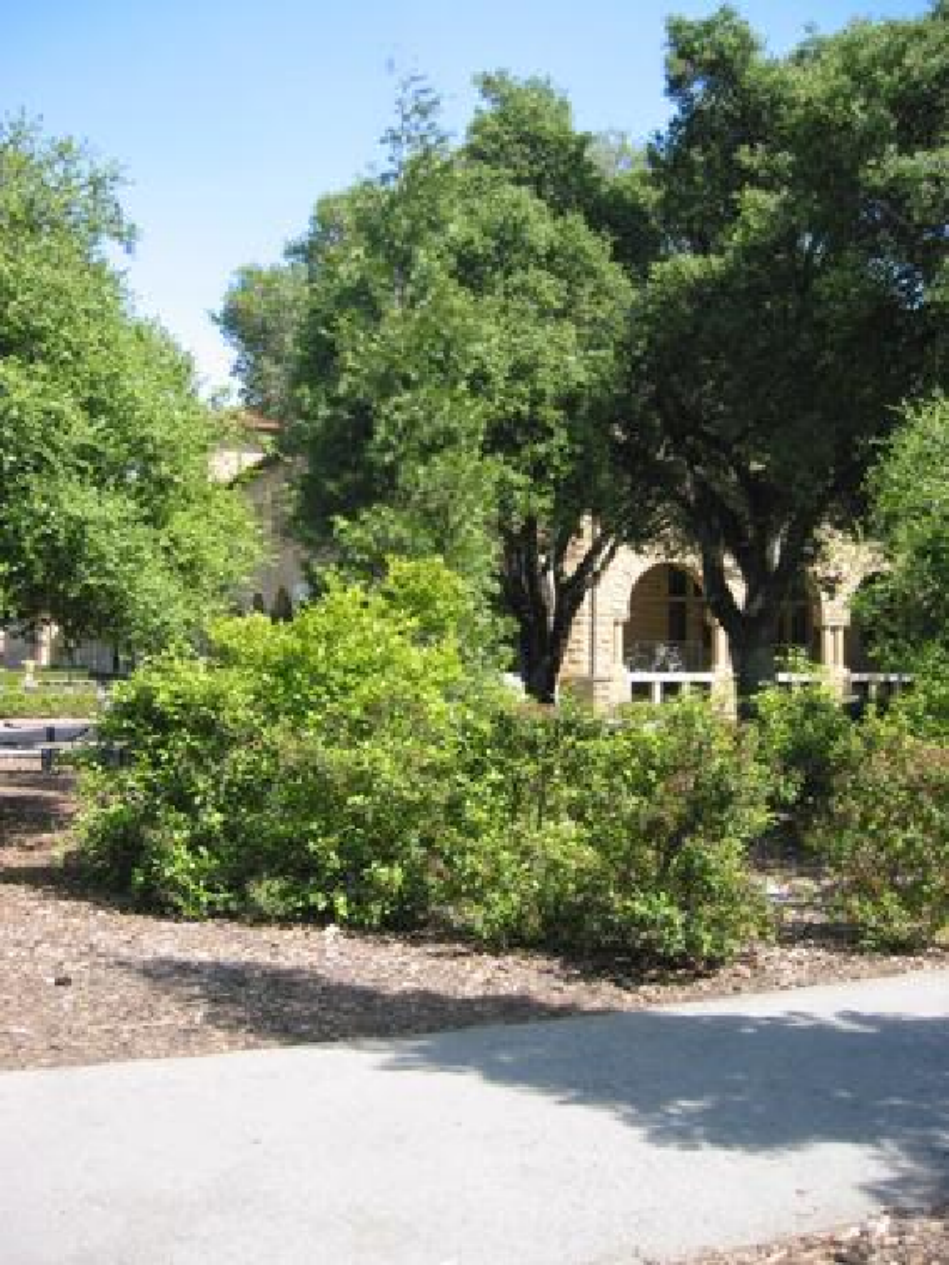}
	&\includegraphics[width=0.165\textwidth]{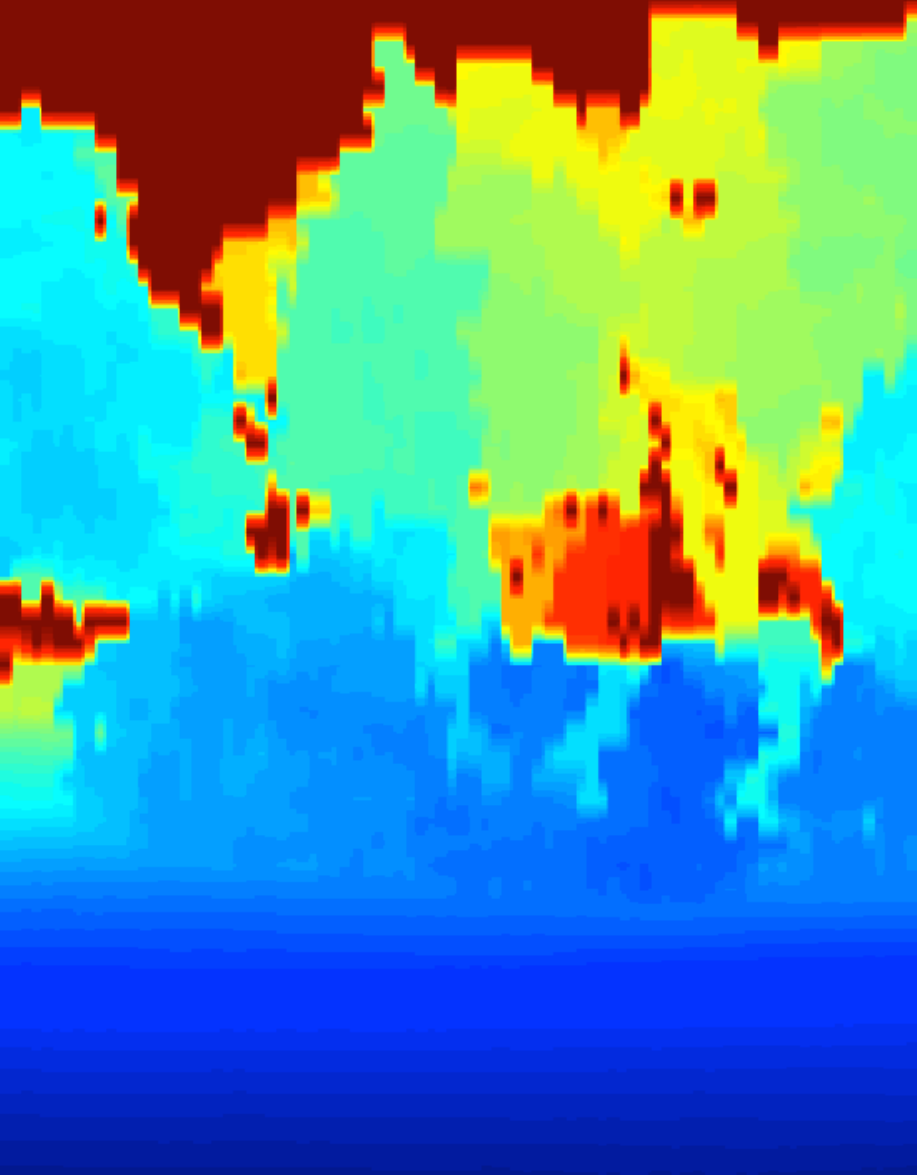}
	&\includegraphics[width=0.165\textwidth]{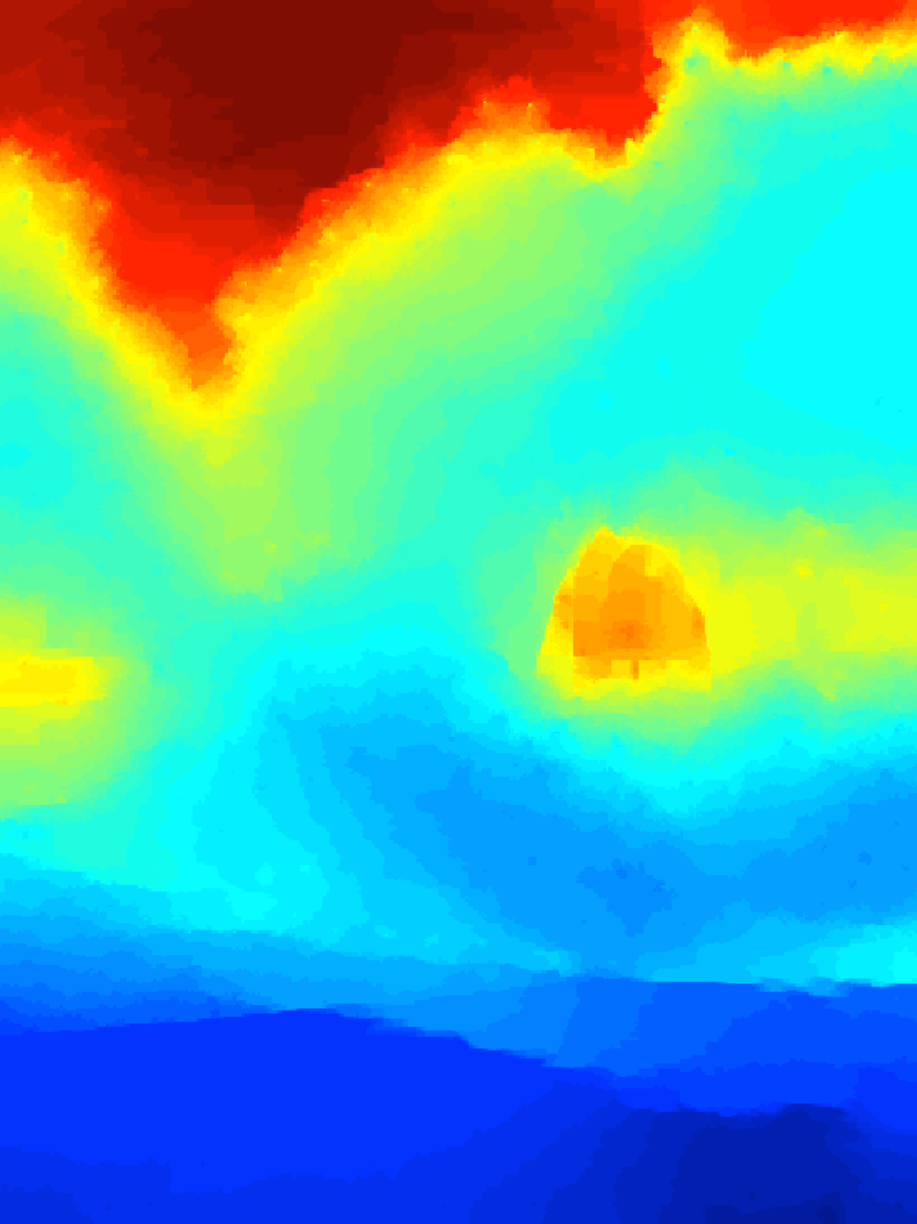}  \\

	\includegraphics[width=0.165\textwidth]{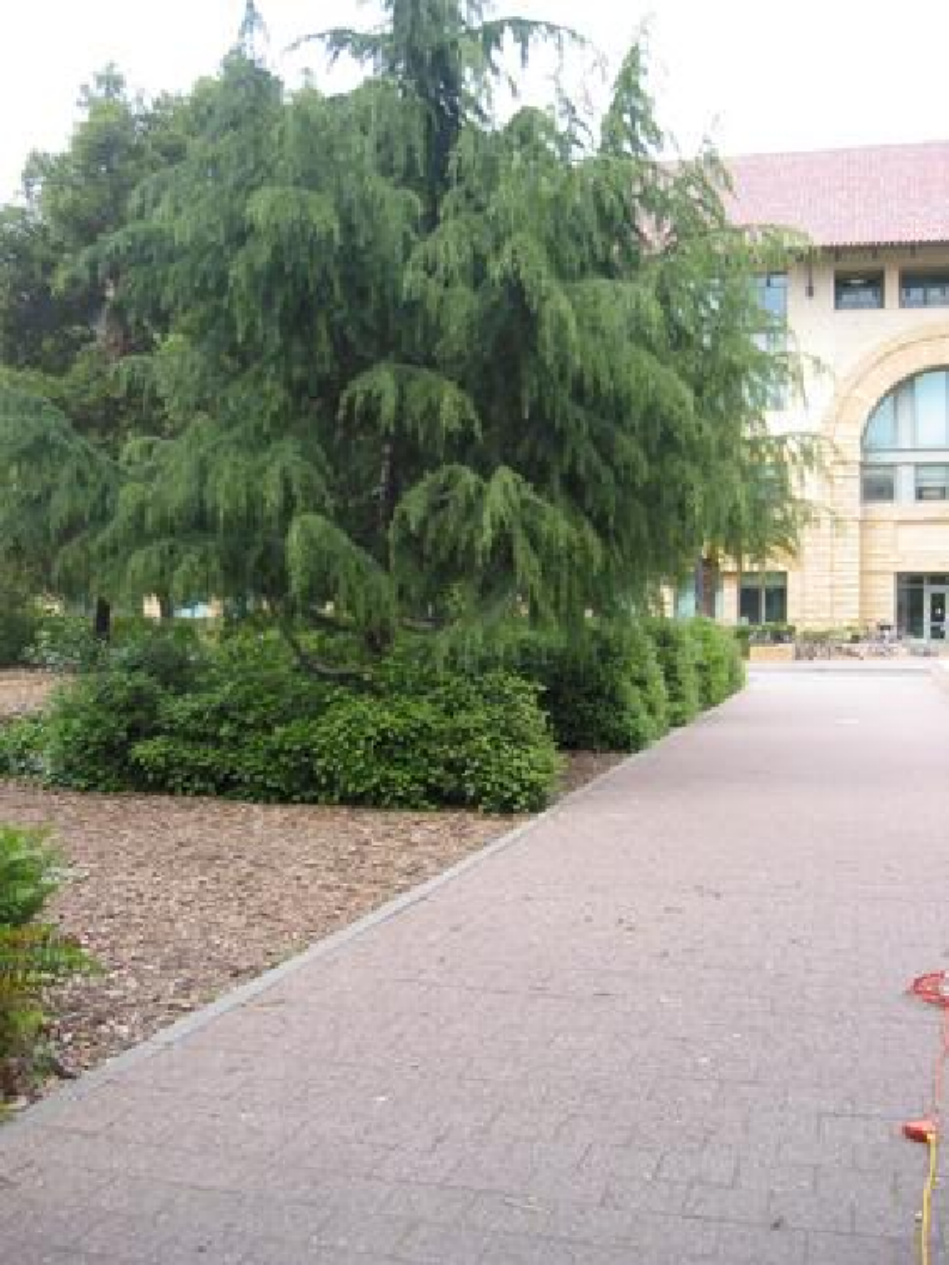}
	&\includegraphics[width=0.165\textwidth]{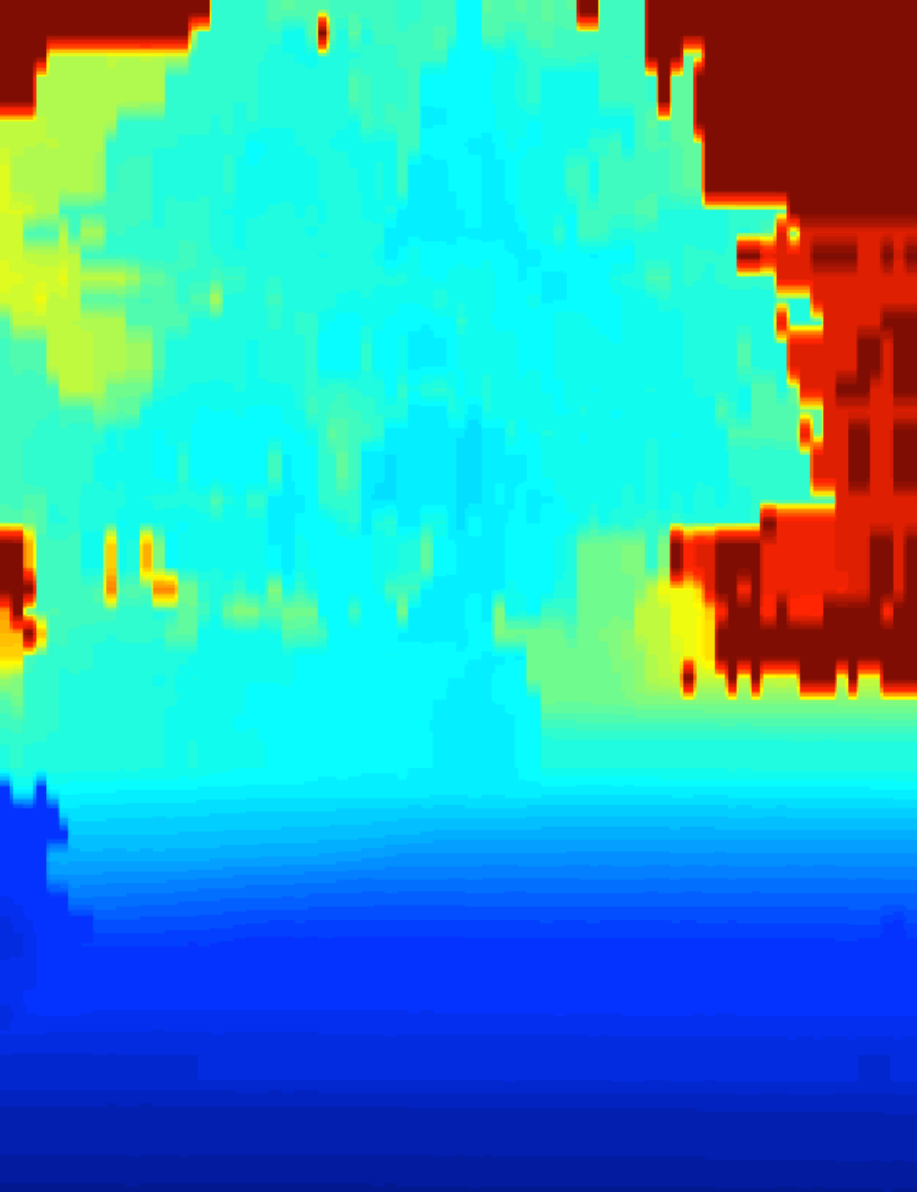}
	&\includegraphics[width=0.165\textwidth]{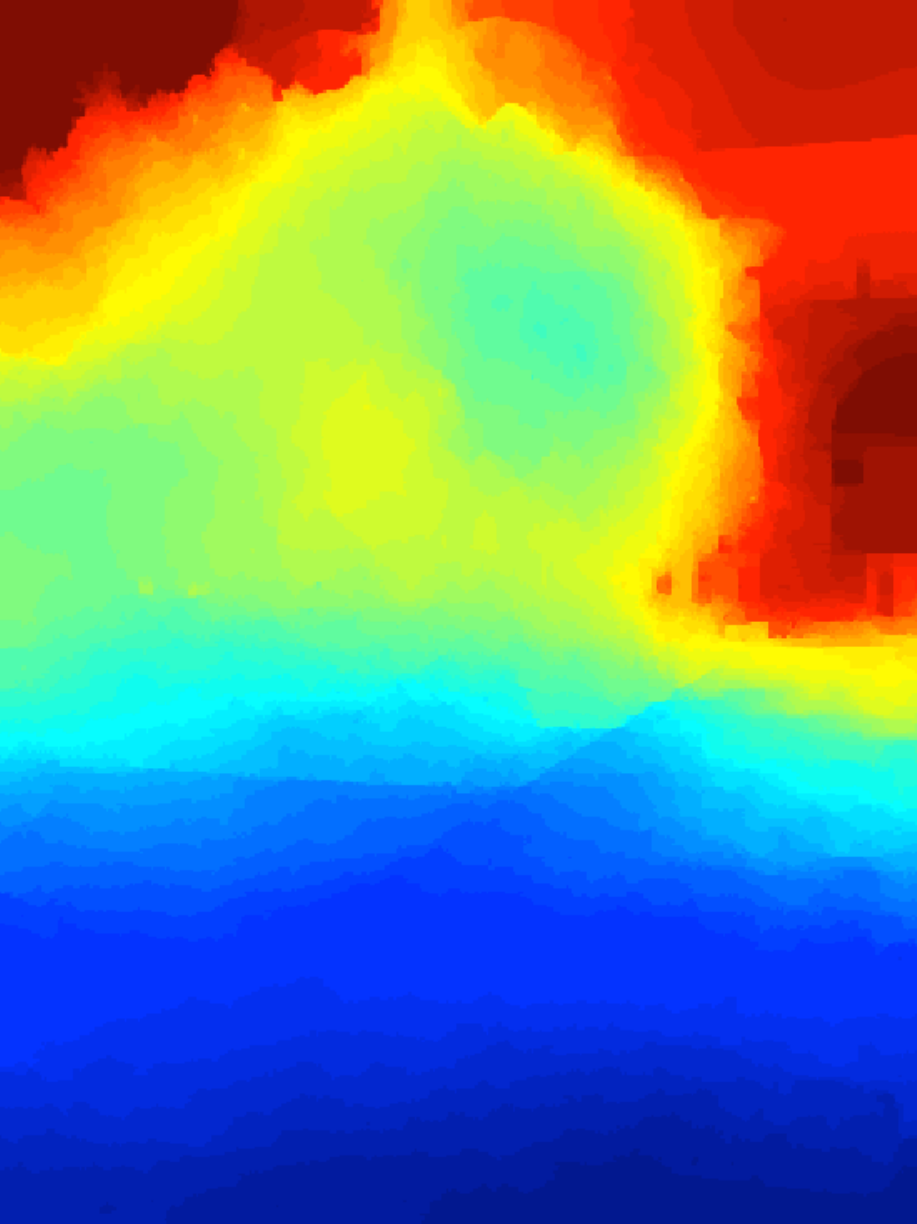}
	&\includegraphics[width=0.165\textwidth]{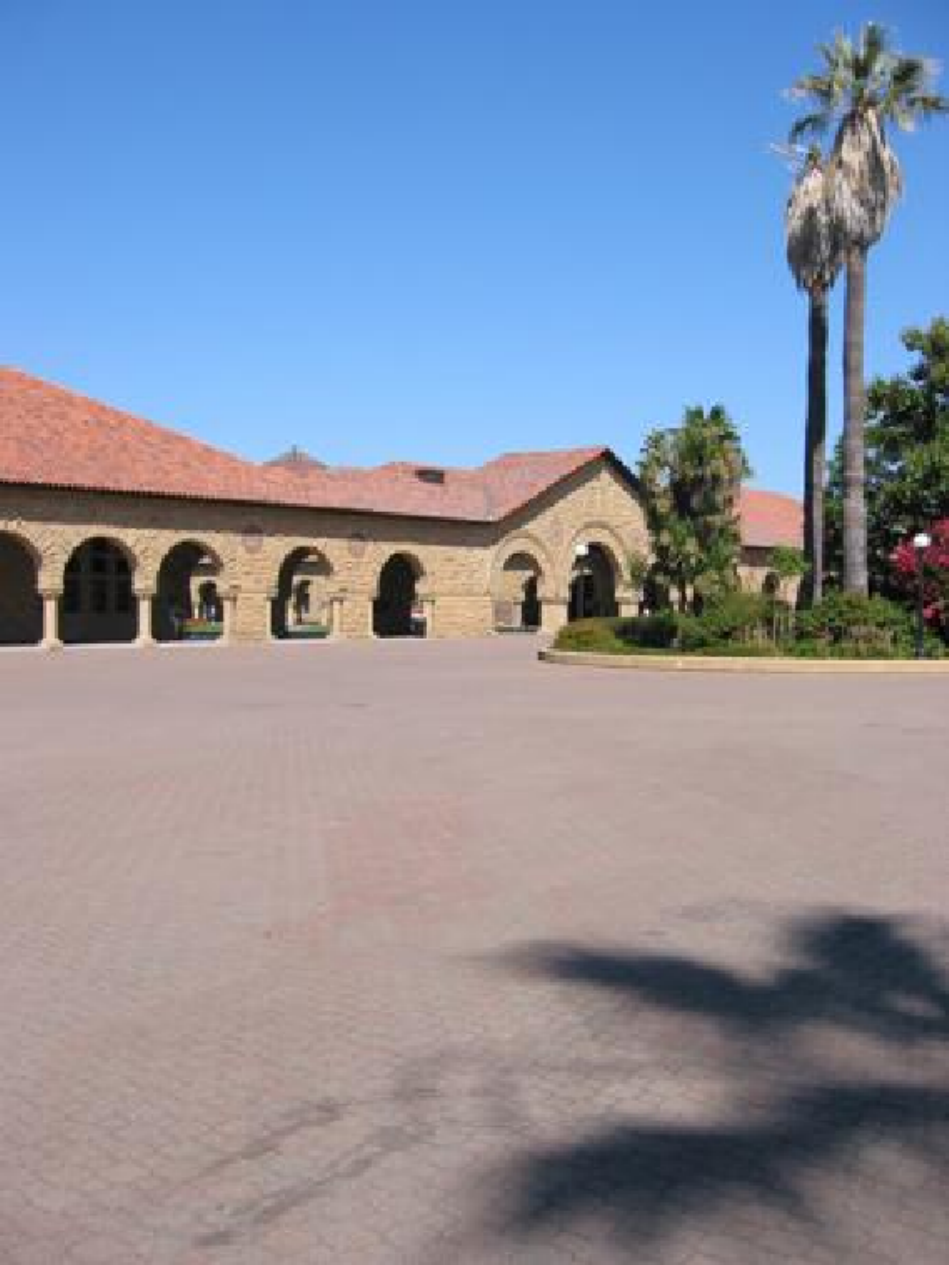}
	&\includegraphics[width=0.165\textwidth]{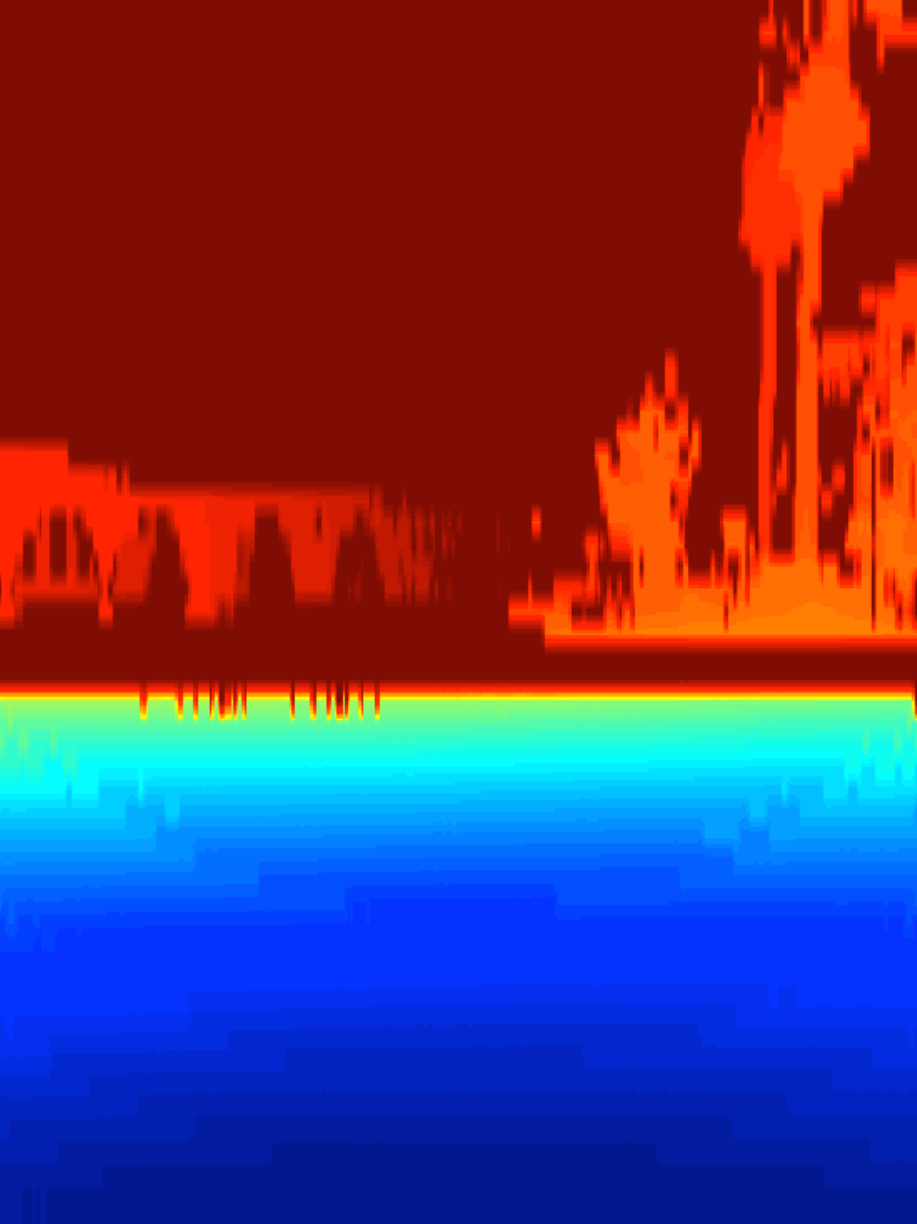}
	&\includegraphics[width=0.165\textwidth]{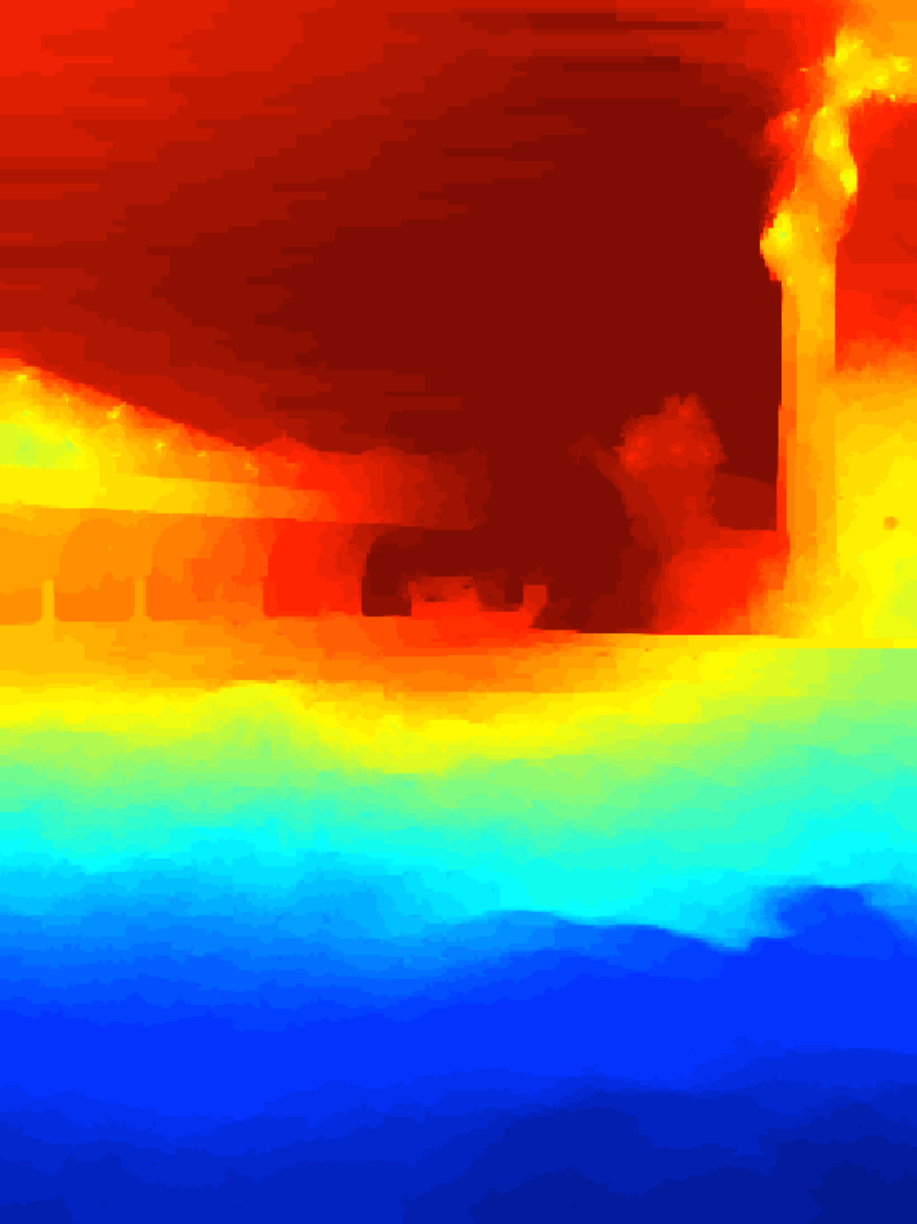} \\

  {\footnotesize Test image}  & {\footnotesize   Ground-truth }  &  {\footnotesize   \dcnfsp }   &  {\footnotesize  Test image}
                                                                           & {\footnotesize  Ground-truth }  & {\footnotesize  \dcnfsp}  \\
\end{tabular}
 }
\caption{Examples of depth predictions on the Make3D dataset (Best viewed on screen).
Depths are shown in log scale and in color (red is far, blue is close).
}  \label{fig:nmake3d}
\end{figure*}

\begin{figure*} [t] \center
{
	\includegraphics[width=0.28\textwidth]{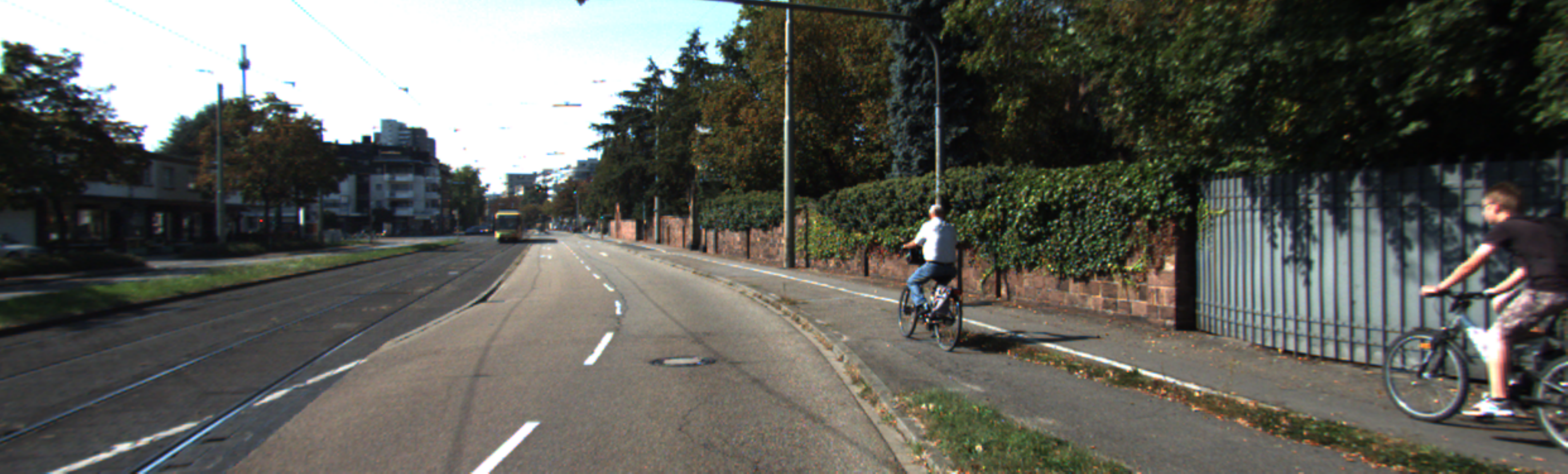}
	\includegraphics[width=0.28\textwidth]{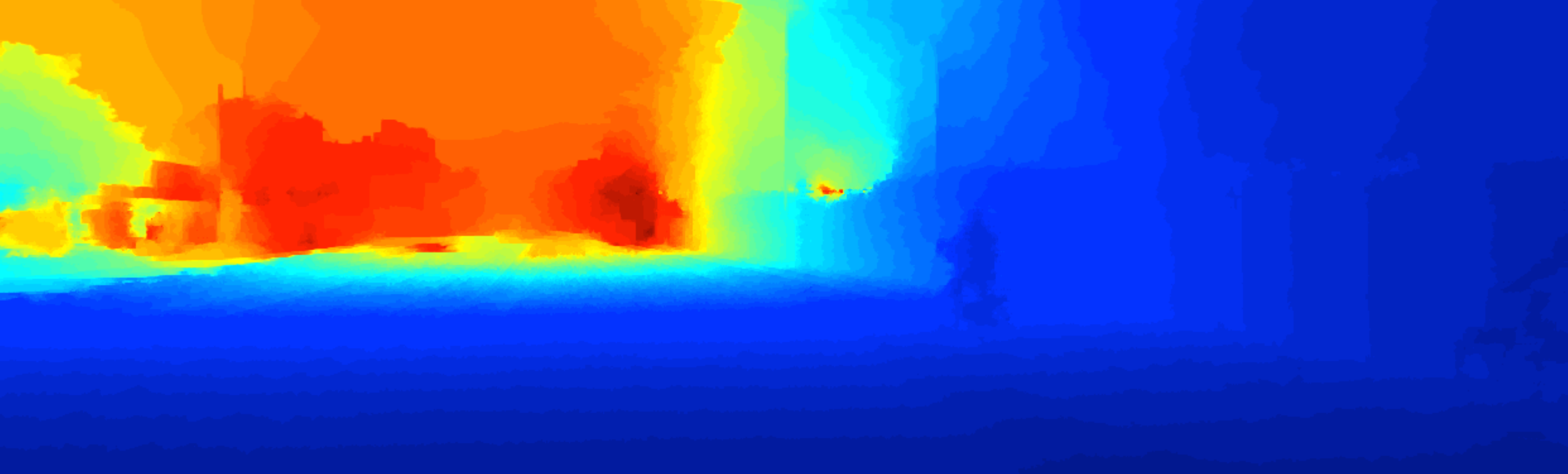}
	\includegraphics[width=0.28\textwidth]{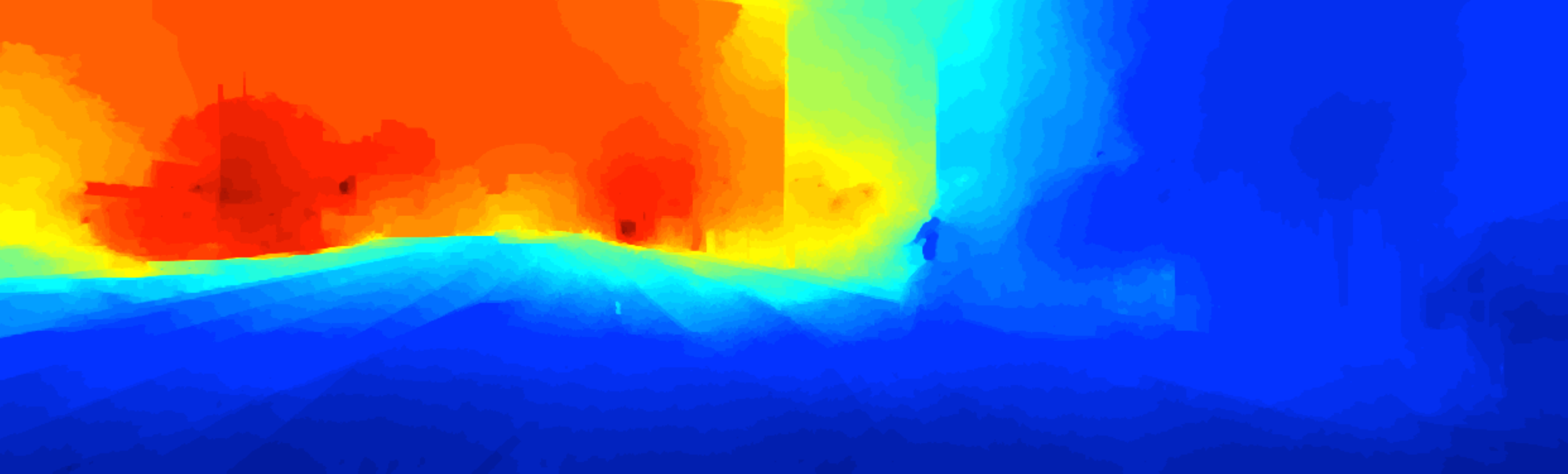}\\

	\includegraphics[width=0.28\textwidth]{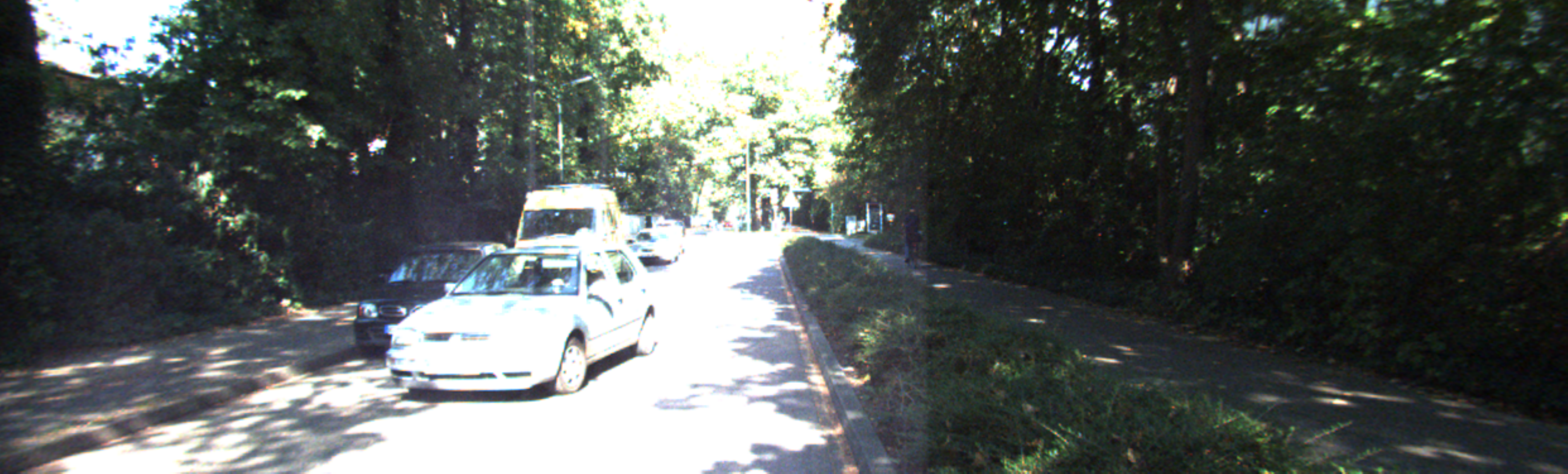}
	\includegraphics[width=0.28\textwidth]{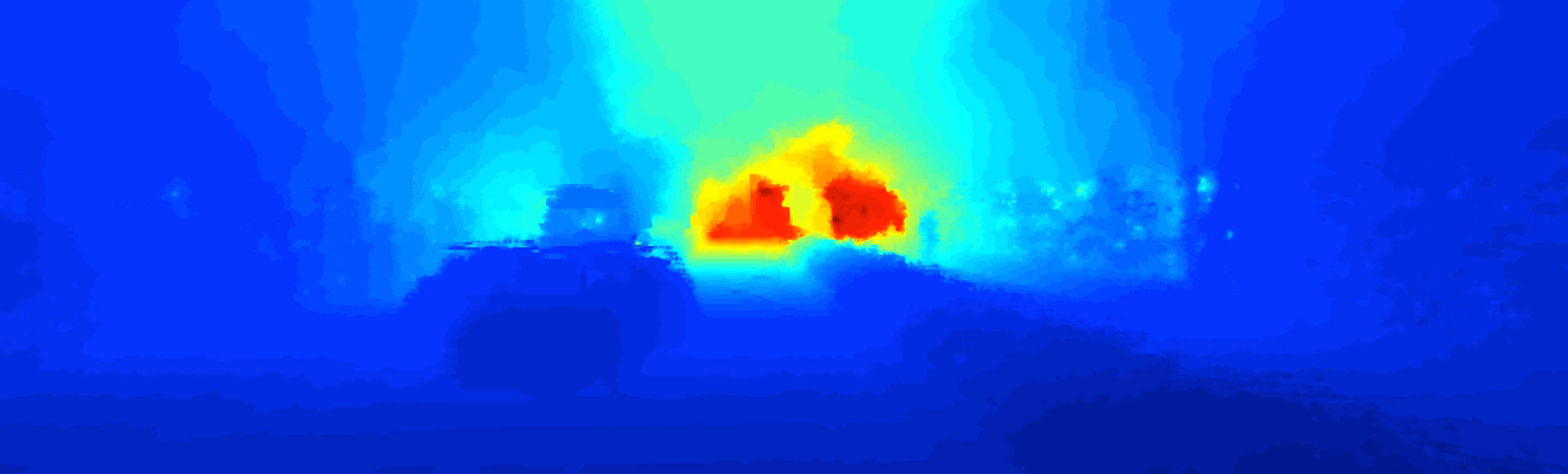}
	\includegraphics[width=0.28\textwidth]{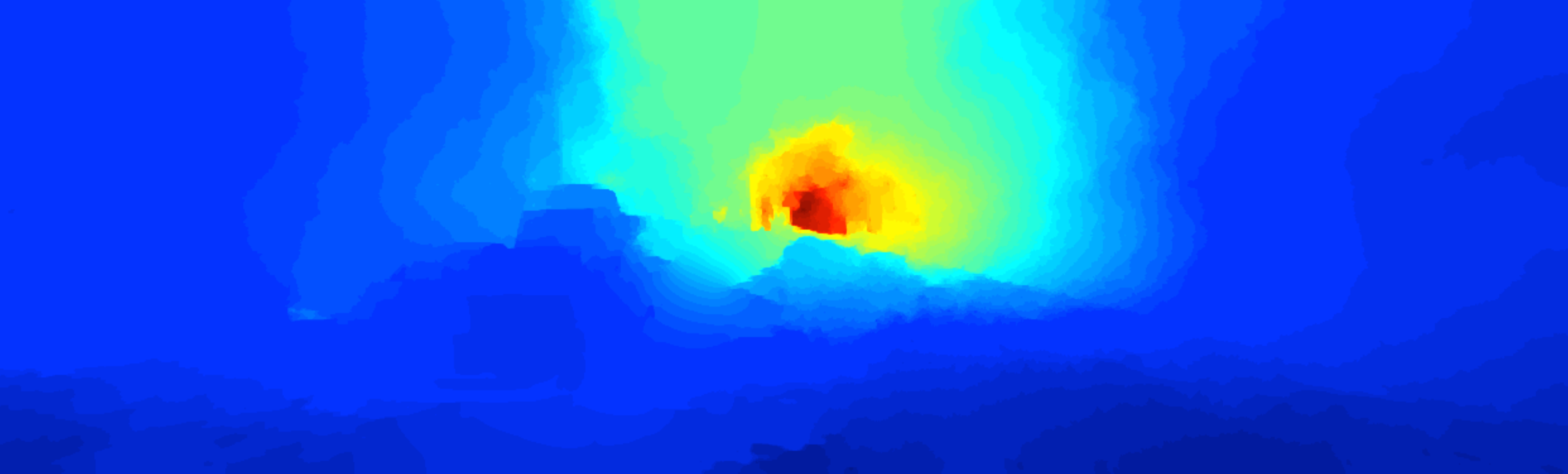}\\

	\includegraphics[width=0.28\textwidth]{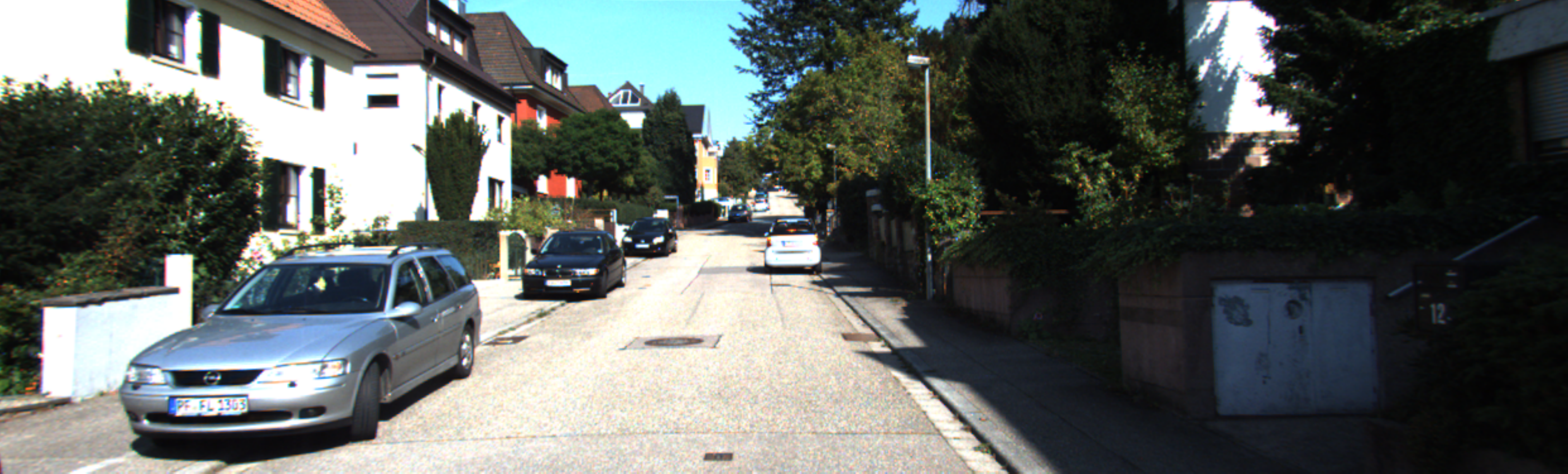}
	\includegraphics[width=0.28\textwidth]{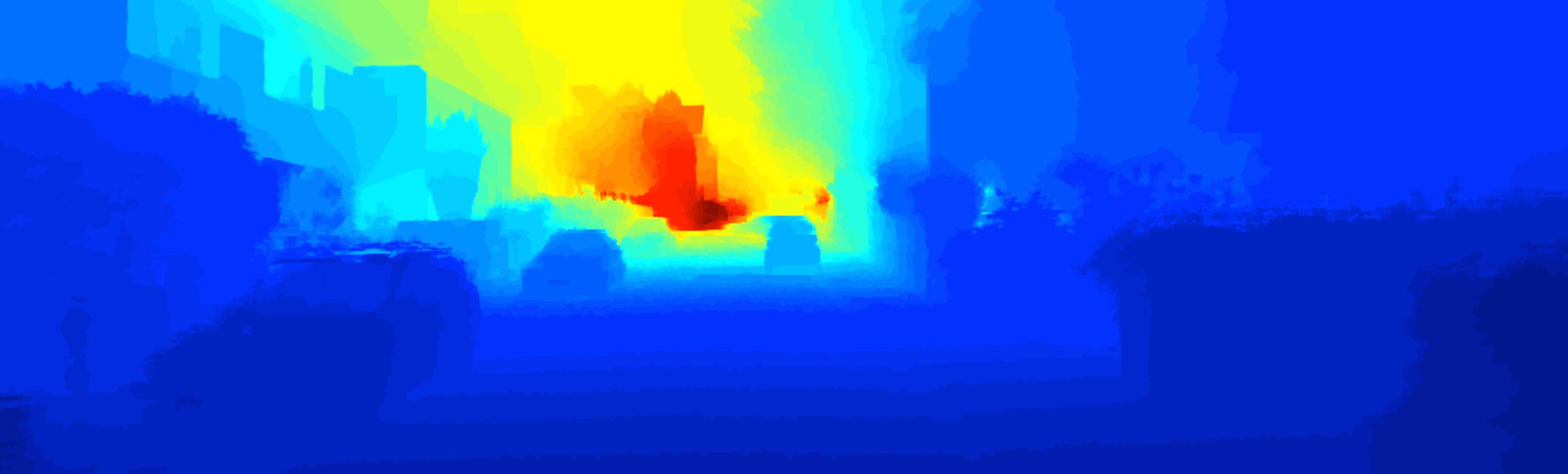}
	\includegraphics[width=0.28\textwidth]{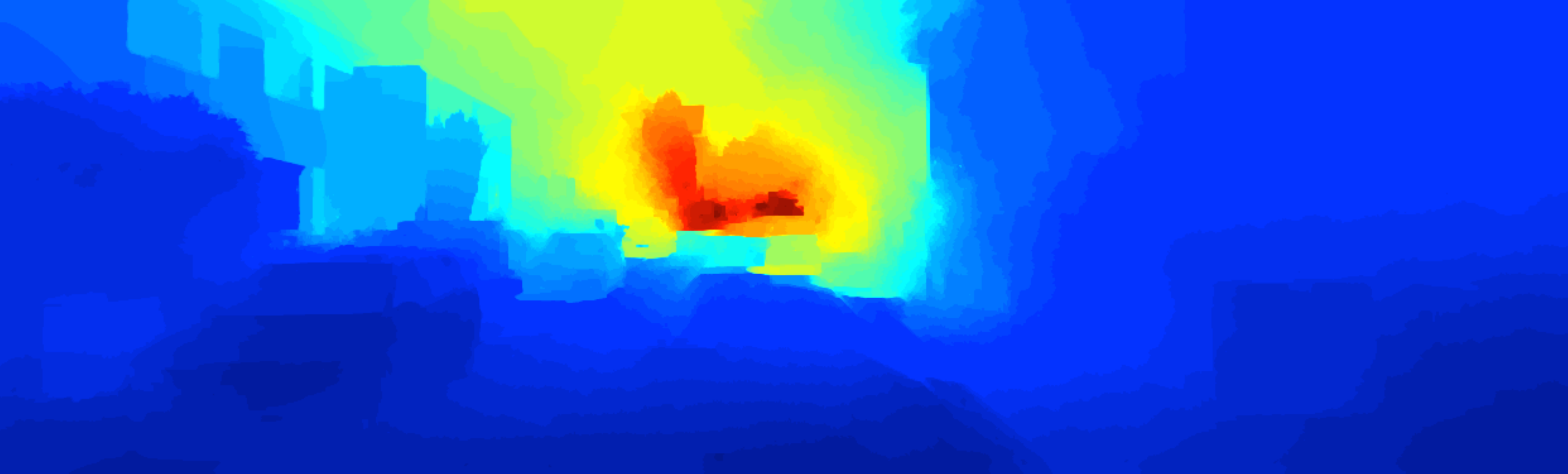}\\

	\includegraphics[width=0.28\textwidth]{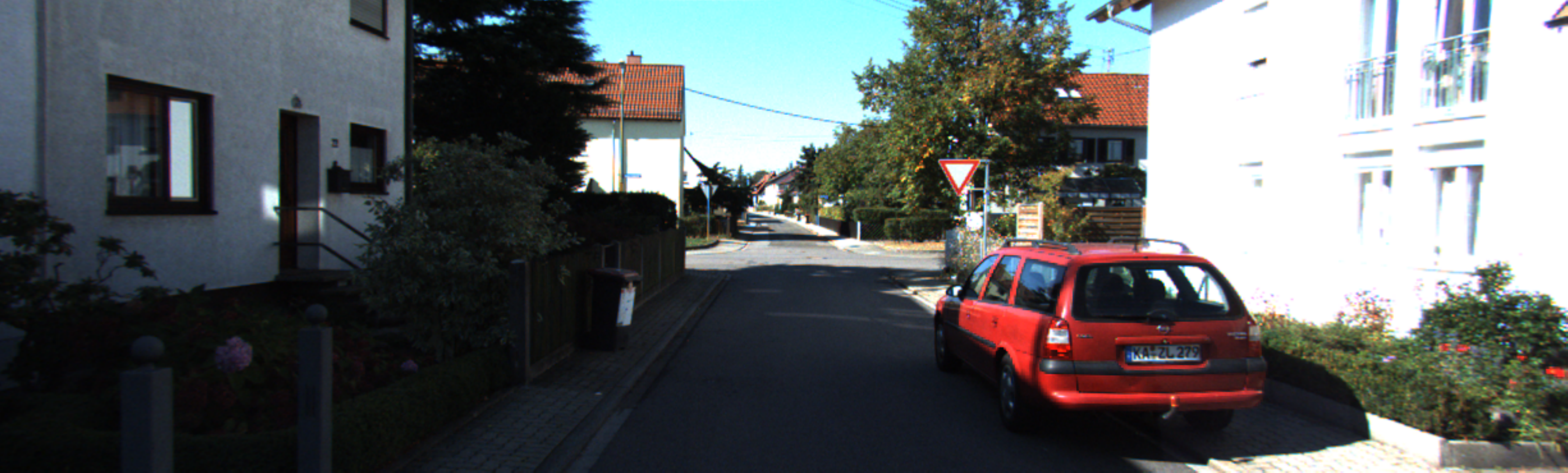}
	\includegraphics[width=0.28\textwidth]{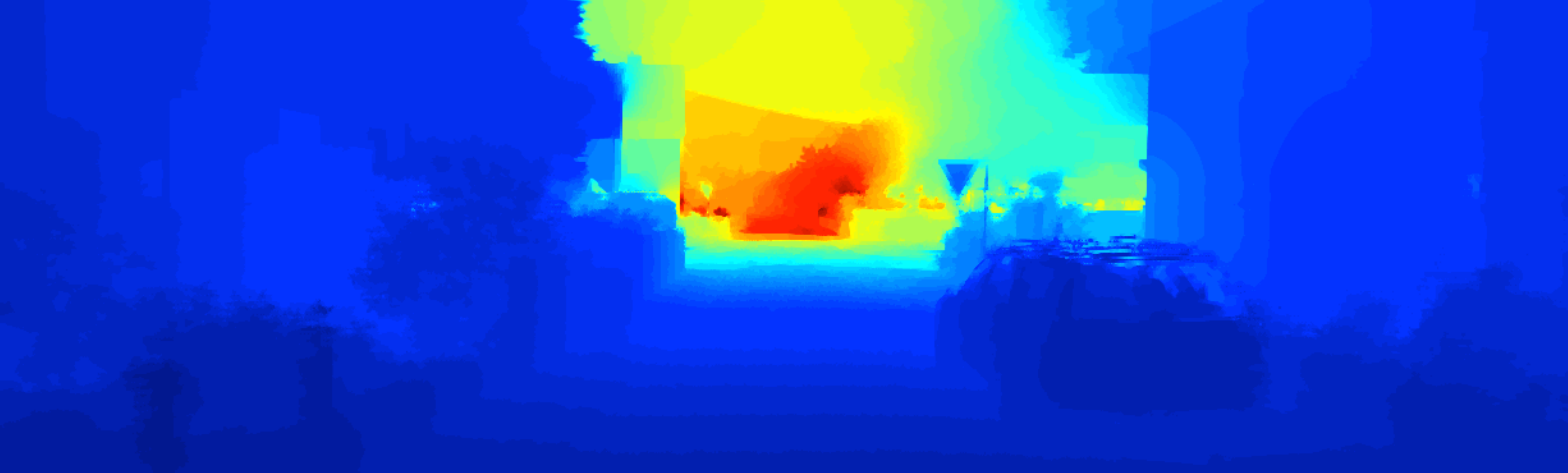}
	\includegraphics[width=0.28\textwidth]{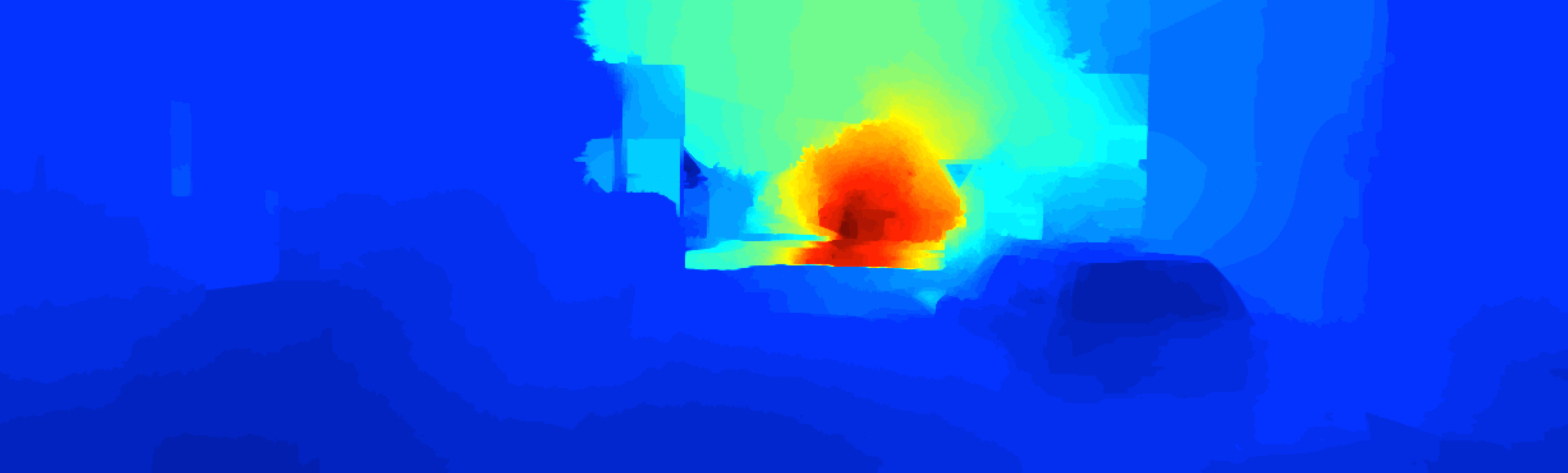}\\

	\includegraphics[width=0.28\textwidth]{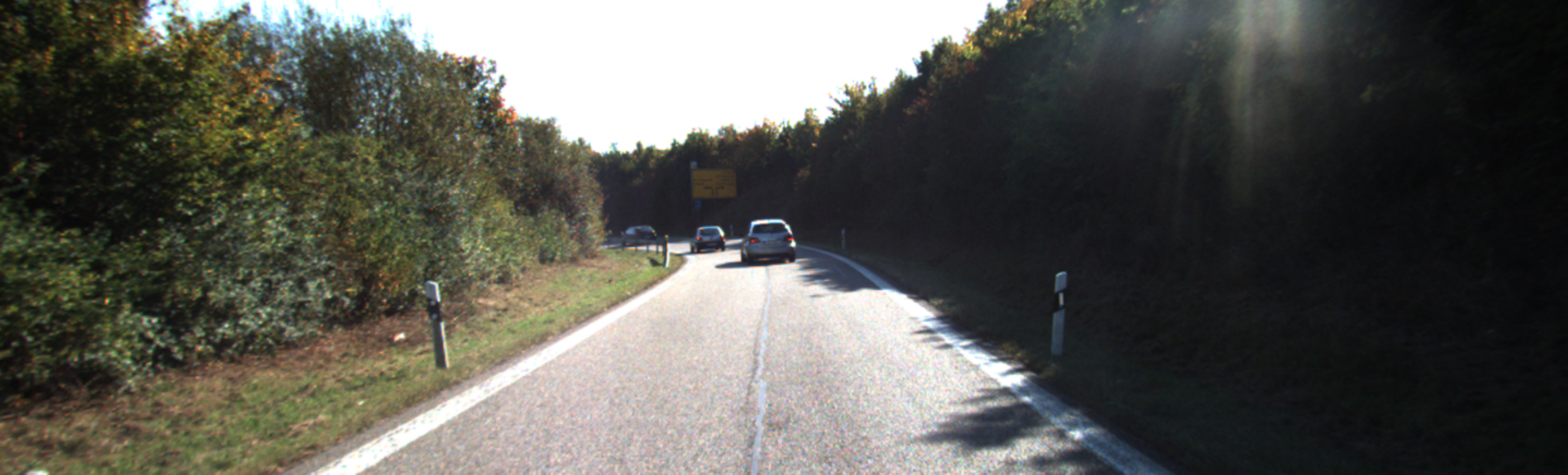}
	\includegraphics[width=0.28\textwidth]{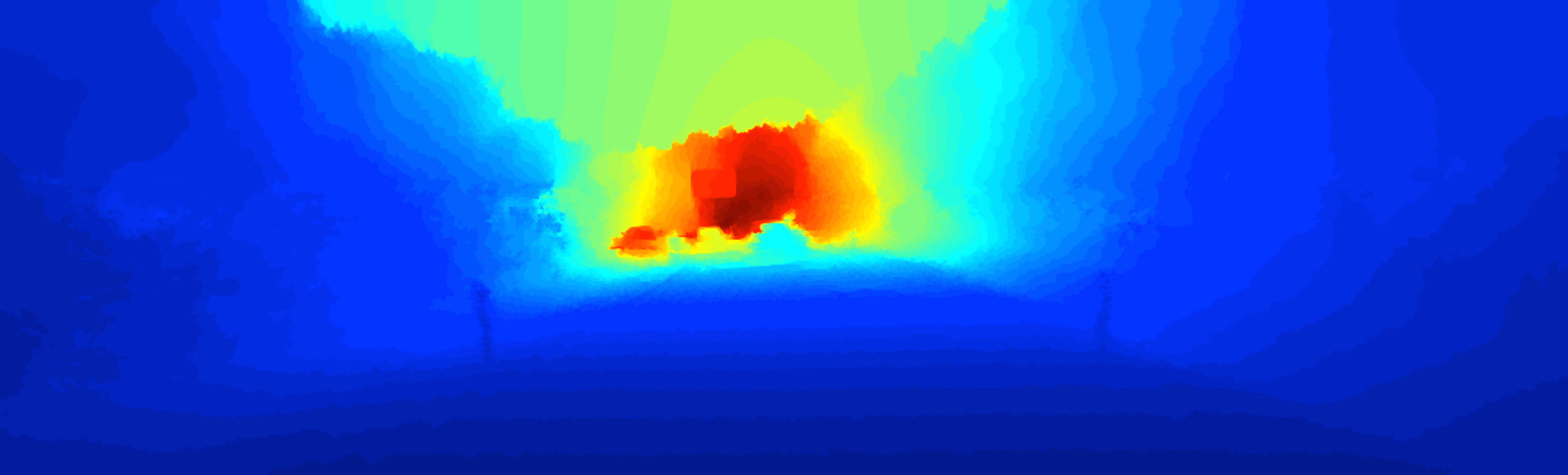}
	\includegraphics[width=0.28\textwidth]{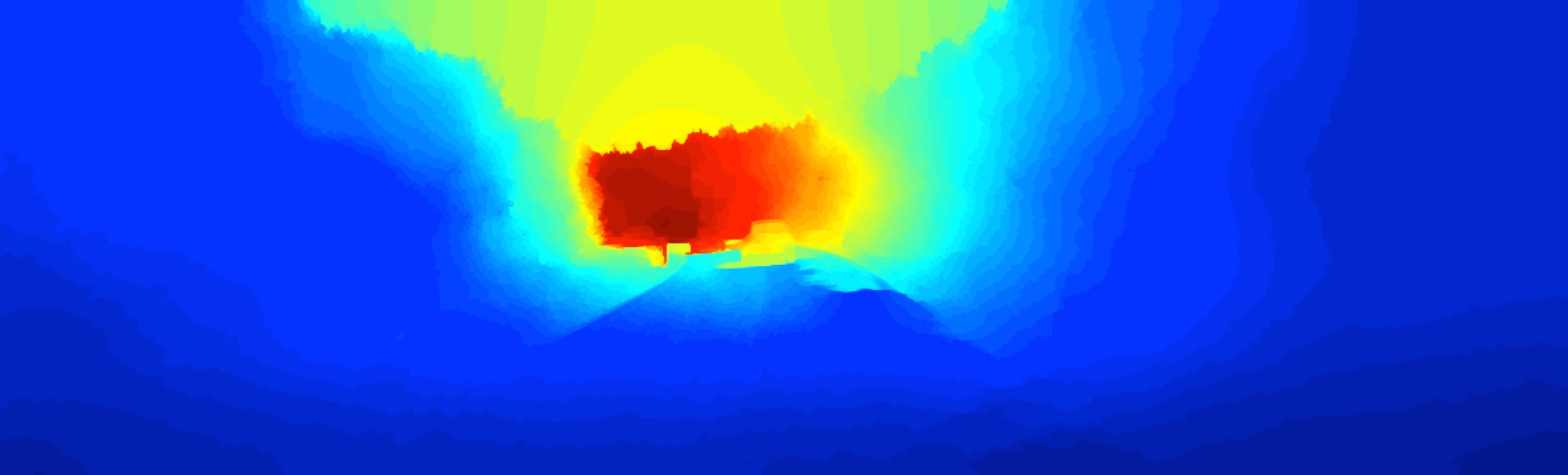}\\

	\includegraphics[width=0.28\textwidth]{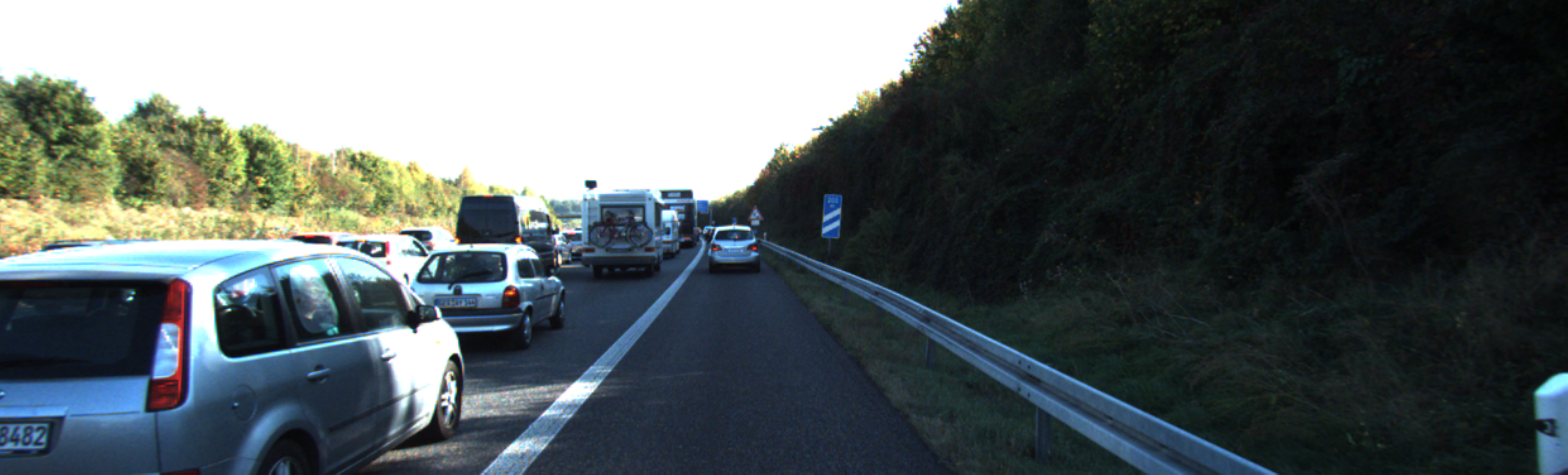}
	\includegraphics[width=0.28\textwidth]{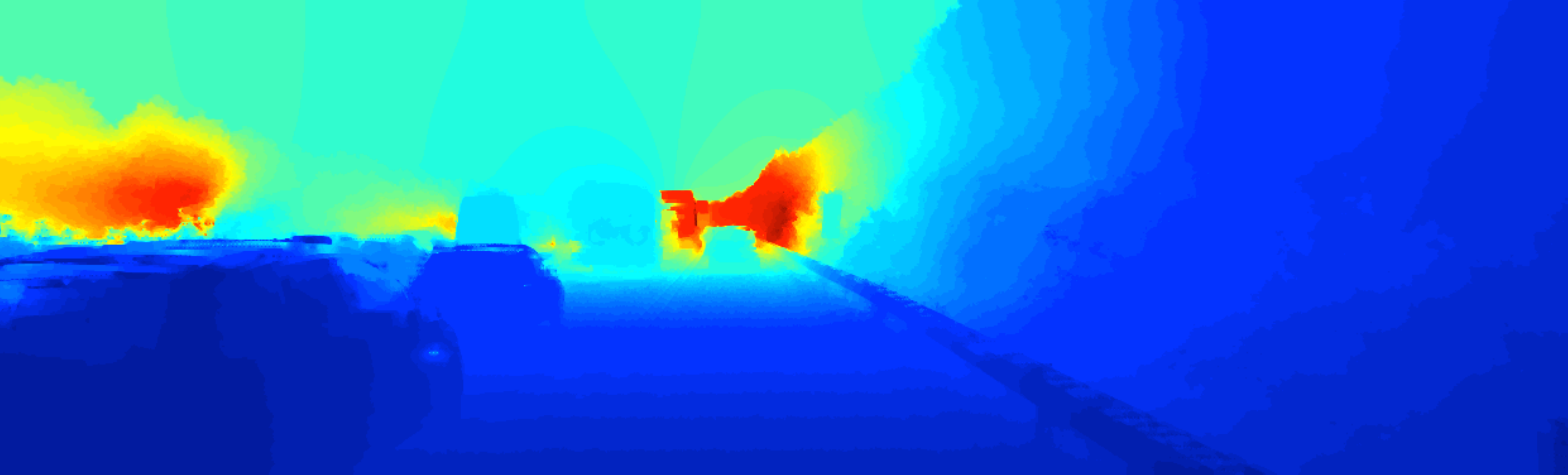}
	\includegraphics[width=0.28\textwidth]{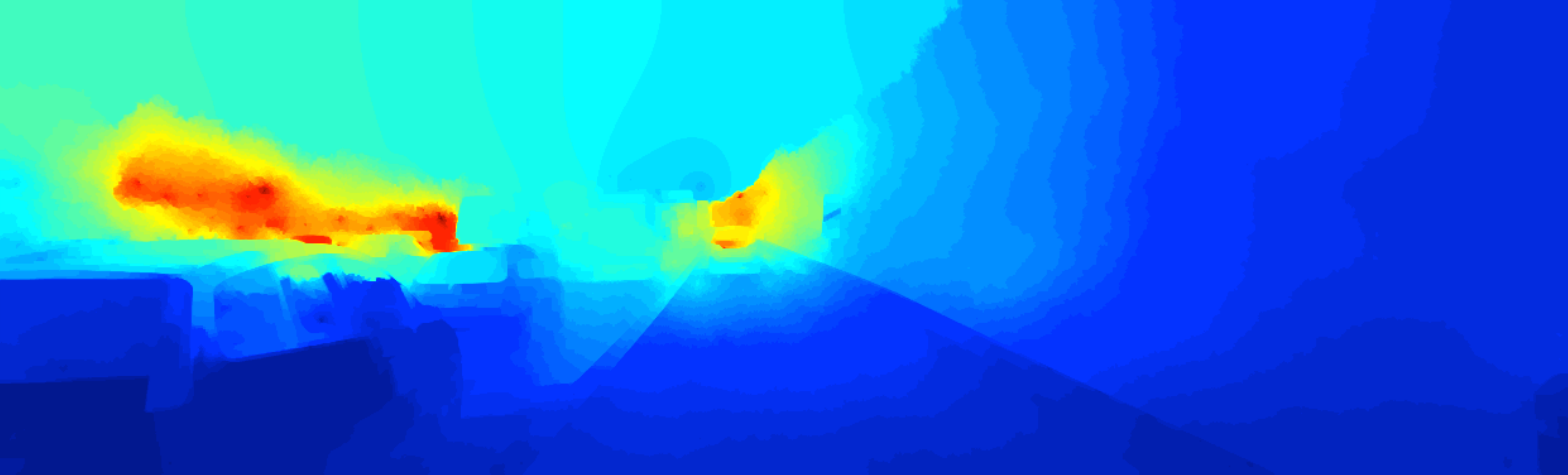}\\

\footnotesize{
Test image \quad\quad\quad\quad\quad\quad\quad\quad\quad\quad\quad\quad
 Ground-truth \quad\quad\quad\quad\quad\quad\quad\quad\quad\quad\quad\quad \dcnfsp
 }
 \\
}
\caption{Examples of depth predictions on the KITTI dataset (Best viewed on screen).
Depths are shown in log scale and in color (red is far, blue is close).
}  \label{fig:kitti}
\end{figure*}

\begin{figure*} [t] \center
\resizebox{.98\linewidth}{!} {
\begin{tabular}{ccc|ccc}
	\includegraphics[width=0.155\textwidth, height=0.12\textwidth]{./fig/NYUD2/images/1.pdf}
	&\includegraphics[width=0.155\textwidth, height=0.12\textwidth]{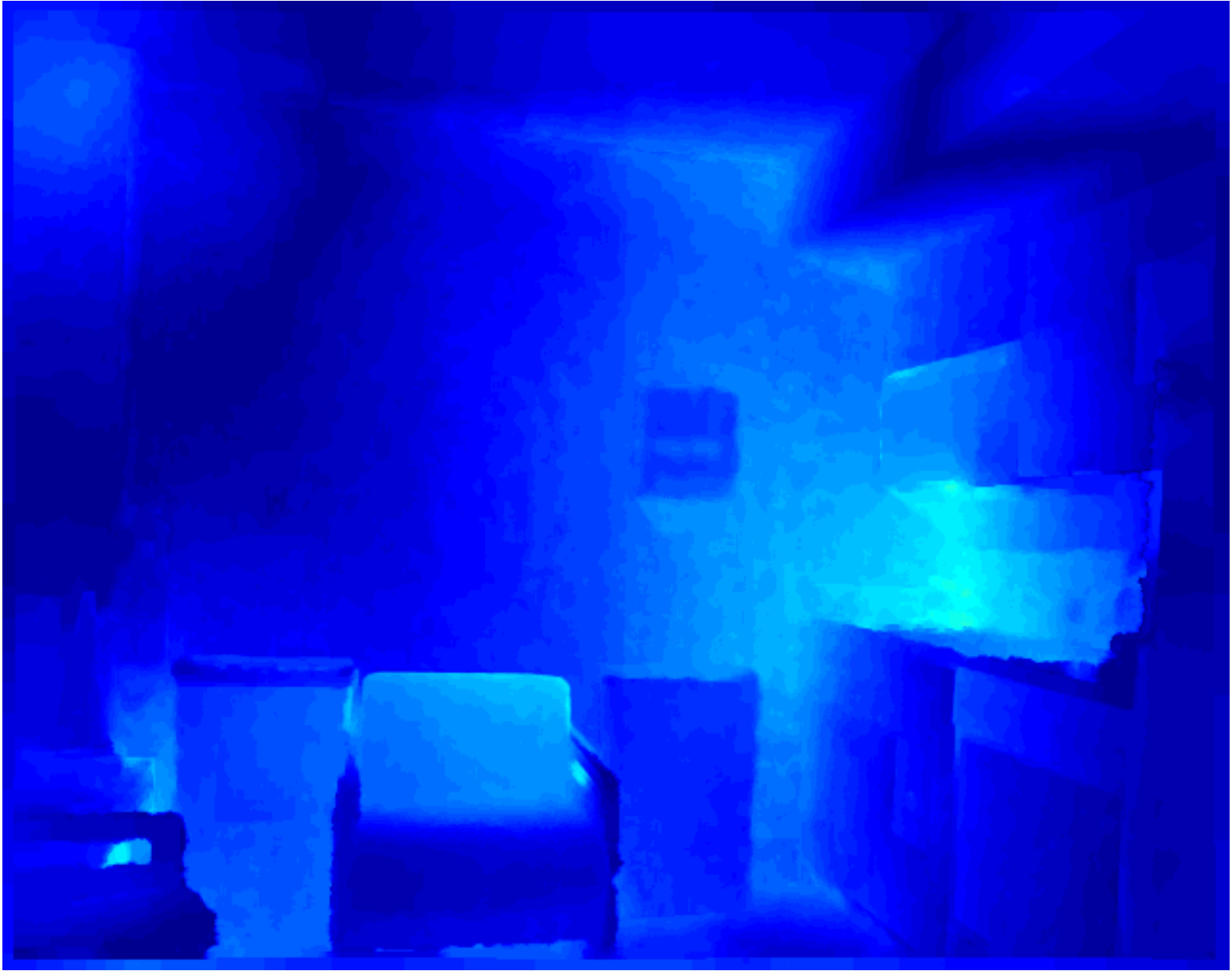}
	&\includegraphics[width=0.18\textwidth, height=0.12\textwidth]{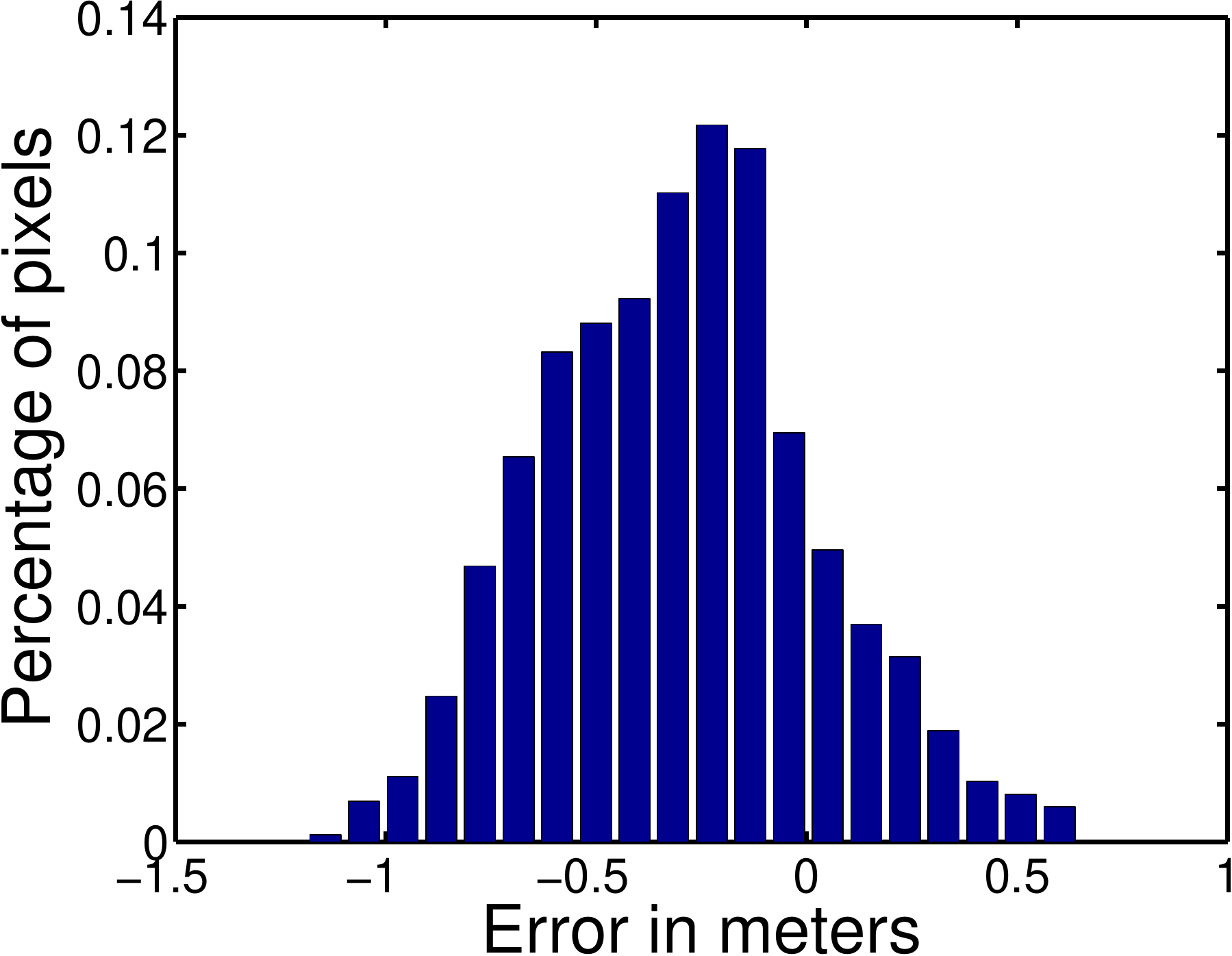}
	&\includegraphics[width=0.1\textwidth, height=0.12\textwidth]{./fig/Make3D/images/403.pdf}
	&\includegraphics[width=0.1\textwidth, height=0.12\textwidth]{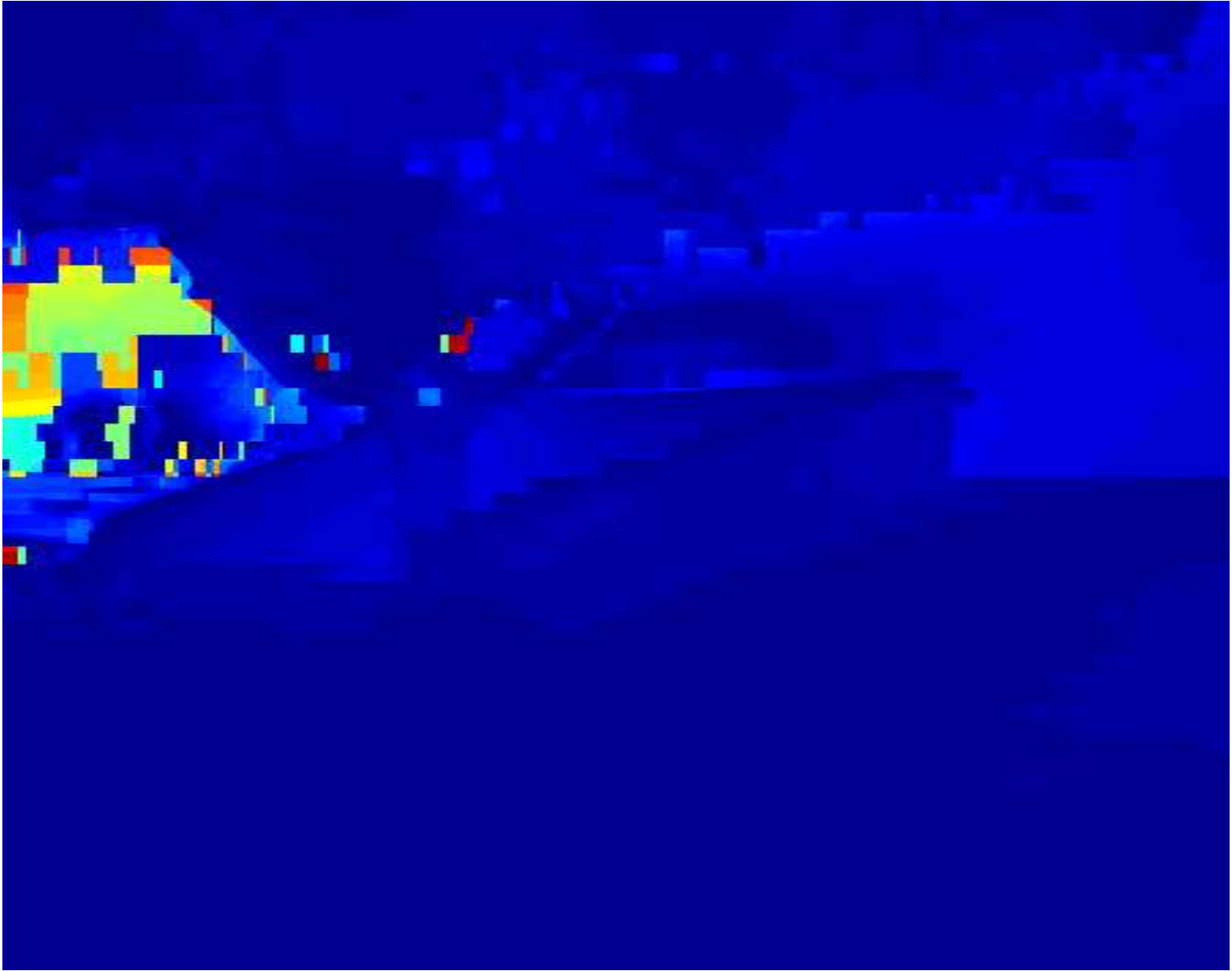}
	&\includegraphics[width=0.18\textwidth, height=0.12\textwidth]{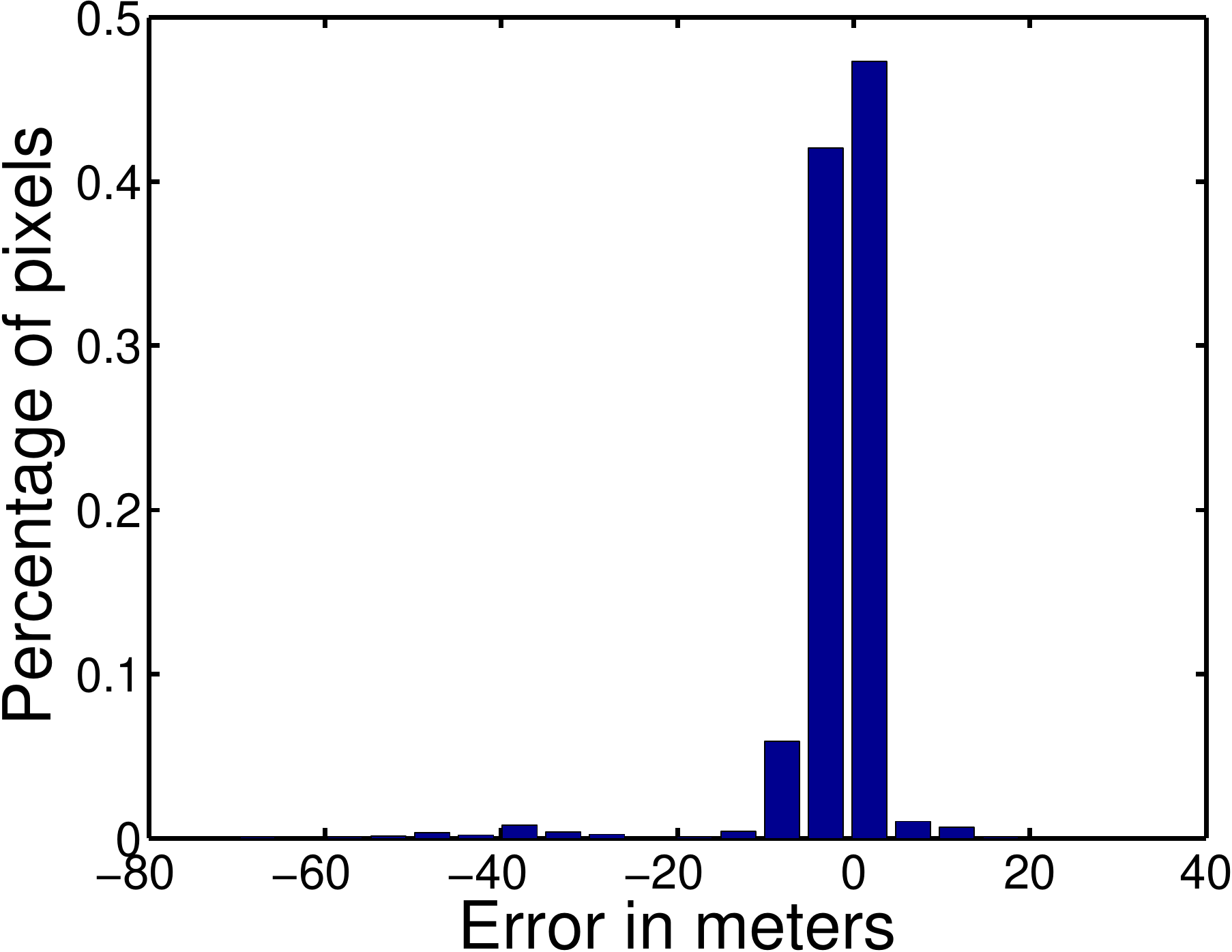} \\

	\includegraphics[width=0.155\textwidth, height=0.12\textwidth]{./fig/NYUD2/images/1167.pdf}
	&\includegraphics[width=0.155\textwidth, height=0.12\textwidth]{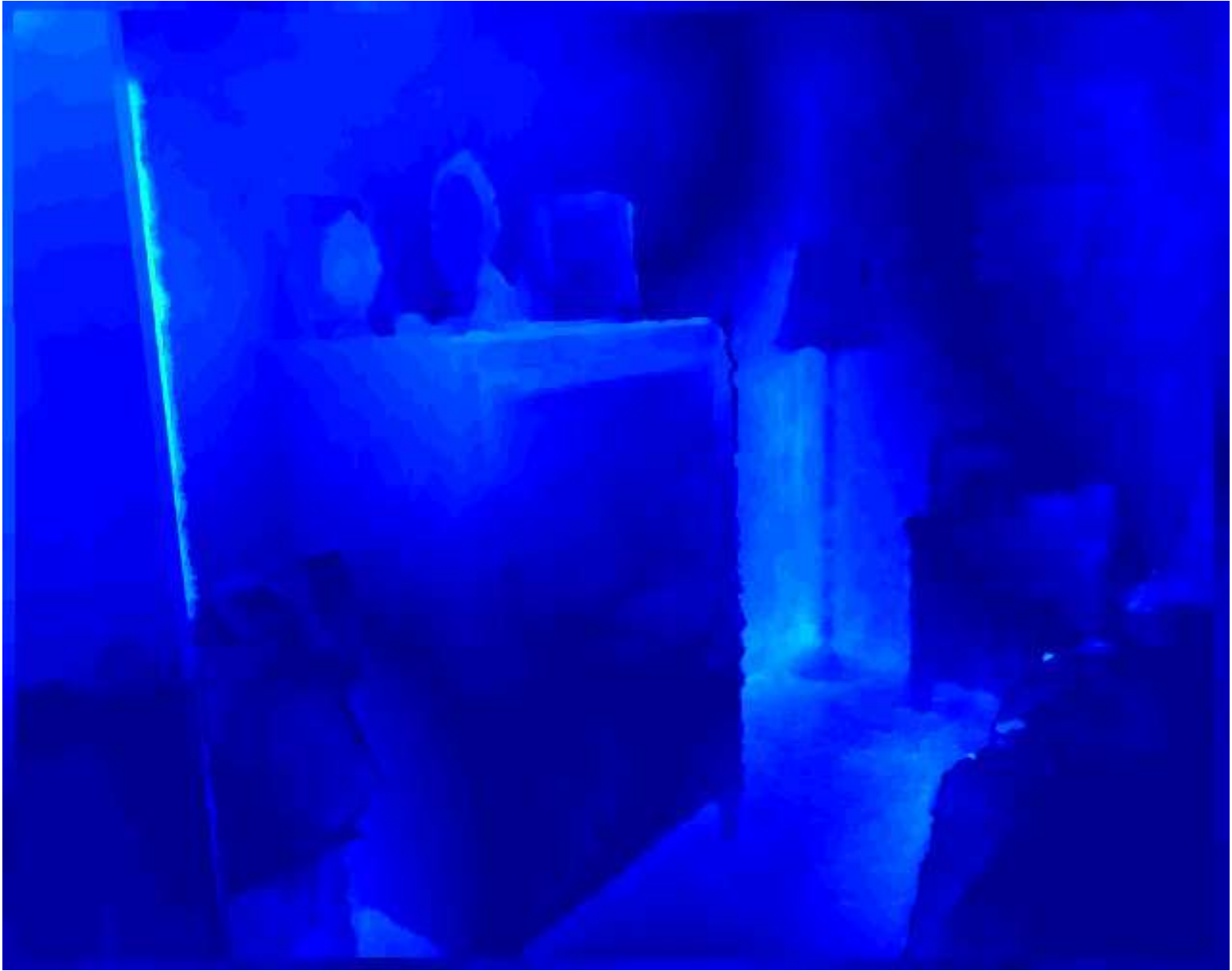}
	&\includegraphics[width=0.18\textwidth, height=0.12\textwidth]{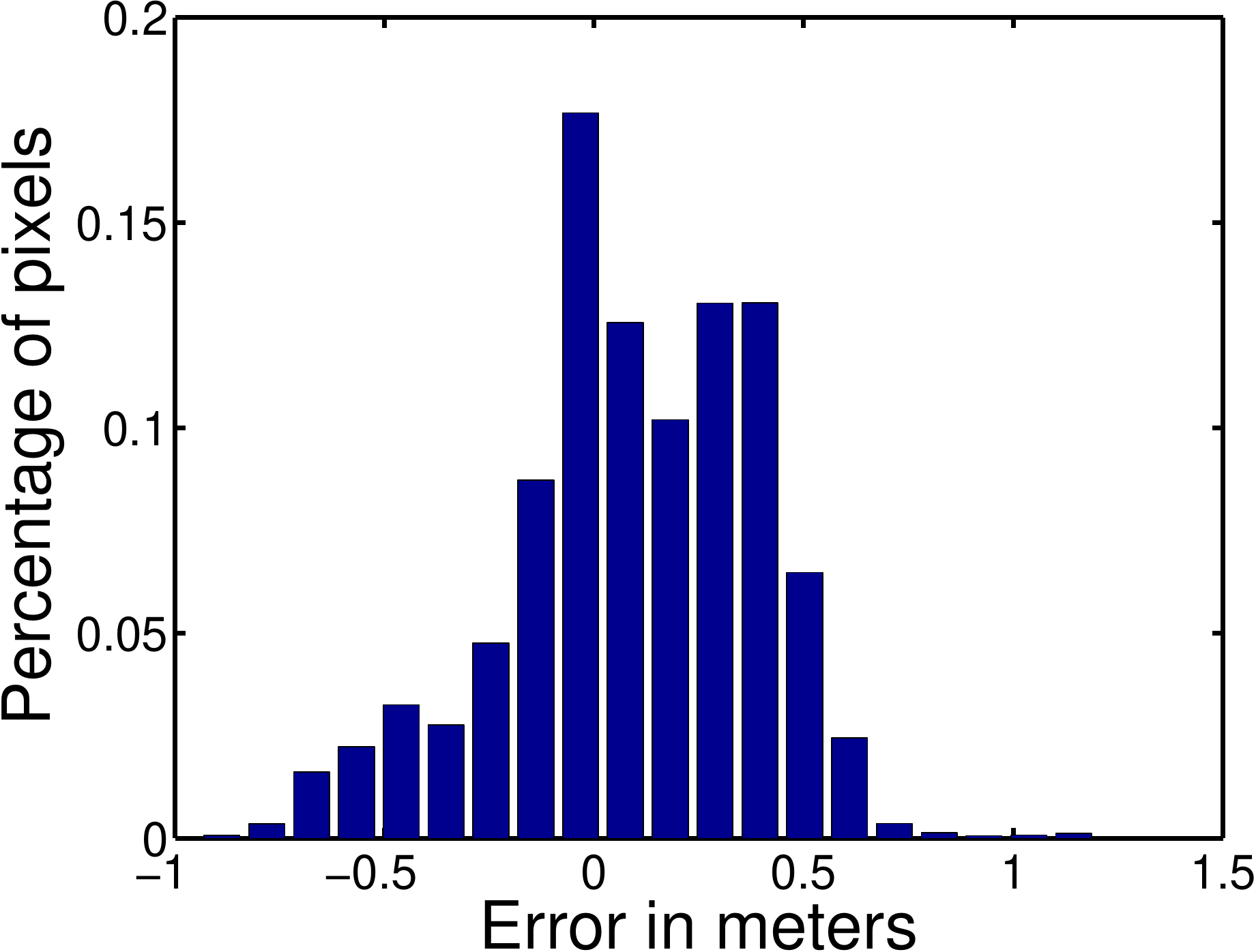}
	&\includegraphics[width=0.1\textwidth, height=0.12\textwidth]{./fig/Make3D/images/407.pdf}
	&\includegraphics[width=0.1\textwidth, height=0.12\textwidth]{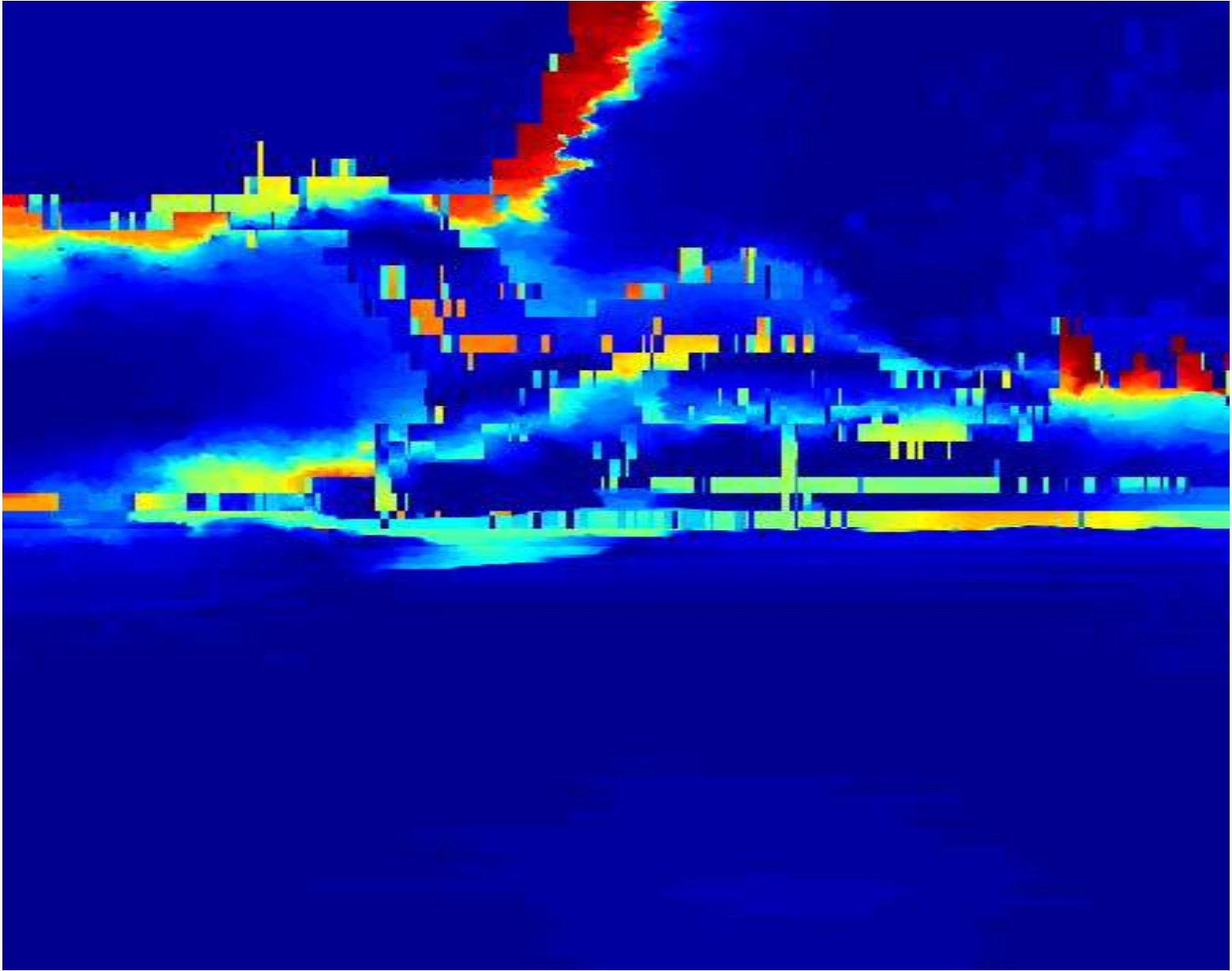}
	&\includegraphics[width=0.18\textwidth, height=0.12\textwidth]{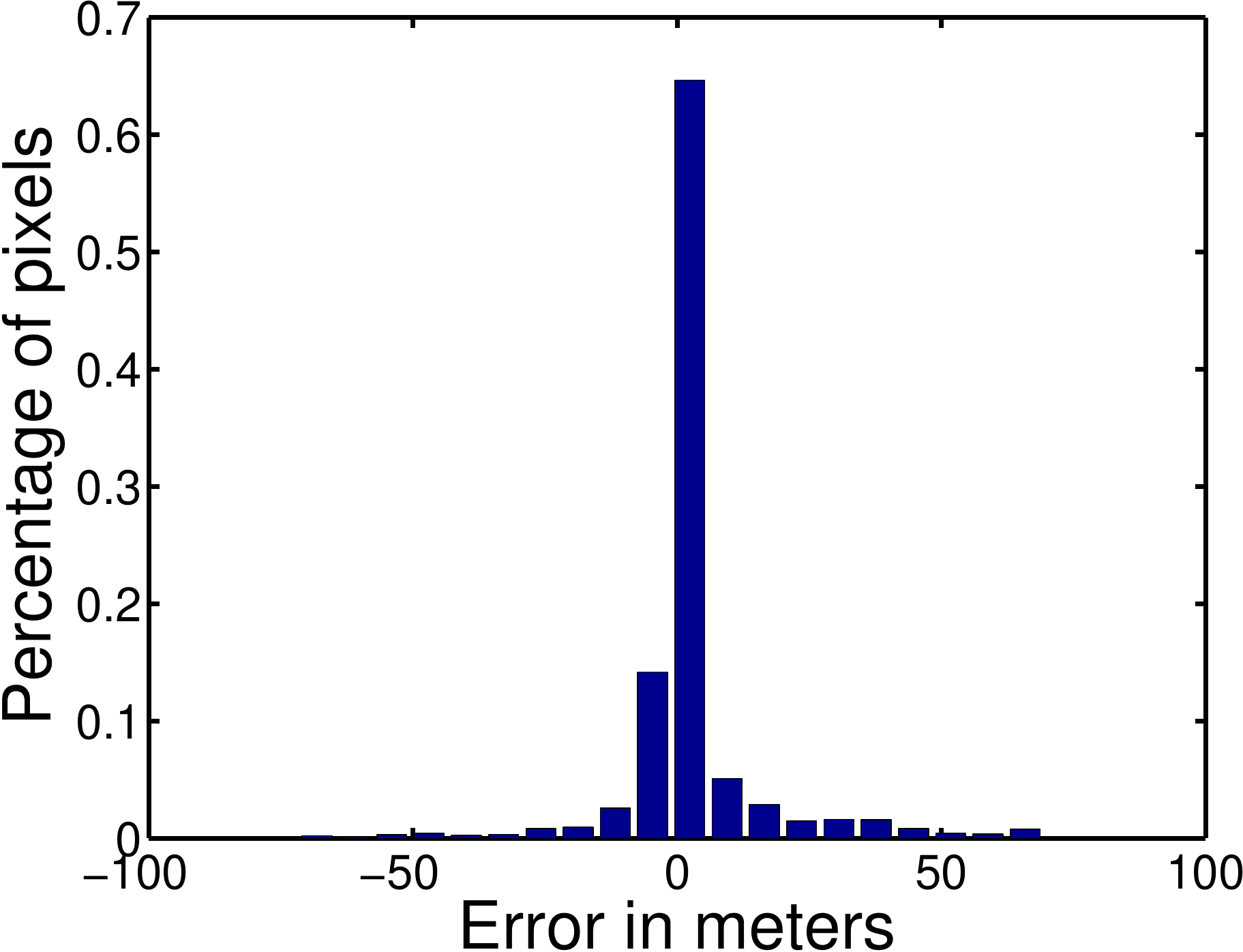}\\

	\includegraphics[width=0.155\textwidth, height=0.12\textwidth]{./fig/NYUD2/images/300.pdf}
	&\includegraphics[width=0.155\textwidth, height=0.12\textwidth]{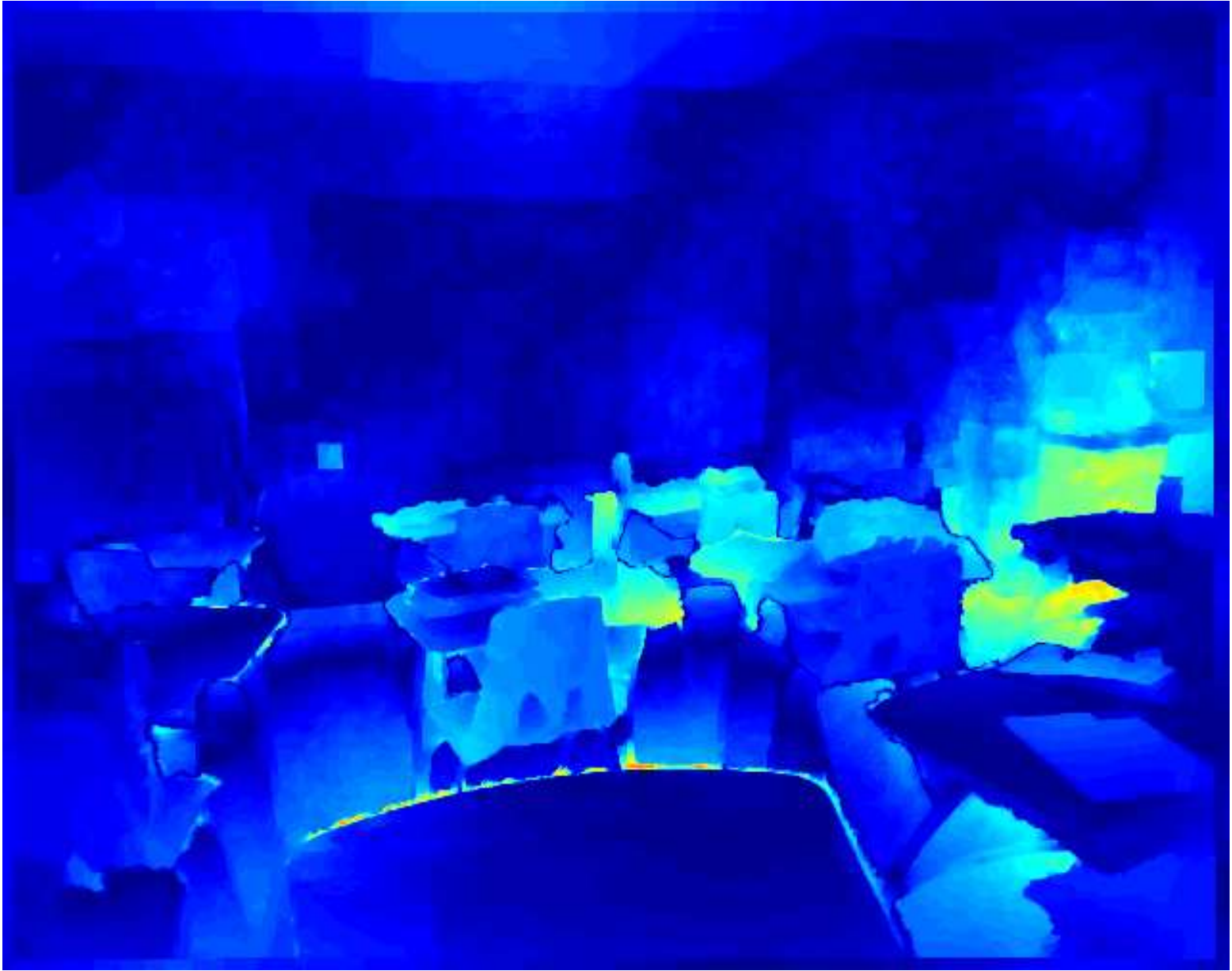}
	&\includegraphics[width=0.18\textwidth, height=0.12\textwidth]{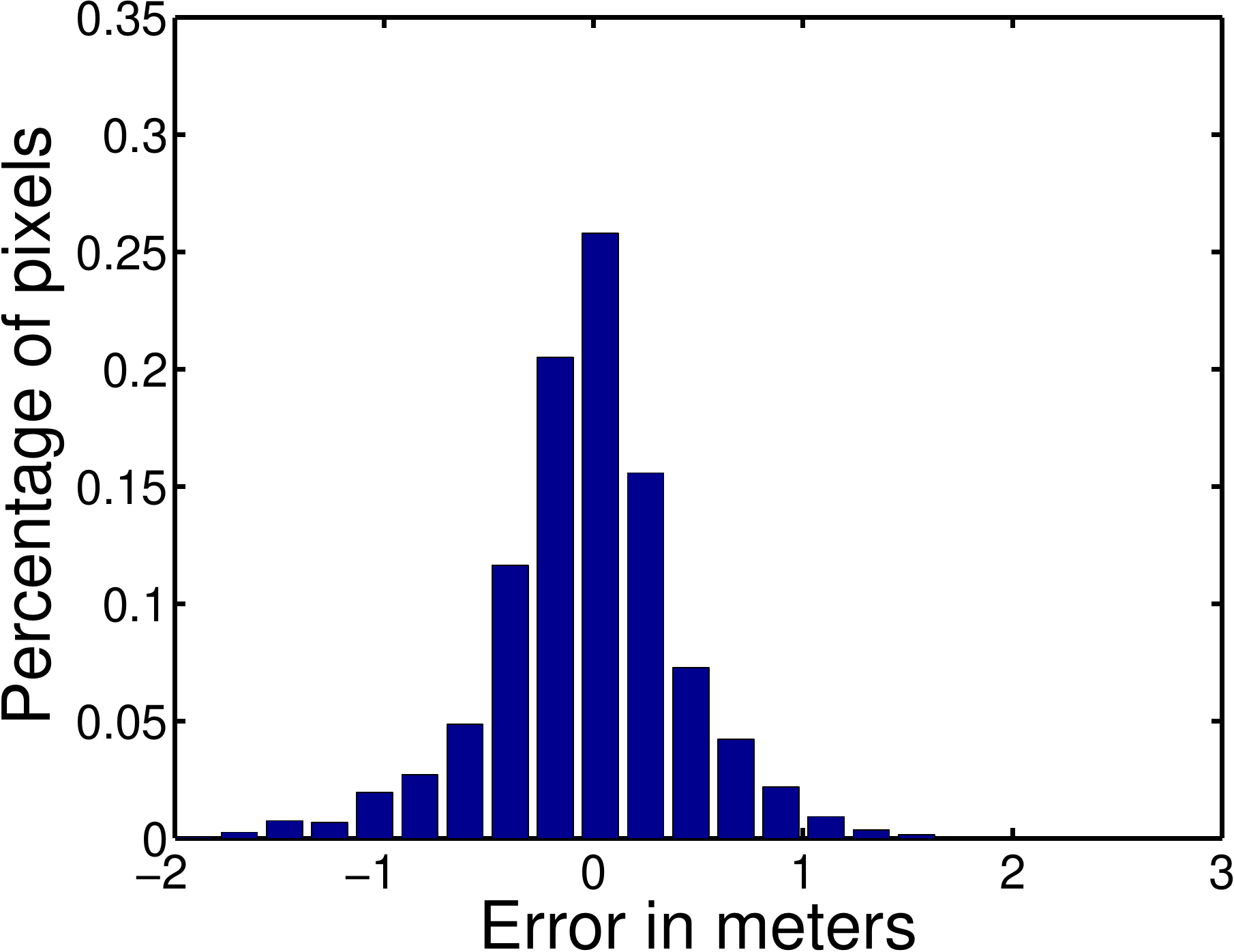}
	&\includegraphics[width=0.1\textwidth, height=0.12\textwidth]{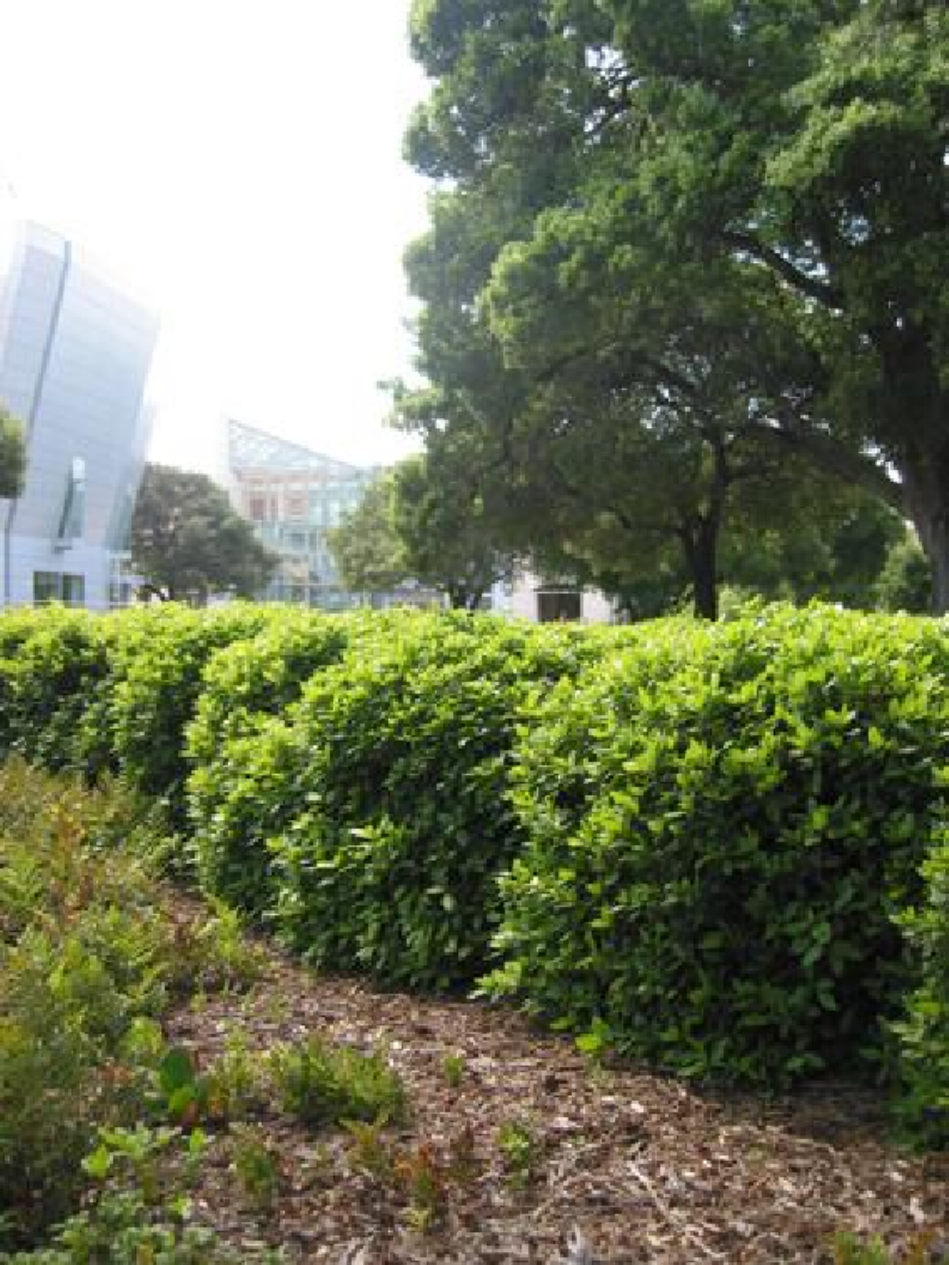}
	&\includegraphics[width=0.1\textwidth, height=0.12\textwidth]{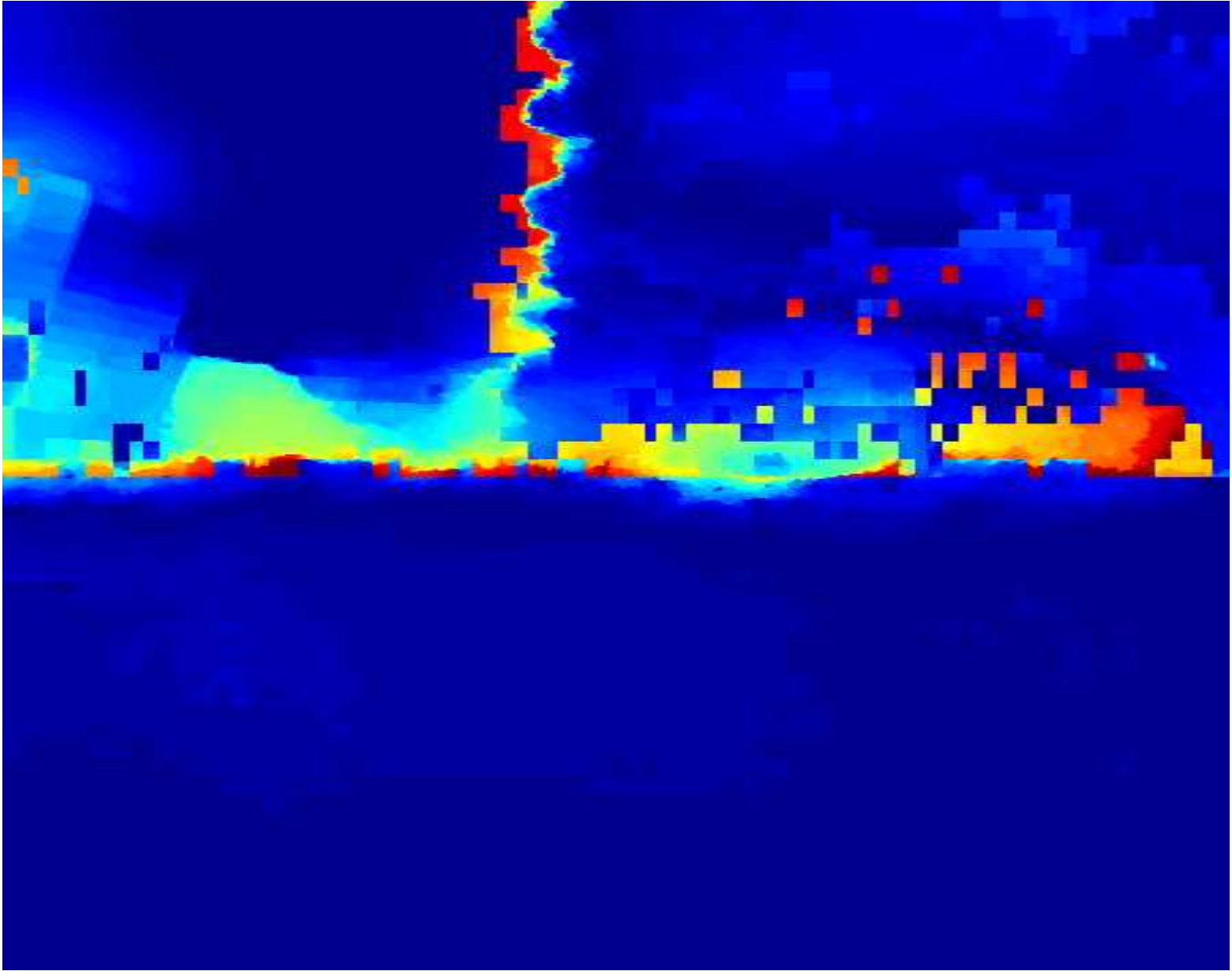}
	&\includegraphics[width=0.18\textwidth, height=0.12\textwidth]{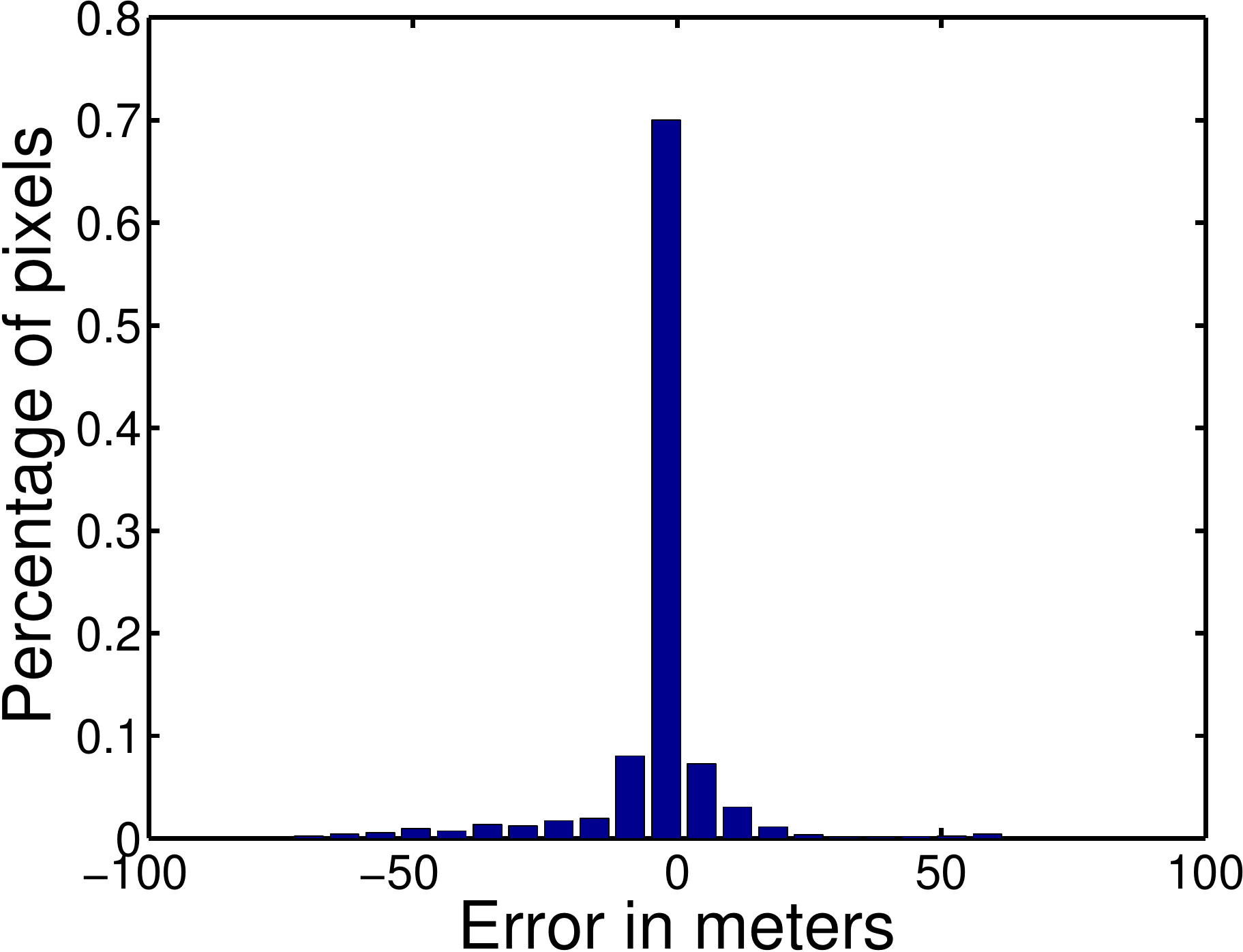}\\

	\includegraphics[width=0.155\textwidth, height=0.12\textwidth]{./fig/NYUD2/images/470.pdf}
	&\includegraphics[width=0.155\textwidth, height=0.12\textwidth]{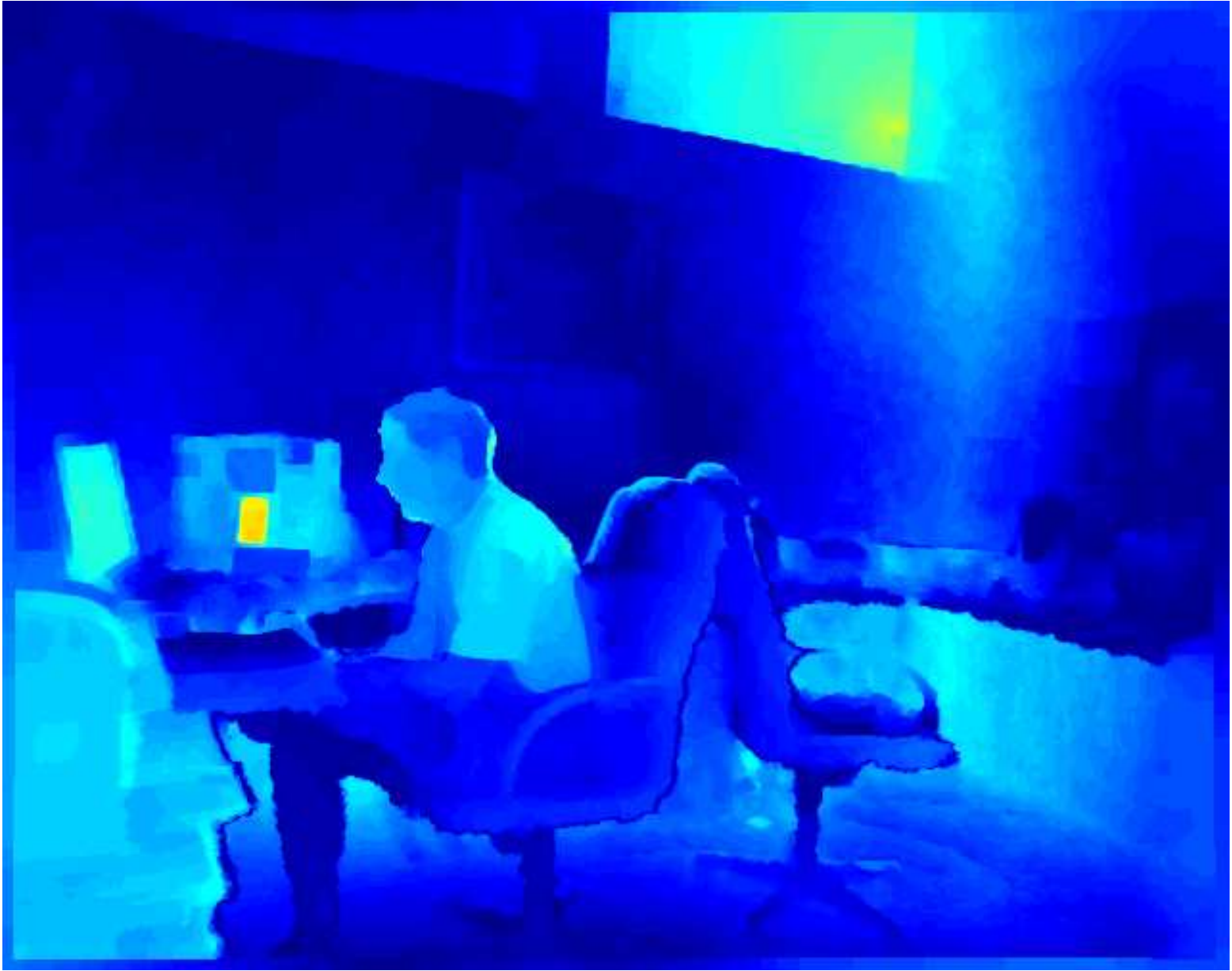}
	&\includegraphics[width=0.18\textwidth, height=0.12\textwidth]{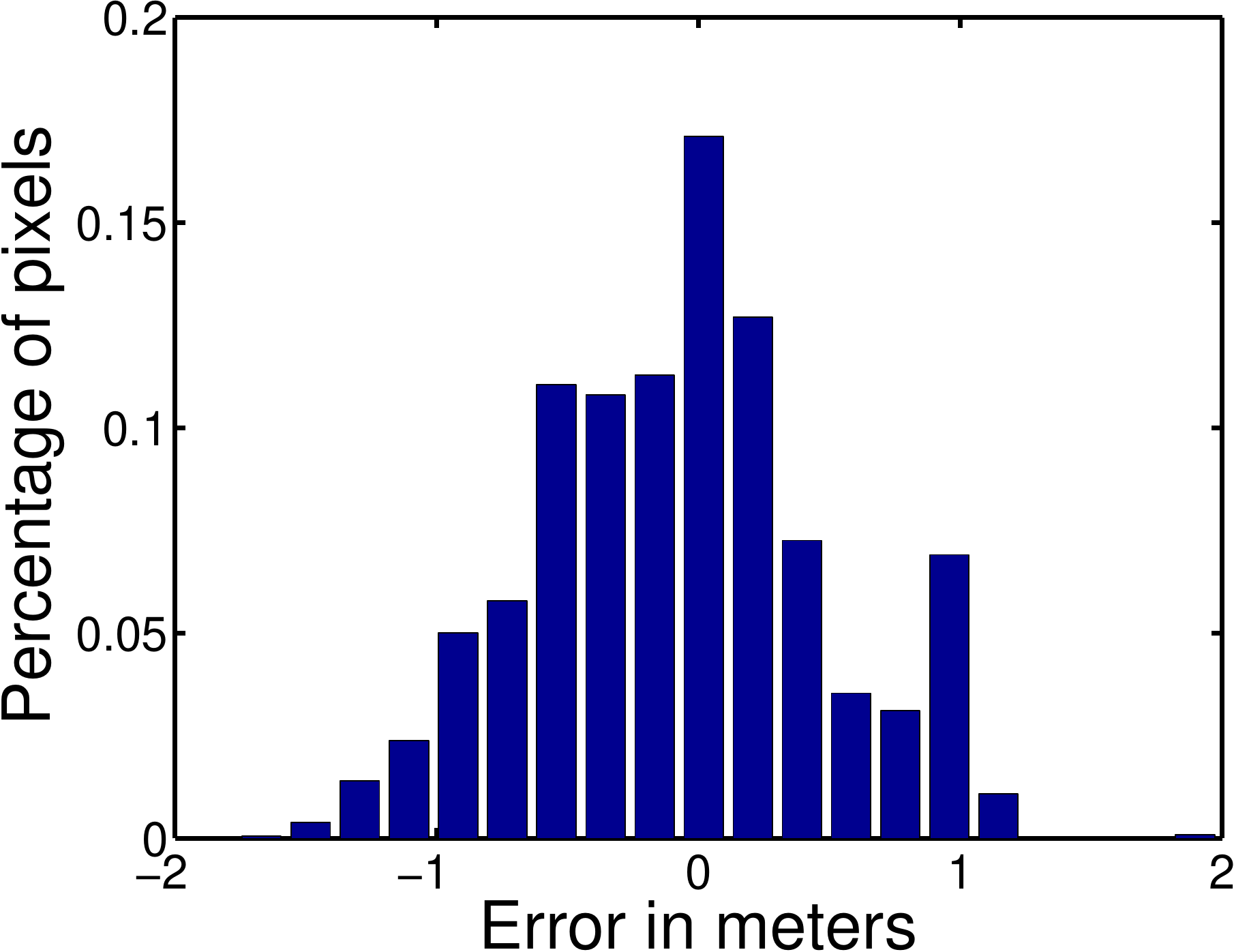}
	&\includegraphics[width=0.1\textwidth, height=0.12\textwidth]{./fig/Make3D/images/429.pdf}
	&\includegraphics[width=0.1\textwidth, height=0.12\textwidth]{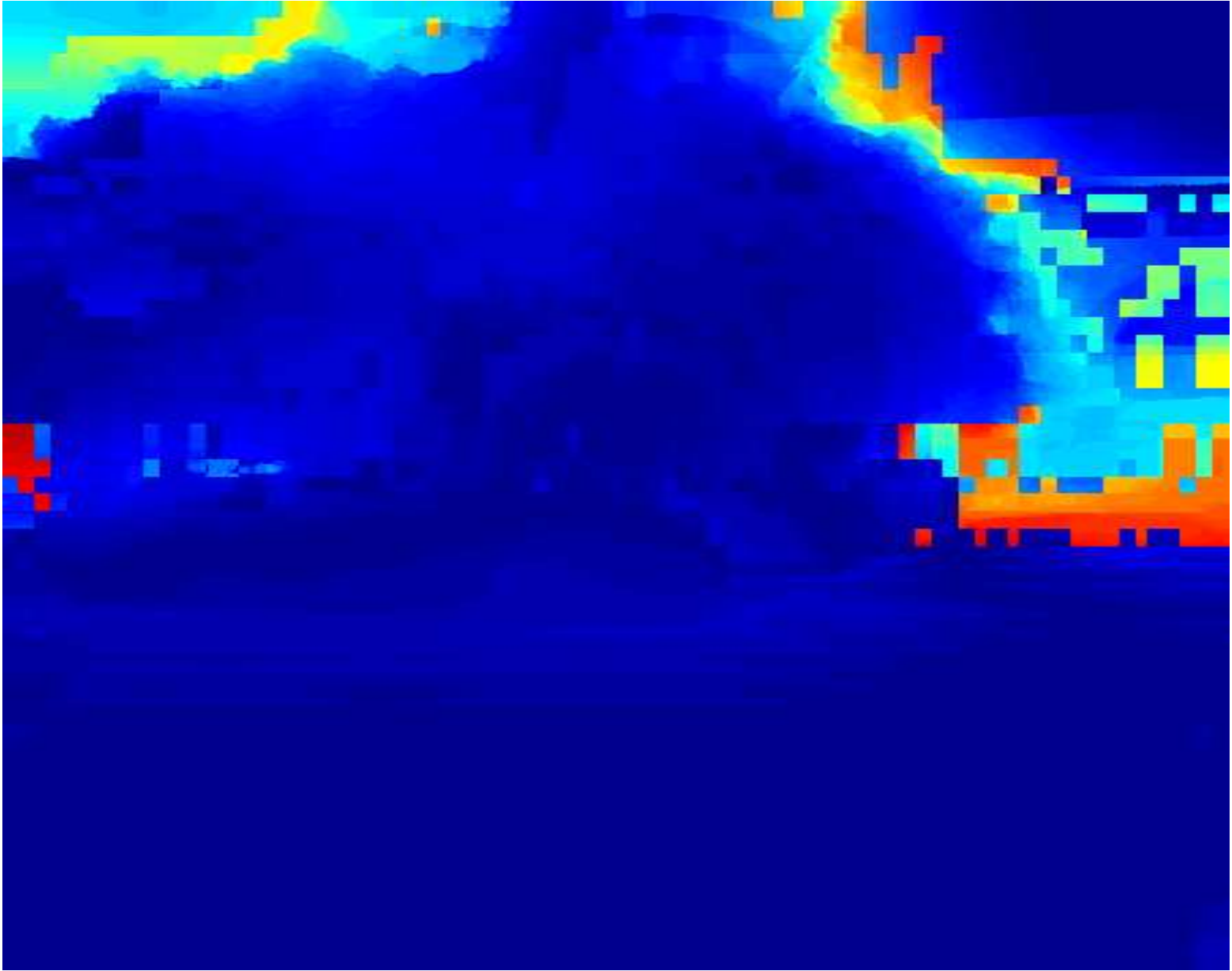}
	&\includegraphics[width=0.18\textwidth, height=0.12\textwidth]{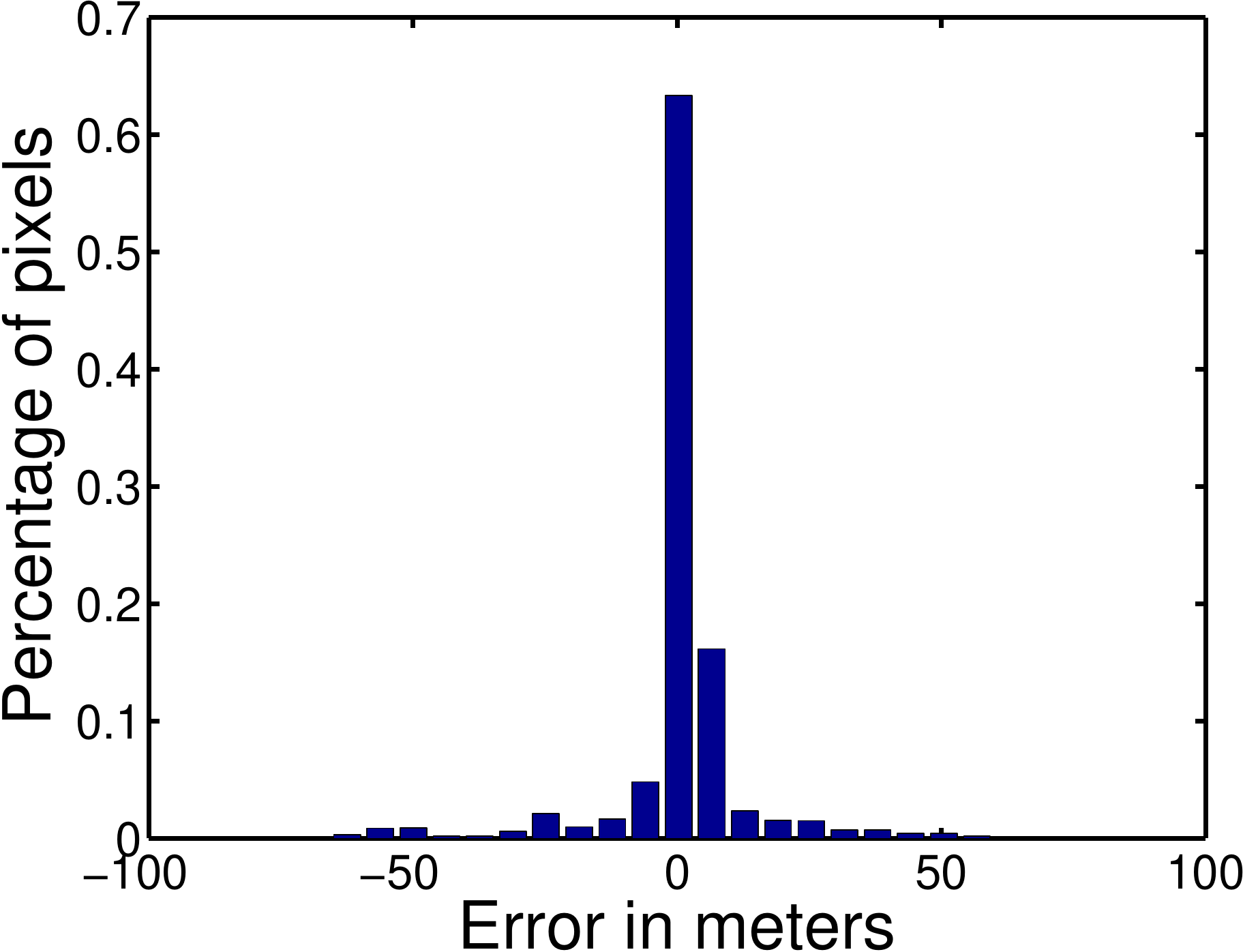}\\

	\includegraphics[width=0.155\textwidth, height=0.12\textwidth]{./fig/NYUD2/images/1424.pdf}
	&\includegraphics[width=0.155\textwidth, height=0.12\textwidth]{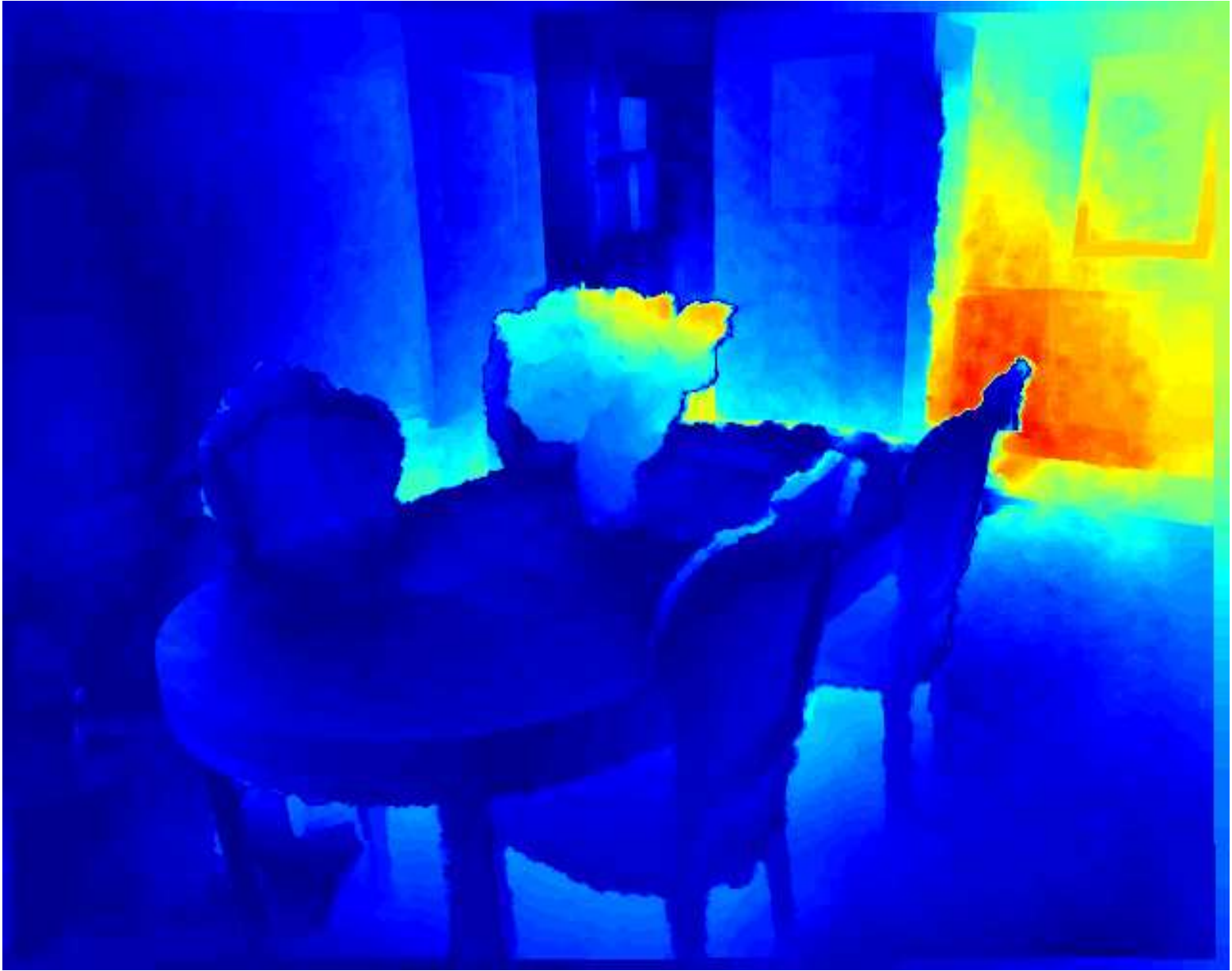}
	&\includegraphics[width=0.18\textwidth, height=0.12\textwidth]{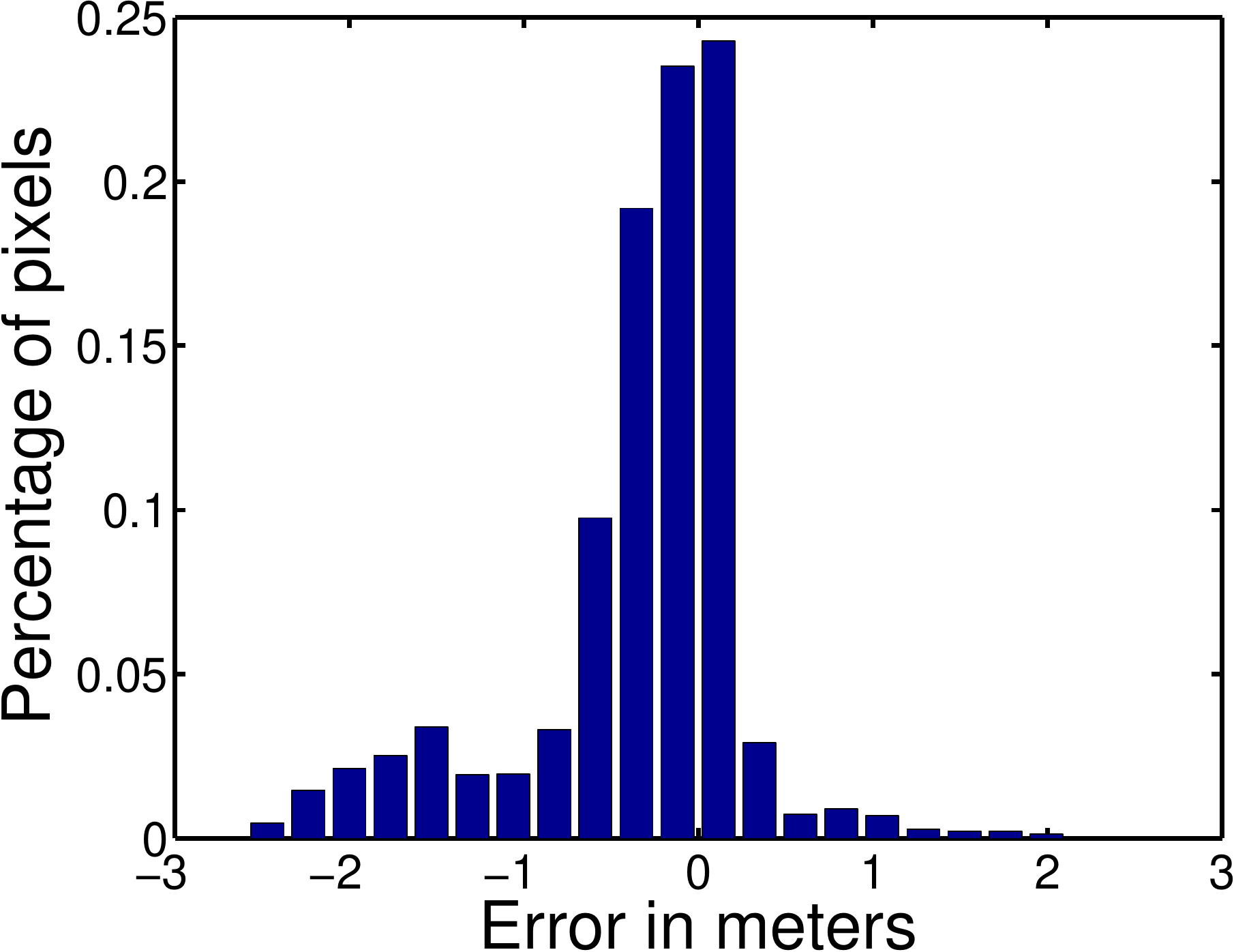}
	&\includegraphics[width=0.1\textwidth, height=0.12\textwidth]{./fig/Make3D/images/424.pdf}
	&\includegraphics[width=0.1\textwidth, height=0.12\textwidth]{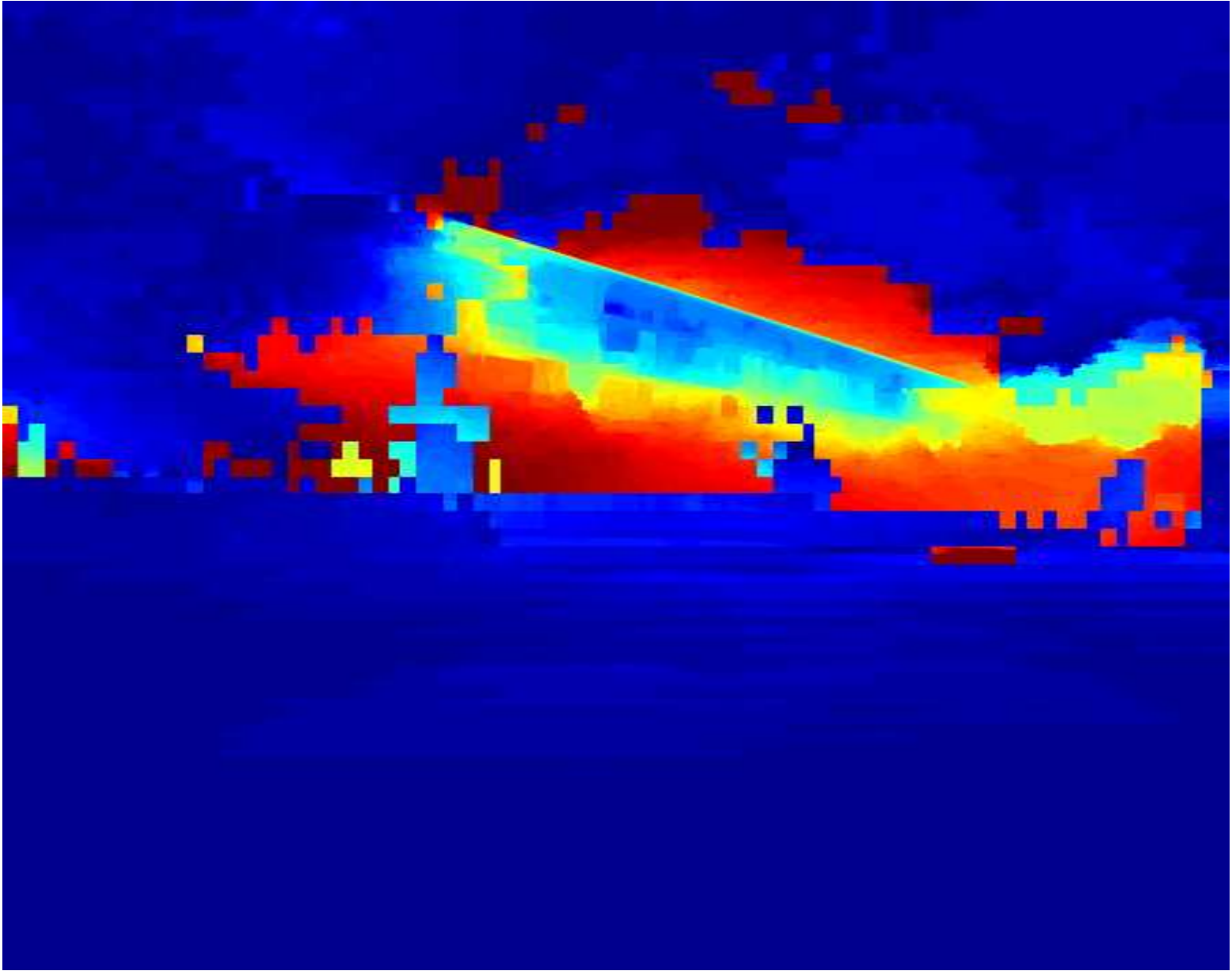}
	&\includegraphics[width=0.18\textwidth, height=0.12\textwidth]{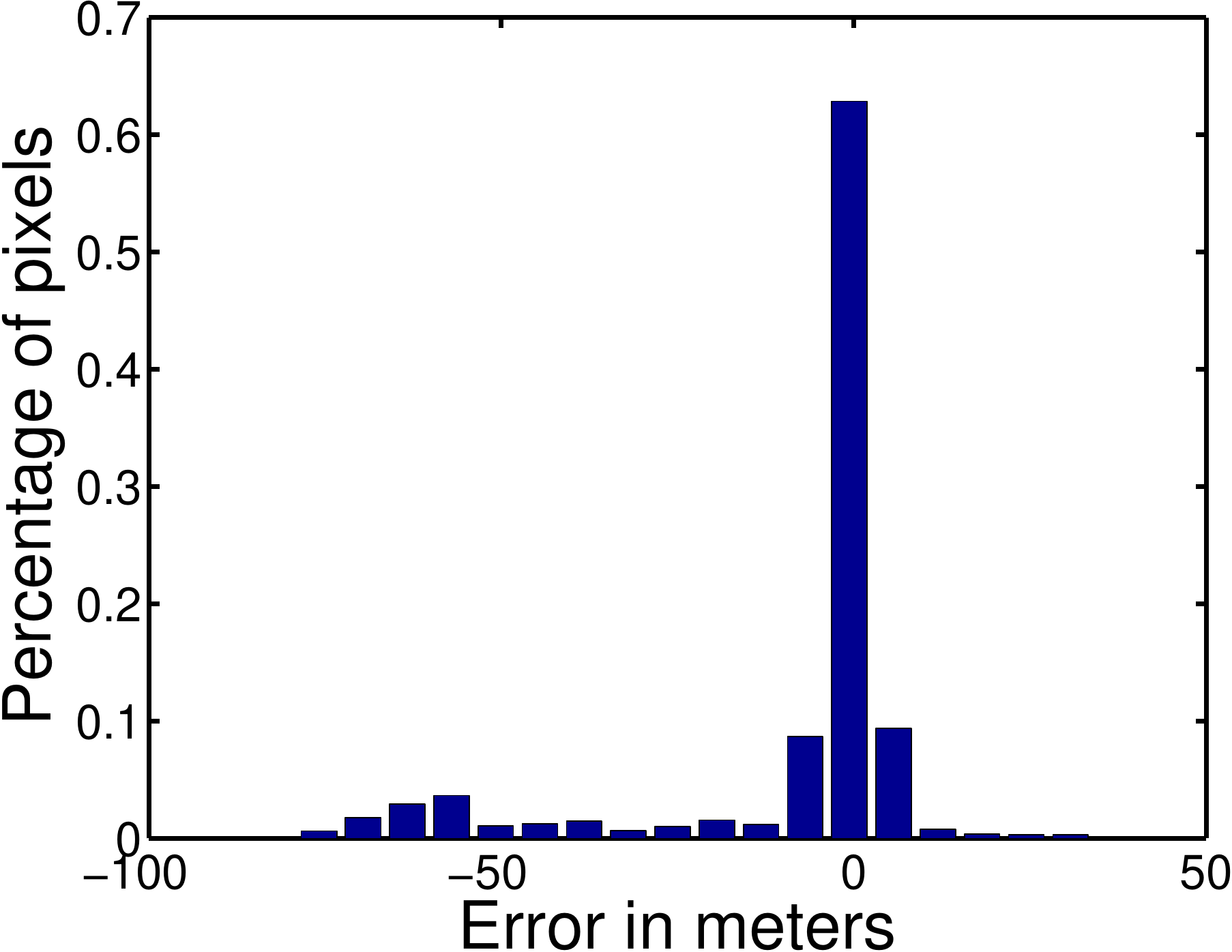} \\

	\multicolumn{3}{c|}{\includegraphics[width=0.3\textwidth, height=0.03\textwidth]{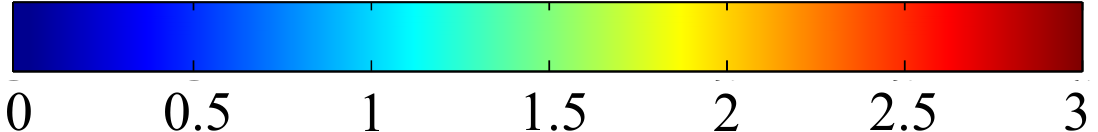}}
	&\multicolumn{3}{c}{\includegraphics[width=0.3\textwidth, height=0.03\textwidth]{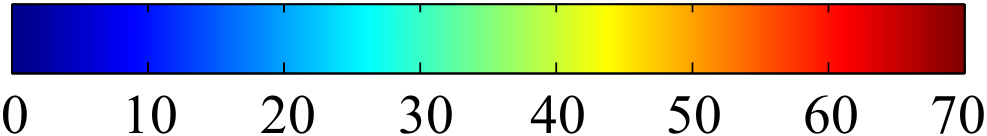}} \\

  {\footnotesize Test image} & {\footnotesize Absolute error map}  &{\footnotesize Error histogram}
  & {\footnotesize  Test image} & {\footnotesize  Absolute error map}  & {\footnotesize  Error histogram}\\
\end{tabular}
}
\caption{An illustration of the absolute error maps and the pixel-wise error histograms of our predictions (Left: NYU v2; Right: Make3D).
The absolute error maps are shown in meters, with the color bar shown in the last row.
For the error histogram plot, the horizontal axis shows the prediction error in meters (quantized into 20 bins), and the vertical axis shows the percentage of pixels in each bin.
} \label{fig:err_pixel}
\end{figure*}

\begin{figure} \center
	\includegraphics[width=0.42\textwidth]{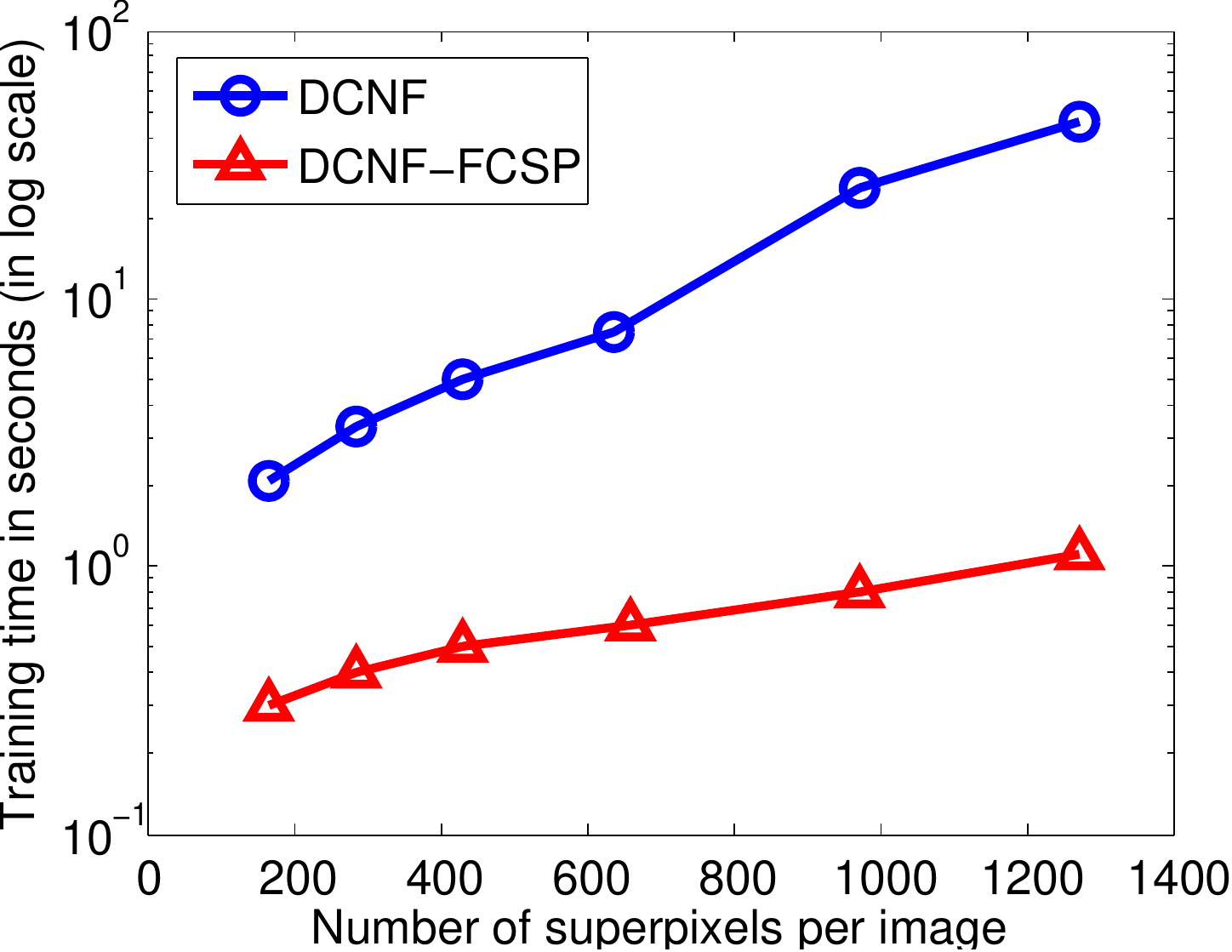}
  \vspace{0.1cm}
\caption{Comparison of the whole model training time (network forward + backward) in seconds (in $log$ scale) for one image on the NYU v2 dataset  with respect to different numbers of superpixels per image. The \dcnfsp model is orders of magnitude faster than the \dcnf model.}  \label{fig:timeComp}
\end{figure}

\begin{figure} \center
	\includegraphics[width=0.42\textwidth]{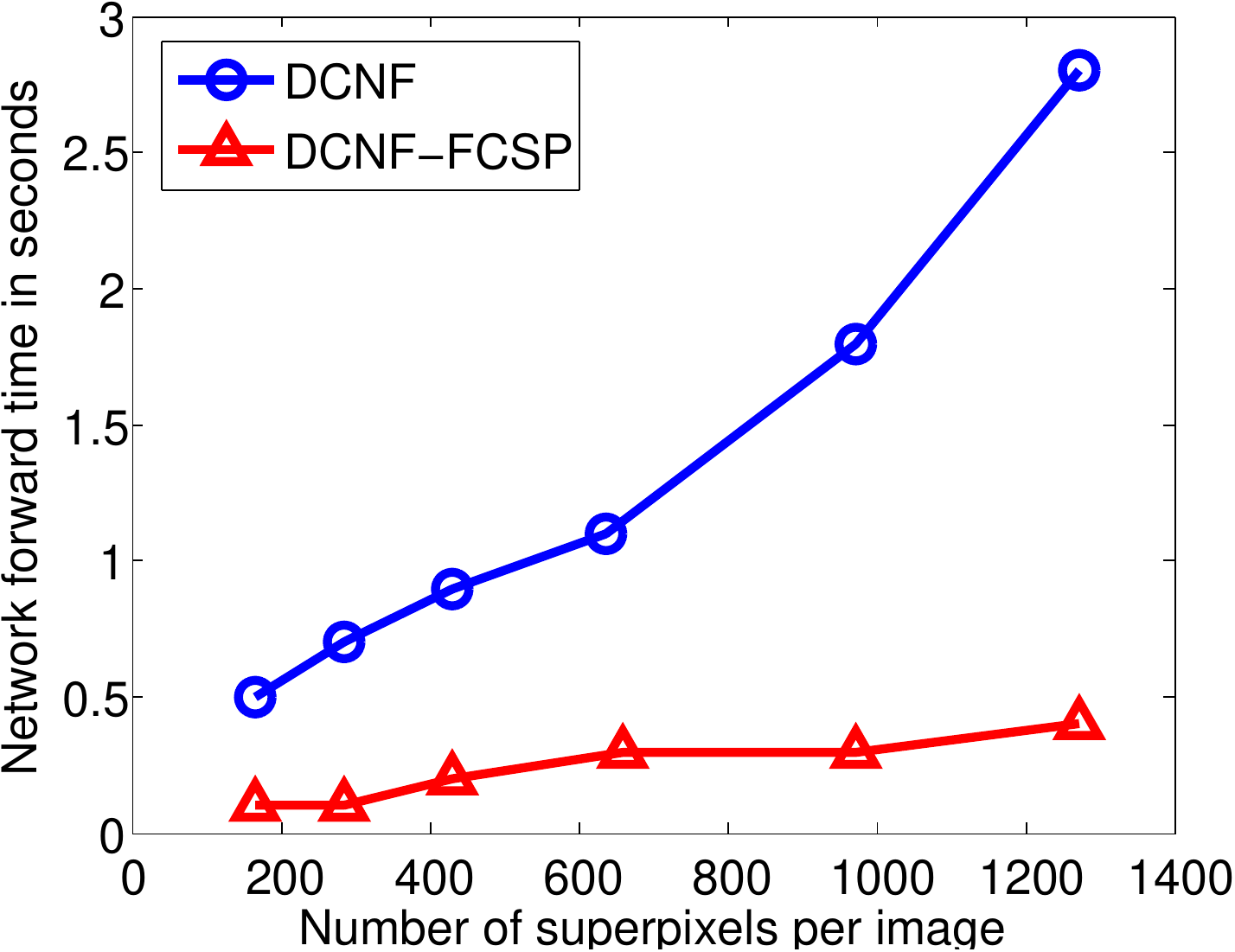}
\caption{Comparison of the network forward time of the whole model during depth prediction (in seconds) for one image on the NYU v2 dataset with respect to different numbers of superpixels per image. The \dcnfsp model is significantly faster than the \dcnf model.}  \label{fig:timeComp_pred}
\end{figure}

\begin{table*}
\caption{Baseline comparisons on the NYU v2 dataset.
Our method with the whole network training performs the best.
} \label{tab:anal_nyud2}
\center
\resizebox{0.7\linewidth}{!} {
\begin{tabular}{ | l |  c  c  c | c  c  c |}
\hline
\multirow{3}{*}{{{Method}}} &\multicolumn{3}{c|}{Error} &\multicolumn{3}{c|}{Accuracy} \\
&\multicolumn{3}{c|}{(lower is better)} &\multicolumn{3}{c|}{(higher is better)} \\
\cline{2-7}
&rel &log10 &rms &$\delta < 1.25$ &$\delta < 1.25^2$ &$\delta < 1.25^3$  \\
\hline
\hline
SVR  &0.313	&0.128	&1.068	&0.490	&0.787	&0.921 \\
SVR (smooth) &0.290	 &0.116	&0.993	&0.514	&0.821	&0.943\\
Unary only  &0.295 	 &0.117 	 &0.985   &0.516 	 &0.815 	 &0.938  \\
Unary only (smooth) &0.287  &0.112  &0.956  &0.535  &0.828  &0.943 \\
\dcnf (pre-train) &0.257	 &0.101	  &0.843  	&0.588   	&0.868	&0.961  \\
\dcnf (fine-tune)  &\textbf{0.230} 	 &\textbf{0.095} 	 &\textbf{0.824}  &\textbf{0.614} 	 &\textbf{0.883} 	 &\textbf{0.971}\\
\hline
\end{tabular}
}

\end{table*}

\begin{table*}
\caption{Baseline comparisons on the Make3D dataset.
Our method with the whole network training performs the best.
} \label{tab:anal_make3d}
\center
\resizebox{0.6\linewidth}{!} {
\begin{tabular}{ | l |  c  c  c | c  c  c |}
\hline
\multirow{3}{*}{{{Method}}} &\multicolumn{3}{c|}{Error (C1)} &\multicolumn{3}{c|}{Error (C2)} \\
&\multicolumn{3}{c|}{(lower is better)} &\multicolumn{3}{c|}{(lower is better)} \\
\cline{2-7}
&rel &log10 &rms  &rel &log10 &rms  \\
\hline
\hline
SVR  &0.433	&0.158	&8.93  &0.429	&0.170	&15.29  \\
SVR (smooth) &0.380	&0.140	&\textbf{8.12}  &0.384	&0.155	&15.10 \\
Unary only &0.366 	 &0.137 	 &8.63  &0.363 	 &0.148 	 &14.41 \\
Unary only (smooth) &0.341  &0.131  &8.49  &0.349  &0.144  &14.37 \\
\dcnf (pre-train) &0.331	 &0.127	 &8.82  &0.324	&0.134	&13.29 \\
\dcnf (fine-tune)   &\textbf{0.314} &\textbf{0.119} &8.60 &\textbf{0.307} &\textbf{0.125} &\textbf{12.89} \\
\hline
\end{tabular}
}
\end{table*}

\begin{table*}
\caption{Performance comparisons of \dcnf and \dcnfsp on the NYU v2 dataset.
The two models show comparable performance.
} \label{tab:dcnf_cmp_nyud2}
\center
\resizebox{0.7\linewidth}{!} {
\begin{tabular}{ | l |  c  c  c | c  c  c |}
\hline
\multirow{3}{*}{{{Method}}} &\multicolumn{3}{c|}{Error} &\multicolumn{3}{c|}{Accuracy} \\
&\multicolumn{3}{c|}{(lower is better)} &\multicolumn{3}{c|}{(higher is better)} \\
\cline{2-7}
&rel &log10 &rms &$\delta < 1.25$ &$\delta < 1.25^2$ &$\delta < 1.25^3$  \\
\hline
\hline
\dcnf (pre-train) &0.257	 &0.101	  &0.843  	&0.588   	&0.868	&0.961  \\
\dcnf (fine-tune)  &0.230 	 &0.095 	 &0.824  &\textbf{0.614} 	 &0.883 	 &\textbf{0.971}\\
\dcnfsp (pre-train) &0.261 &0.100 &0.842 &0.583 &0.869 &0.964  \\
\dcnfsp (fine-tune)  &0.237	&\textbf{0.082}	&\textbf{0.822}	&0.608	&\textbf{0.889}	&0.969 \\
\hline
\end{tabular}
}

\end{table*}

\begin{table*}
\caption{Performance comparisons of \dcnf and \dcnfsp on the Make3D dataset.
The two models perform on par in general.
} \label{tab:dcnf_cmp_make3d}
\center
\resizebox{0.6\linewidth}{!} {
\begin{tabular}{ | l |  c  c  c | c  c  c |}
\hline
\multirow{3}{*}{{{Method}}} &\multicolumn{3}{c|}{Error (C1)} &\multicolumn{3}{c|}{Error (C2)} \\
&\multicolumn{3}{c|}{(lower is better)} &\multicolumn{3}{c|}{(lower is better)} \\
\cline{2-7}
&rel &log10 &rms  &rel &log10 &rms  \\
\hline
\hline
\dcnf (pre-train) &0.331	 &0.127	 &8.82  &0.324	&0.134	&13.29 \\
\dcnf (fine-tune)   &0.314 &0.119 &\textbf{8.60} &0.307 &0.125 &\textbf{12.89} \\
\dcnfsp (pre-train) &0.323	&0.127	&9.01	&0.318	&0.136	&13.89  \\
\dcnfsp (fine-tune)   &\textbf{0.312}	&\textbf{0.113}	&9.10	  &\textbf{0.305}  &\textbf{0.120} 	&13.24  \\
\hline
\end{tabular}
}

\end{table*}

\begin{table*} [t]
\caption{State-of-the-art comparisons on the NYU v2 dataset.
Our method performs the best in most cases.
Note that the results of Eigen \etal \cite{dcnn_nips14} are obtained by using extra training data (in the millions in total) while ours are obtained using the standard training set.} \label{tab:nyud2}
\center
\resizebox{.7\linewidth}{!} {
\begin{tabular}{ | l |  c  c  c | c  c  c |}
\hline
\multirow{3}{*}{{{Method}}} &\multicolumn{3}{c|}{Error} &\multicolumn{3}{c|}{Accuracy} \\
&\multicolumn{3}{c|}{(lower is better)} &\multicolumn{3}{c|}{(higher is better)} \\
\cline{2-7}
&rel &log10 &rms &$\delta < 1.25$ &$\delta < 1.25^2$ &$\delta < 1.25^3$  \\
\hline
\hline
Saxena \etal \cite{make3d_pami09}		  &0.349  &-  &1.214	&0.447	&0.745	&0.897 \\
DepthTransfer \cite{depthTransfer_pami14}	&0.35 	&0.131	&1.2		&- &- &-	 \\
Discrete-continuous \crf \cite{Miaomiao_cvpr14}	    &0.335	&0.127	&1.06 &- &- &-	 \\
Ladicky \etal \cite{Ladicky_cvpr14}		&- &- &-	&0.542	&0.829	&0.941  \\
Eigen \etal \cite{dcnn_nips14}      &0.215	&-	&0.907	&0.611	&0.887	&0.971 \\
\hline
\dcnfsp (pre-train) &0.234	&0.095	&0.842	&0.604	&0.885	&0.973 \\
\dcnfsp (fine-tune) &\textbf{0.213}	&\textbf{0.087}	&\textbf{0.759}	&\textbf{0.650} 	&\textbf{0.906}	&\textbf{0.976} \\
\hline
\end{tabular}
}

\end{table*}

\begin{table*}
\caption{State-of-the-art comparisons on the Make3D dataset.
Our method performs the best.
Note that the C2 errors of the Discrete-continuous \crf \cite{Miaomiao_cvpr14} are reported with an ad-hoc post-processing step (train a classifier to label sky pixels and set the corresponding regions to the maximum depth).} \label{tab:make3d}
\center
\resizebox{.6\linewidth}{!} {
\begin{tabular}{ | l |  c  c  c | c  c  c |}
\hline
\multirow{2}{*}{{{Method}}} &\multicolumn{3}{c|}{Error (C1)} &\multicolumn{3}{c|}{Error (C2)} \\
&\multicolumn{3}{c|}{(lower is better)} &\multicolumn{3}{c|}{(lower is better)} \\
\cline{2-7}
&rel &log10 &rms  &rel &log10 &rms  \\
\hline
\hline
Saxena \etal \cite{make3d_pami09}		  &-  &-  &- 	&0.370	&0.187  &-  \\
Semantic Labelling \cite{Liu_cvpr12}  &-  &-  &- 	&0.379 &0.148 &- \\
DepthTransfer \cite{depthTransfer_pami14}	&0.355	&0.127	&9.20  &0.361	  &0.148	  &15.10 \\
Discrete-continuous \crf \cite{Miaomiao_cvpr14}	    &0.335	&0.137	&9.49  &0.338	  &0.134	 &\textbf{12.60}  \\
\hline
\dcnfsp (pre-train) &0.331	&0.119	&7.77	&0.330	&0.133	&14.46 \\
\dcnfsp (fine-tune)  &\textbf{0.287}	&\textbf{0.109}	&\textbf{7.36}	&\textbf{0.287}	 &\textbf{0.122} &14.09 \\
\hline
\end{tabular}
}

\end{table*}

\begin{table*} [t]
\caption{State-of-the-art comparisons on the KITTI dataset.
Our method achieves the best RMS error.
Note that the results of Eigen \etal \cite{dcnn_nips14} are obtained by using extra training data (in the millions in total) while ours are obtained using 700 training images.
The results of Saxena \etal \cite{make3d_pami09} are reproduced from \cite{dcnn_nips14}.} \label{tab:kitti}
\center
\resizebox{.7\linewidth}{!} {
\begin{tabular}{ | l |  c  c  c | c  c  c |}
\hline
\multirow{3}{*}{{{Method}}} &\multicolumn{3}{c|}{Error} &\multicolumn{3}{c|}{Accuracy} \\
&\multicolumn{3}{c|}{(lower is better)} &\multicolumn{3}{c|}{(higher is better)} \\
\cline{2-7}
&rel &log10 &rms &$\delta < 1.25$ &$\delta < 1.25^2$ &$\delta < 1.25^3$  \\
\hline
\hline
Saxena \etal \cite{make3d_pami09}		  &0.280  &-  &8.734 	&0.601	&0.820	&0.926 \\
Eigen \etal \cite{dcnn_nips14}      &\textbf{0.190}	&-	&7.156	&\textbf{0.692}	&\textbf{0.899}	&\textbf{0.967} \\
\hline
\dcnfsp (pre-train) &0.236	&0.101	&7.421	&0.613	&0.858	&0.949 \\
\dcnfsp (fine-tune) &0.217	&\textbf{0.092}	&\textbf{7.046}	&0.656 	&0.881	&0.958 \\
\hline
\end{tabular}
}

\end{table*}

\section{Experiments}
We organize our experiments into the following three parts:
1) We compare our \dcnf model with several baseline methods to show the benefits of jointly learning \cnn and \crf ;
2) We perform comparisons between the two proposed models, \ie, \dcnf and \dcnfsp, to show that the \dcnfsp model is equally effective while generally being $\sim 10$ times faster;
3) We compare our \dcnfsp model using deeper network design with state-of-the-art methods to show that our model performs significantly better.
We evaluate on three popular datasets which are available online: the indoor NYU v2 Kinect dataset \cite{nyud2_eccv12}, the outdoor Make3D range image dataset \cite{make3d_pami09} and the KITTI dataset \cite{kitti}.
Several measures commonly used in prior works are applied here for quantitative evaluations:
\begin{itemize}
\item
average relative error (rel):
$\frac{1}{T} \sum_p \frac{|d_p^{gt} - d_p|}{d_p^{gt}} $;
\item
root mean squared error (rms):
$\sqrt{\frac{1}{T} \sum_p (d_p^{gt} - d_p)^2}$;
\item
average $\log_{10}$ error (log10): \\
$\frac{1}{T} \sum_p | \log_{10}d_p^{gt} - \log_{10}d_p|$;
\item
accuracy with threshold $thr$: \\
percentage ($\%$) of $d_p \; \; \st \max (\frac{d_p^{gt}}{d_p}, \frac{d_p}{d_p^{gt}}) = \delta < thr$;
\end{itemize}
where $d_p^{gt}$ and $d_p$ are the ground-truth and predicted depths respectively at pixel indexed by $p$, and $T$ is the total number of pixels in all the evaluated images.

We use SLIC \cite{slic_pami12} to segment the images into a set of non-overlapping superpixels.
For the \dcnf model, we consider the image patch within a rectangular box centred on the centroid of each of the superpixels, which contains a large portion of its background surroundings. More specifically, we use a box size of 168$\times$168 pixels for the NYU v2 and $120 \times 120$ pixels for the Make3D dataset.
Following \cite{make3d_pami09,Liu_cvpr12,dcnn_nips14}, we transform the depths into log-scale before training.
For better visualizations, we apply a cross-bilateral filter \cite{cbfilter} for inpainting using the provided toolbox \cite{nyud2_eccv12} after obtaining the superpixel depth predictions.
Our experiments empirically show that this post-processing has negligibly impact on the evaluation performance.

\subsection{Baseline comparisons}
To demonstrate the effectiveness of the proposed method, we first conduct experimental comparisons against several baseline methods:
\begin{itemize}
\item
SVR: We train a support vector regressor using the \cnn representations from the first 6 layers of Fig. \ref{fig:dcnf_unary};
\item
SVR (smooth): We add a smoothness term to the trained SVR during prediction by solving the inference problem in Eq. \eqref{eq:inf_solution}. As tuning multiple pairwise parameters is not straightforward, we only use color difference as the pairwise potential and choose the parameter $\beta$ by hand-tuning on a validation set;
\item
Unary only: We replace the \crf loss layer in Fig. \ref{fig:dcnf} with a least-square regression layer (by setting the pairwise outputs $\pws_{pq}=0$, $p,q=1,...,n$), which degenerates to a deep regression model trained by SGD.
\item
Unary only (smooth): We add similar smoothness term to our unary only model, as did in the SVR (smooth) case.
\end{itemize}

\subsubsection{NYU v2 data}
The NYU v2 dataset consists of $1449$ RGBD images of indoor scenes, among which $795$ are used for training and 654 for test (we use the standard training/test split provided with the dataset).
We report the baseline comparisons in Table \ref{tab:anal_nyud2}.
From the table, several conclusions can be made: 1) When trained with only unary term, deeper network is beneficial for better performance, which is demonstrated by the fact that our unary only model outperforms the SVR model;
2) Adding smoothness term to the SVR or our unary only model helps improve the prediction accuracy;
3) Our \dcnf model achieves the best performance by jointly learning the unary and the pairwise parameters in a unified deep \cnn framework. Moreover, fine-tuning the whole network yields further performance gain.
These well demonstrate the efficacy of our model.

\subsubsection{Make3D data}

The Make3D dataset contains 534 images depicting outdoor scenes,
 with $400$ for training and $134$ images for test.
As pointed out in \cite{make3d_pami09,Miaomiao_cvpr14}, this dataset is with limitations: the maximum value of depths is 81m with far-away objects are all mapped to the one distance of $81$ meters.
As a remedy, two criteria are used in \cite{Miaomiao_cvpr14} to report the prediction errors: ($C_1$) Errors are calculated only in the regions with the ground-truth depth less than $70$ meters; ($C_2$) Errors are calculated over the entire image.
We follow this protocol to report the evaluation results in Table \ref{tab:anal_make3d}.
As we can see, our full \dcnf model with the whole network training performs the best among all the compared baseline methods.
Using deeper networks and adding smoothness terms generally help improve the performance.

\subsection{\dcnf vs. \dcnfsp}
\label{sec:exp_time}
In this section, we compare the performance of the proposed \dcnf and \dcnfsp in terms of both prediction accuracy and computational efficiency.
The compared prediction performance are reported in Table \ref{tab:dcnf_cmp_nyud2} and Table \ref{tab:dcnf_cmp_make3d}.
We can see that the proposed \dcnfsp model performs very close to the \dcnf model.
Next, we compare the computational efficiency of these two models.
Specifically, we report the training time (network forward + backward ) of one image for both whole models in terms of different numbers of superpixels we use per image.
The comparison is conducted on the NYU v2 dataset and is shown in Fig. \ref{fig:timeComp}.
As demonstrated, the \dcnfsp model is generally orders of magnitude faster than the \dcnf model.
Moreover, the speedup becomes more significant with the number of superpixels increases.
We also compare the network forward time of whole models during depth predictions, and plot the results in Fig. \ref{fig:timeComp_pred}.
The shown time is for processing one image.
We can see that the \dcnfsp model is much faster as well as more scalable than the \dcnf model.
Most importantly, with this more efficient \dcnfsp model, we can design deeper network for better performance, as we will show in the sequel in Sec. \ref{sec:exp_stateComp}.

\subsection{State-of-the-art comparisons}
\label{sec:exp_stateComp}
Recent studies have shown that very deep networks can significantly improve the image classifications  performance \cite{googleNet,vgg16_arxiv}.
Thanks to the speedup brought about by the superpixel pooling method, we are now able to design deeper networks in our framework.
We transfer the popular VGG-16 net trained on the ImageNet from \cite{vgg16_arxiv}.
Specifically, we replace the AlexNet part (the first $5$ convolutional layers in Fig. \ref{fig:dcnfsp_fcnn}) with all the convolutional layers (including the $5$-th pooling layer) in VGG-16.
These layers are followed by the $2$ newly added convolutional layers with $512$ channels each,
and then the $3$ fully connected layer to construct the unary potentials.
We follow the same training protocol, \ie, first pre-train the remaining layers by fixing the transferred layers (the VGG-16 net part) and then fine-tune the whole network.

\subsubsection{NYU v2 data}
In Table \ref{tab:nyud2}, we report the results compared to several popular state-of-the-art methods on the NYU v2 dataset.
As can be observed, our method outperforms classic methods like Make3d \cite{make3d_pami09}, DepthTransfer \cite{depthTransfer_pami14} with large margins.
Most notably, our results are significantly better than that of \cite{Ladicky_cvpr14}, which jointly exploits depth estimation and semantic labelling.
Compared to the recent work of Eigen \etal \cite{dcnn_nips14}, our method  also exhibits better performance in terms of all metrics.
Note that, to overcome overfit, they \cite{dcnn_nips14} have to collect millions of additional labelled images to train their model.
In contrast, we only use the standard training sets ($795$) without any extra data,
yet we achieve better performance.
Fig. \ref{fig:nyud2} illustrates some qualitative evaluations of our method compared against Eigen \etal \cite{dcnn_nips14} (We download the predictions of \cite{dcnn_nips14} from the authors' website.).
Compared to the coarse predictions of \cite{dcnn_nips14},
our method yields better visualizations with sharper transitions, aligning to local details.
To better illustrate how our predictions deviate from the ground-truth depths, we plot the absolute depth error maps and the pixel-wise error histograms in Fig. \ref{fig:err_pixel}.
Specifically, the absolute error maps are shown in meters, with the color bar shown in the last row.
For the error histogram plot, the horizontal axis shows the prediction error in meters (quantized into 20 bins), and the vertical axis shows the percentage of pixels in each bin.
As we can see, our predictions are mostly well aligned to the ground truth depth maps.

\subsubsection{Make3D data}
We show the compared results on the Make3D dataset in Table \ref{tab:make3d}.
We can see that our \dcnfsp model with
the whole network training
ranks the first in overall performance,
outperforming the compared methods by large margins.
Note that the C2 errors of \cite{Miaomiao_cvpr14} are reported with an ad-hoc post-processing step, which trains a classifier to label sky pixels and set the corresponding regions to the maximum depth.
In contrast, we do not employ any of those heuristics to refine our results, yet we achieve better results in terms of relative error.
Compared to the results of the \dcnfsp model using smaller net in Table \ref{tab:dcnf_cmp_make3d},
we get better $C1$ error but degraded $C2$ error.
This can be explained from the limitations of this dataset that the depths of all far away objects are all set to one maximum depth value.
Some examples of qualitative  evaluations are shown in Fig. \ref{fig:nmake3d}.
By jointly learning the unary and pairwise potentials, our \dcnfsp model produce predictions that well capture local details.
The absolute depth error maps and the pixel-wise error histograms are shown in the right part of Fig. \ref{fig:err_pixel}.
As can be observed, our predictions are mostly well aligned to the ground truth depth maps, with most of the prediction errors are on the boundary regions which show extremely large depth jumps.
In these cases, our predictions exhibit relatively mild depth transitions.

\subsubsection{KITTI data}
We further perform depth estimation on the KITTI dataset \cite{kitti}, which consists of videos taken from
a driving vehicle with depths captured by a LiDAR sensor.
We use the same test set, \ie, 697 images from 28 scenes, as provided by Eigen \cite{dcnn_nips14}.
As for the training set, we use the same 700 images that Eigen \etal \cite{dcnn_nips14} used to train the method of Saxena \etal \cite{make3d_pami09}.
Since the ground-truth depths of the KITTI dataset are scattered at irregularly spaced points, which only consists of $\sim 5\%$ pixels of each image, we extract the ground-truth depth closest to each superpixel centroid as the superpixel depth label.
We then construct our \crf graph only on those superpixels that have ground-truth labels.
The compared results are presented in Table \ref{tab:kitti}.
In summary, Eigen \etal \cite{dcnn_nips14} have achieved slightly better performance on this dataset by
leveraging large amounts of training data (in the millions).
This can be explained by the fact that the highly sparse ground-truth depth maps lessened the benefits of the pairwise term in our model.
Fig. \ref{fig:kitti} shows some prediction examples.

\section{Conclusion}
We have presented a deep convolutional neural field model for depth estimation from a single image.
The proposed method combines the strength of deep \cnn and continuous \crf
in a unified \cnn framework.
We show that the log-likelihood optimization in our method can be directly solved using back propagation without any approximations required.
Predicting the depths of a new image by solving the MAP inference can be efficiently performed as closed-form solutions exist.
We further propose an improved model that  is based on fully convolutional networks and a novel superpixel pooling method.
We experimentally demonstrate that it is equally effective while brings orders of magnitude faster training speedup, which enables the use of deeper networks for better performance.
Experimental results demonstrate that the proposed method outperforms state-of-the-art methods on both indoor and outdoor scene datasets.

The main limitation of our method is that it does not exploit any geometric cues, which can be explored in the future work.
Given the general learning framework of our method, it is also possible to be applied to other vision applications
with minimum modification, \eg, image denoising, and deblurring.
Another potential working direction is to apply our depth estimation models for benefiting other vision tasks, e.g., semantic segmentation, object detection, on traditional vision datasets where ground-truth depths are not available.

\section*{Acknowledgements}

  C. Shen's participation was in part support by ARC Future Fellowship.
  The authors would like to thank NVIDIA for GPU donation.
  Correspondence should be addressed to C. Shen (chunhua.shen@adelaide.edu.au).
{
    \bibliographystyle{IEEEtran}
    \bibliography{CSRef}

\begin{thebibliography}{10}
\providecommand{\url}[1]{#1}
\csname url@samestyle\endcsname
\providecommand{\newblock}{\relax}
\providecommand{\bibinfo}[2]{#2}
\providecommand{\BIBentrySTDinterwordspacing}{\spaceskip=0pt\relax}
\providecommand{\BIBentryALTinterwordstretchfactor}{4}
\providecommand{\BIBentryALTinterwordspacing}{\spaceskip=\fontdimen2\font plus
\BIBentryALTinterwordstretchfactor\fontdimen3\font minus
  \fontdimen4\font\relax}
\providecommand{\BIBforeignlanguage}[2]{{%
\expandafter\ifx\csname l@#1\endcsname\relax
\typeout{** WARNING: IEEEtran.bst: No hyphenation pattern has been}%
\typeout{** loaded for the language `#1'. Using the pattern for}%
\typeout{** the default language instead.}%
\else
\language=\csname l@#1\endcsname
\fi
#2}}
\providecommand{\BIBdecl}{\relax}
\BIBdecl

\bibitem{Ladicky_cvpr14}
L.~Ladický, J.~Shi, and M.~Pollefeys, ``Pulling things out of perspective,''
  in \emph{Proc. IEEE Conf. Comp. Vis. Patt. Recogn.}, 2014.

\bibitem{Shotton_pami13}
J.~Shotton, R.~B. Girshick, A.~W. Fitzgibbon, T.~Sharp, M.~Cook, M.~Finocchio,
  R.~Moore, P.~Kohli, A.~Criminisi, A.~Kipman, and A.~Blake, ``Efficient human
  pose estimation from single depth images,'' \emph{{IEEE} Trans. Pattern Anal.
  Mach. Intell.}, 2013.

\bibitem{dcnn_nips14}
D.~Eigen, C.~Puhrsch, and R.~Fergus, ``Depth map prediction from a single image
  using a multi-scale deep network,'' in \emph{Proc. Adv. Neural Inf. Process.
  Syst.}, 2014.

\bibitem{Hedau_eccv10}
V.~Hedau, D.~Hoiem, and D.~A. Forsyth, ``Thinking inside the box: Using
  appearance models and context based on room geometry,'' in \emph{Proc. Eur.
  Conf. Comp. Vis.}, 2010.

\bibitem{Lee_nips10}
D.~C. Lee, A.~Gupta, M.~Hebert, and T.~Kanade, ``Estimating spatial layout of
  rooms using volumetric reasoning about objects and surfaces,'' in \emph{Proc.
  Adv. Neural Inf. Process. Syst.}, 2010.

\bibitem{Gupta_eccv10}
A.~Gupta, A.~A. Efros, and M.~Hebert, ``Blocks world revisited: Image
  understanding using qualitative geometry and mechanics,'' in \emph{Proc. Eur.
  Conf. Comp. Vis.}, 2010.

\bibitem{depthTransfer_pami14}
K.~Karsch, C.~Liu, and S.~B. Kang, ``Depthtransfer: Depth extraction from video
  using non-parametric sampling,'' \emph{{IEEE} Trans. Pattern Anal. Mach.
  Intell.}, 2014.

\bibitem{Russell_cvpr09}
B.~C. Russell and A.~Torralba, ``Building a database of 3d scenes from user
  annotations,'' in \emph{Proc. IEEE Conf. Comp. Vis. Patt. Recogn.}, 2009.

\bibitem{Liu_cvpr12}
B.~Liu, S.~Gould, and D.~Koller, ``Single image depth estimation from predicted
  semantic labels,'' in \emph{Proc. IEEE Conf. Comp. Vis. Patt. Recogn.}, 2010.

\bibitem{Lafferty_icml01}
J.~D. Lafferty, A.~McCallum, and F.~C.~N. Pereira, ``Conditional random fields:
  Probabilistic models for segmenting and labeling sequence data,'' in
  \emph{Proc. Int. Conf. Mach. Learn.}, 2001.

\bibitem{ccrf_nips08}
T.~Qin, T.-Y. Liu, X.-D. Zhang, D.-S. Wang, and H.~Li, ``Global ranking using
  continuous conditional random fields,'' in \emph{Proc. Adv. Neural Inf.
  Process. Syst.}, 2008.

\bibitem{ccrf_ecai10}
V.~Radosavljevic, S.~Vucetic, and Z.~Obradovic, ``Continuous conditional random
  fields for regression in remote sensing,'' in \emph{Proc. Eur. Conf.
  Artificial Intell.}, 2010.

\bibitem{ccrf_aaai13}
K.~Ristovski, V.~Radosavljevic, S.~Vucetic, and Z.~Obradovic, ``Continuous
  conditional random fields for efficient regression in large fully connected
  graphs,'' in \emph{{AAAI} Conf. Artificial Intell.}, 2013.

\bibitem{cnn_Lecun89}
Y.~LeCun, B.~Boser, J.~S. Denker, D.~Henderson, R.~E. Howard, W.~Hubbard, and
  L.~D. Jackel, ``Backpropagation applied to handwritten zip code
  recognition,'' \emph{Neural Comput.}, 1989.

\bibitem{AlexNet12}
A.~Krizhevsky, I.~Sutskever, and G.~E. Hinton, ``{ImageNet} classification with
  deep convolutional neural networks,'' in \emph{Proc. Adv. Neural Inf.
  Process. Syst.}, 2012.

\bibitem{astound_cvprW14}
A.~Sharif~Razavian, H.~Azizpour, J.~Sullivan, and S.~Carlsson, ``{CNN} features
  off-the-shelf: An astounding baseline for recognition,'' in \emph{Proc. IEEE
  Conf. Comput. Vis. \& Pattern Recogn. Workshops}, June 2014.

\bibitem{Long_arxiv14}
J.~Long, E.~Shelhamer, and T.~Darrell, ``Fully convolutional networks for
  semantic segmentation,'' in \emph{Proc. IEEE Conf. Comp. Vis. Patt. Recogn.},
  2015.

\bibitem{Cogswell_arxiv14}
\BIBentryALTinterwordspacing
M.~Cogswell, X.~Lin, S.~Purushwalkam, and D.~Batra, ``Combining the best of
  graphical models and convnets for semantic segmentation,'' 2014. [Online].
  Available: \url{http://arxiv.org/abs/1412.4313}
\BIBentrySTDinterwordspacing

\bibitem{CVPR2015LIUDepth}
F.~Liu, C.~Shen, and G.~Lin, ``Deep convolutional neural fields for depth
  estimation from a single image,'' in \emph{Proc. IEEE Conf. Comp. Vis. Patt.
  Recogn.}, 2015.

\bibitem{vgg16_arxiv}
K.~Simonyan and A.~Zisserman, ``Very deep convolutional networks for
  large-scale image recognition,'' in \emph{Proc. Int. Conf. Learn.
  Representations}, 2015.

\bibitem{RCNN14}
R.~Girshick, J.~Donahue, T.~Darrell, and J.~Malik, ``Rich feature hierarchies
  for accurate object detection and semantic segmentation,'' in \emph{Proc.
  IEEE Conf. Comp. Vis. Patt. Recogn.}, 2014.

\bibitem{Saxena_nips05}
A.~Saxena, S.~H. Chung, and A.~Y. Ng, ``Learning depth from single monocular
  images,'' in \emph{Proc. Adv. Neural Inf. Process. Syst.}, 2005.

\bibitem{make3d_pami09}
A.~Saxena, M.~Sun, and A.~Y. Ng, ``{Make3D}: Learning 3d scene structure from a
  single still image,'' \emph{{IEEE} Trans. Pattern Anal. Mach. Intell.}, 2009.

\bibitem{Miaomiao_cvpr14}
M.~Liu, M.~Salzmann, and X.~He, ``Discrete-continuous depth estimation from a
  single image,'' in \emph{Proc. IEEE Conf. Comp. Vis. Patt. Recogn.}, 2014.

\bibitem{haosu_sig14}
H.~Su, Q.~Huang, N.~J. Mitra, Y.~Li, and L.~Guibas, ``Estimating image depth
  using shape collections,'' \emph{ACM Trans. Graph.}, 2014.

\bibitem{Kar_cvpr15}
A.~Kar, S.~Tulsiani, J.~Carreira, and J.~Malik, ``Category-specific object
  reconstruction from a single image,'' in \emph{Proc. IEEE Conf. Comp. Vis.
  Patt. Recogn.}, June 2015.

\bibitem{Lecun13}
C.~Farabet, C.~Couprie, L.~Najman, and Y.~LeCun, ``Learning hierarchical
  features for scene labeling,'' \emph{{IEEE} Trans. Pattern Anal. Mach.
  Intell.}, 2013.

\bibitem{Lecun_nips14}
J.~Tompson, A.~Jain, Y.~LeCun, and C.~Bregler, ``Joint training of a
  convolutional network and a graphical model for human pose estimation,'' in
  \emph{Proc. Adv. Neural Inf. Process. Syst.}, 2014.

\bibitem{Baltrusaitis2014}
T.~Baltru\v{s}aitis, L.-P. Morency, and P.~Robinson, ``Continuous conditional
  neural fields for structured regression,'' in \emph{Proc. Eur. Conf. Comp.
  Vis.}, 2014.

\bibitem{Eigen_iccv13}
D.~Eigen, D.~Krishnan, and R.~Fergus, ``Restoring an image taken through a
  window covered with dirt or rain,'' in \emph{Proc. IEEE Int. Conf. Comp.
  Vis.}, 2013.

\bibitem{Dong_eccv14}
C.~Dong, C.~C. Loy, K.~He, and X.~Tang, ``Learning a deep convolutional network
  for image super-resolution,'' in \emph{Proc. Eur. Conf. Comp. Vis.}, 2014.

\bibitem{overfeat}
P.~Sermanet, D.~Eigen, X.~Zhang, M.~Mathieu, R.~Fergus, and Y.~{LeCun},
  ``{OverFeat}: Integrated recognition, localization and detection using
  convolutional networks,'' in \emph{Proc. Int. Conf. Learn. Representations},
  2014.

\bibitem{LBP_icpr94}
T.~Ojala, M.~Pietikainen, and D.~Harwood, ``Performance evaluation of texture
  measures with classification based on kullback discrimination of
  distributions,'' in \emph{Proc. IEEE Int. Conf. Patt. Recogn.}, 1994.

\bibitem{vgg_bmvc14}
K.~Chatfield, K.~Simonyan, A.~Vedaldi, and A.~Zisserman, ``Return of the devil
  in the details: Delving deep into convolutional nets,'' in \emph{Proc.
  British Machine Vis. Conf.}, 2014.

\bibitem{nyud2_eccv12}
P.~K. Nathan~Silberman, Derek~Hoiem and R.~Fergus, ``Indoor segmentation and
  support inference from {RGBD} images,'' in \emph{Proc. Eur. Conf. Comp.
  Vis.}, 2012.

\bibitem{kitti}
A.~Geiger, P.~Lenz, C.~Stiller, and R.~Urtasun, ``Vision meets robotics: The
  {KITTI} dataset,'' \emph{Int. J. Robotics Res.}, 2013.

\bibitem{slic_pami12}
R.~Achanta, A.~Shaji, K.~Smith, A.~Lucchi, P.~Fua, and S.~S{\"{u}}sstrunk,
  ``{SLIC} superpixels compared to state-of-the-art superpixel methods,''
  \emph{{IEEE} Trans. Pattern Anal. Mach. Intell.}, 2012.

\bibitem{cbfilter}
F.~Durand and J.~Dorsey, ``Fast bilateral filtering for the display of
  high-dynamic-range images,'' \emph{ACM Trans. Graph.}, 2002.

\bibitem{googleNet}
C.~Szegedy, W.~Liu, Y.~Jia, P.~Sermanet, S.~Reed, D.~Anguelov, D.~Erhan,
  V.~Vanhoucke, and A.~Rabinovich, ``Going deeper with convolutions,'' in
  \emph{Proc. IEEE Conf. Comp. Vis. Patt. Recogn.}, 2015.

\end{thebibliography}
}

\begin{IEEEbiography}
{Fayao Liu}
received the B.Eng. and M.Eng. degrees from School of Computer Science, National University of Defense Technology, Hunan, China, in 2008 and 2010 respectively.
She is currently pursuing the Ph.D. degree at the University of Adelaide, Adelaide, Australia.
Her current research interests include machine learning and computer vision.
\end{IEEEbiography}

\begin{IEEEbiography}
  {Chunhua Shen}
is a Professor  of Computer Science at the
University of Adelaide.  His research interests are in the
intersection of computer vision and statistical machine learning.

He studied at Nanjing University, at Australian National University,
and received his PhD degree from the University of Adelaide.
Before he moved back  to the University of Adelaide in 2011, he was with the computer vision program
at NICTA (National ICT Australia), Canberra Research Laboratory for about six years.
In 2012,
he was awarded the Australian Research Council Future Fellowship.
\end{IEEEbiography}

\begin{IEEEbiography}
  {Guosheng Lin}
 is a Research Fellow at School of Computer Science at the  University of Adelaide.
His research interests are on computer vision and machine learning.
He received his PhD degree in computer vision from the University of Adelaide in 2015;
Bachelor degree and  Master degree from the
South China University of Technology in computer science in 2007 and 2010, respectively.

\end{IEEEbiography}

\begin{IEEEbiography}
  {Ian Reid}
received the BSc degree in computer
science and mathematics with first class honors
from the University of Western Australia in 1987
and was awarded a Rhodes Scholarship in 1988
in order to study at the University of Oxford,
where he received the DPhil degree in 1991. He
is a Professor of Computer Science at the
University of Adelaide.   His
research interests include active vision, visual navigation, visual
geometry, human motion capture, and intelligent visual surveillance,
with an emphasis on real-time aspects of the computations.
\end{IEEEbiography}

\end{document}